%% file: paper.tex
\documentclass{article}

\usepackage{microtype}
\usepackage{graphicx}
\usepackage{subfigure}
\usepackage{booktabs} 

\usepackage{hyperref}

\usepackage[accepted]{icml2024}

\usepackage{amsmath}
\usepackage{amssymb}
\usepackage{mathtools}
\usepackage{amsthm}

\usepackage[capitalize,noabbrev]{cleveref}

\theoremstyle{plain}
\newtheorem{theorem}{Theorem}[section]

\theoremstyle{definition}
\newtheorem{definition}[theorem]{Definition}

\theoremstyle{remark}

\usepackage[textsize=tiny]{todonotes}

\input{my_imports.tex}

\icmltitlerunning{Prospector Heads: Generalized Feature Attribution for Large Models \& Data}

\begin{document}

\twocolumn[
\icmltitle{Prospector Heads: Generalized Feature Attribution for Large Models \& Data}



\icmlsetsymbol{equal}{*}

\begin{icmlauthorlist}
\icmlauthor{Gautam Machiraju}{bds}
\icmlauthor{Alexander Derry}{bds}
\icmlauthor{Arjun Desai}{cartesia}
\icmlauthor{Neel Guha}{cs}
\icmlauthor{Amir-Hossein Karimi}{waterloo}
\icmlauthor{James Zou}{bds}
\icmlauthor{Russ B. Altman}{bds}
\icmlauthor{Christopher Ré}{cs,equal}
\icmlauthor{Parag Mallick}{rad,equal}
\end{icmlauthorlist}

\icmlaffiliation{bds}{Department of Biomedical Data Science, Stanford University}
\icmlaffiliation{cs}{Department of Computer Science, Stanford University}
\icmlaffiliation{rad}{Department of Radiology, Stanford University}
\icmlaffiliation{waterloo}{Department of Electrical \& Computer Engineering, University of Waterloo}
\icmlaffiliation{cartesia}{Cartesia AI}

\icmlcorrespondingauthor{Gautam Machiraju}{gmachi@stanford.edu}

\icmlkeywords{explainability \& interpretability, concept discovery, long-range modeling, foundation models, geometric ML, ML for science}

\vskip 0.3in
]



\printAffiliationsAndNotice{$^{*}$Joint senior authorship} 

\begin{abstract}
Feature attribution, the ability to localize regions of the input data that are relevant for classification, is an important capability for ML models in scientific and biomedical domains. Current methods for feature attribution, which rely on \say{explaining} the predictions of end-to-end classifiers, suffer from imprecise feature localization and are inadequate for use with small sample sizes and high-dimensional datasets due to computational challenges. We introduce prospector heads, an efficient and interpretable alternative to explanation-based attribution methods that can be applied to any encoder and any data modality. Prospector heads generalize across modalities through experiments on sequences (text), images (pathology), and graphs (protein structures), outperforming baseline attribution methods by up to 26.3 points in mean localization AUPRC. We also demonstrate how prospector heads enable improved interpretation and discovery of class-specific patterns in input data. Through their high performance, flexibility, and generalizability, prospectors provide a framework for improving trust and transparency for ML models in complex domains.
\end{abstract}

\section{Introduction}
\label{sec:intro}
\input{sections/1-intro}

\section{Related Work}
\label{sec:related}
\input{sections/2-related}

\section{Methods}
\label{sec:methods}
\input{sections/3-methods}

\section{Experiments}
\label{sec:exp}
\input{sections/4-exp}

\input{sections/4-results}

\section{Discussion \& Conclusion}
\label{sec:conc}
\input{sections/5-conc}

\newpage
\clearpage
\newpage

\section*{Impact Statement}

Trust and safety considerations are increasingly important as AI becomes an increasingly prominent part of high-impact disciplines such as science and biomedicine. This concern is particularly relevant for large \say{black box} foundation models. The goal of this work is to provide a new approach to feature attribution for large models and complex datasets to improve transparency of AI systems. It is important to note that that our method is specifically not designed to be an \textit{explanation} of a model's reasoning, and any feature attributions made by prospector heads should be carefully interpreted by the user in the context of the data modality.


\section*{Code Availability}
Our code is made available at: \url{https://github.com/gmachiraju/K2}.

\section*{Acknowledgements}
We thank Mayee Chen, Simran Arora, Eric Nguyen, Silas Alberti, Ben Viggiano, and  B. Anana for their helpful feedback. Gautam Machiraju is supported by the Stanford Data Science scholarship program. Neel Guha is supported by the Stanford Interdisciplinary Graduate Fellowship and the HAI Graduate Fellowship. 

We gratefully acknowledge the support of NIH under No. U54EB020405 (Mobilize), GM102365, LM012409, 1R01CA249899, and 1R01AG078755; NSF under Nos. CCF2247015 (Hardware-Aware), CCF1763315 (Beyond Sparsity), CCF1563078 (Volume to Velocity), and 1937301 (RTML); US DEVCOM ARL under Nos. W911NF-23-2-0184 (Long-context) and W911NF-21-2-0251 (Interactive Human-AI Teaming); ONR under Nos. N000142312633 (Deep Signal Processing); Stanford HAI under No. 247183; NXP, Xilinx, LETI-CEA, Intel, IBM, Microsoft, NEC, Toshiba, TSMC, ARM, Hitachi, BASF, Accenture, Ericsson, Qualcomm, Analog Devices, Google Cloud, Salesforce, Total, the HAI-GCP Cloud Credits for Research program,  the Stanford Data Science Initiative (SDSI), [STANFORD/STUDENT-SPECIFIC FELLOWSHIPS], and members of the Stanford DAWN project: Meta, Google, and VMWare. The U.S. Government is authorized to reproduce and distribute reprints for Governmental purposes notwithstanding any copyright notation thereon. Any opinions, findings, and conclusions or recommendations expressed in this material are those of the authors and do not necessarily reflect the views, policies, or endorsements, either expressed or implied, of NIH, ONR, or the U.S. Government. 

Finally, figures \ref{fig:abs}, \ref{fig:modality}, \ref{fig:prospect}, \ref{fig:layeri}, \ref{fig:layerii}, \ref{fig:unstruct}, and \ref{fig:rollup} were created with \url{BioRender.com}.



\bibliography{paper}
\bibliographystyle{icml2024}

\newpage
\appendix
\onecolumn
\setcounter{figure}{0}
\setcounter{table}{0}
\makeatletter 
\renewcommand{\thefigure}{S\@arabic\c@figure}
\renewcommand{\thetable}{S\@arabic\c@table}
\makeatother
\input{sections/appendix}

\end{document}

%% file: my_imports.tex
\usepackage{amsmath}
\usepackage{amssymb}
\usepackage{mathtools}
\usepackage{amsthm}
\usepackage{accents}
\usepackage[capitalize,noabbrev]{cleveref}

\usepackage[utf8]{inputenc} 
\usepackage[T1]{fontenc}    
\usepackage{url}            
\usepackage{booktabs}       
\usepackage{amsfonts}       
\usepackage{nicefrac}       
\usepackage{dblfloatfix}

\usepackage{enumitem}
\usepackage{comment}
\usepackage{amsmath,amssymb}
\usepackage{color}
\usepackage{url}
\usepackage{graphicx}
\usepackage{algorithm}
\usepackage{algorithmic}
\usepackage[font=footnotesize,skip=0pt,labelfont=bf]{caption}
\PassOptionsToPackage{hyphens}{url}
\usepackage{lipsum}
\usepackage{orcidlink}
\usepackage{cite}
\usepackage{booktabs}
\usepackage{subcaption}
\usepackage{wrapfig}
\usepackage{verbatim}
\usepackage{mathtools} 
\usepackage{mathrsfs}
\usepackage{bbm}  
\usepackage{amsfonts}
\usepackage{physics}
\usepackage{lipsum}  
\usepackage{graphics}
\usepackage{enumitem}
\usepackage[capitalize]{cleveref} 
\usepackage{multirow}
\usepackage{amssymb}
\usepackage{pifont}
\usepackage{dirtytalk}
\usepackage{chngcntr}

\usepackage[normalem]{ulem}

\newcommand{\xmark}{\ding{55}}%
\usepackage{xargs}                      

\usepackage{titlesec}
\titlespacing*{\section}{0pt}{0.7\baselineskip}{0.4\baselineskip}
\usepackage{microtype}

\newcommand{\ie}{\textit{i}.\textit{e}., }
\newcommand{\eg}{\textit{e}.\textit{g}., }

\newcommand{\sprite}{S}
\newcommand{\token}{\mathbf{x}}
\newcommand{\feature}{\mathbf{z}}

\newcommand{\conv}{\texttt{K2conv} }

%% file: sections/1-intro.tex


Most ML models are optimized solely for predictive performance, but many applications also necessitate models that provide insight into features of the data that are unique to a particular class. This capability is known as \textit{feature attribution}, which in unstructured data (\eg text, images, graphs) consists of identifying subsets of the input datum most responsible for that datum’s class membership (\eg pixels or patches of an image, often represented as a heatmap). Feature attribution is especially important for scientific and biomedical applications. For example, for a model to assist a pathologist in making a cancer diagnosis, it ideally should not only accurately classify which images contain tumors, but also precisely locate the tumors in each image \citep{Song2023-sq, Niazi2019-ue}.
\begin{figure}
    \centering
    \vspace{0pt}
    \includegraphics[width=0.48\textwidth, trim={0 0 0 0},clip]{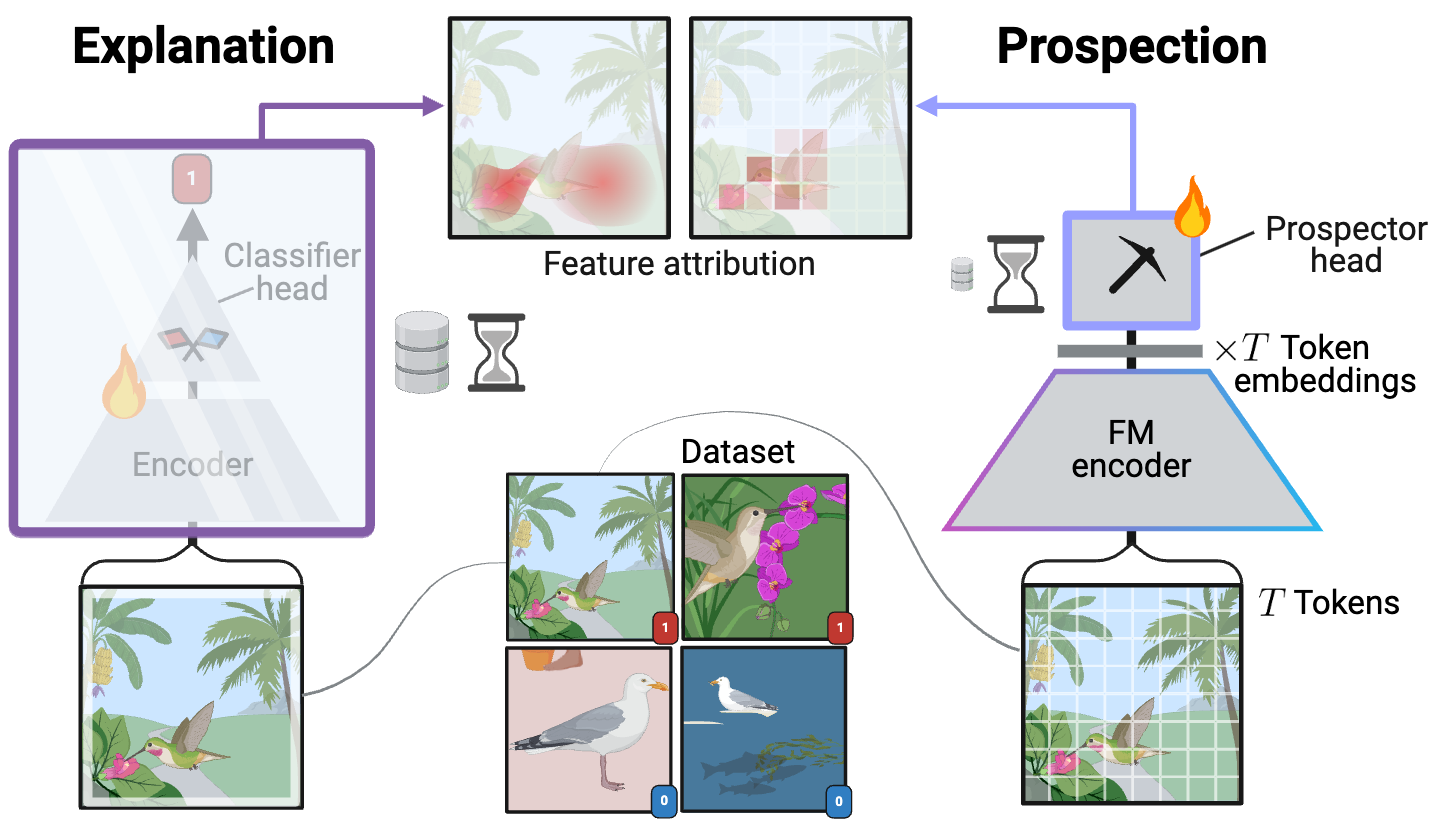}
    \vspace{-7pt}
    \setlength{\belowcaptionskip}{-16pt}
    \caption{Explanation-based attribution can be conceptualized as a \say{wrapper function} for trained classifiers using internals, forward or backward passes, or input perturbations. Prospector heads are instead encoder-equippable like classifier heads and adapt token embeddings with data- and time-efficiency. Flame icon indicates trainable parameters.}
    \label{fig:abs}
\end{figure}

Unfortunately, modern ML systems can struggle to perform feature attribution. Most existing attribution techniques attempt to provide \say{explanations} for trained classifiers (\cref{fig:abs}) — descriptions of how model weights interact with different input features (\eg gradients \citep{Simonyan2014-ta}, attention \citep{Jetley2018-aw}) or of how each feature contributes to prediction (\eg SHAP \citep{Lundberg2017-bw}, LIME \citep{Ribeiro2016-hq}). Explanation-based attribution methods are inherently (a) data-inefficient as they require ample labeled training data to train underlying classifiers. Additionally, methods producing explanations can themselves be (b) computationally inefficient \citep{Ancona2019-kg,Chen2023-ak} and thus may not actually improve tractability relative to annotation by domain experts, particularly for large inputs. Finally, (c) the attributed features are often found to be inaccurate and irrelevant to target classes \citep{Arun2020-jp, Zech2018-vb, Jain2019-to, Zhou2021-lb, Bilodeau2022-lm}.

We explore whether foundation models (FMs) can be used to solve challenges (a–c) without traditional explanations. Prior work demonstrates that FMs learn high quality data representations and can learn class-specific properties through a few labeled examples \citep{Bommasani2021-ca, Brown2020-bh, Gondal2023-bg}. However, it is unclear whether FM representations can be used to perform feature attribution in a scalable and accurate manner. Our key insight is to build on top of FM representations, rather than explain an FM fine-tuned as an end-to-end classifier.


In this work we present \textit{prospector heads} (a.k.a. \say{prospectors}), simple modules that aim to \textit{equip} feature attribution to any encoder — including FMs — just as one would equip classification heads. Prospectors inductively reason over two layers: layer (I) categorizes learned representations into a finite set of \say{concepts} and layer (II) learns concepts' spatial associations and how those associations correlate with a target class. To (a) enable data efficiency, prospectors are parameter-efficient and with only hundreds of parameters. To (b) limit time complexity, prospectors operate with efficient data structures and linear-time convolutions, all without model backpropagation. To (c) improve attribution accuracy, prospector heads are explicitly trained to perform feature attribution, unlike explanation methods.

We show that prospector heads outperform attribution baselines over multiple challenging data modalities. Prospector-equipped models achieve gains in mean area under the precision-recall curve (AUPRC) of 8.5 points in sequences (text documents), 26.3 points in images (pathology slides), and 22.0 points in graphs (protein structures) over the top modality-specific baselines. Additionally, we show that prospector-equipped FMs are particularly robust to variation in the prevalence and dispersion of class-specific features. Finally, we also present visualizations of prospectors' internals and outputs to demonstrate their interpretability in complex domain applications.

%% file: sections/2-related.tex
To adequately motivate our approach (\cref{sec:methods}), this section focuses on central methods ideas. We present a full version of Related Work, including baselines, in \cref{supp:rw}.

\textbf{Feature attribution via explanation:} In the current explanation-based paradigm, feature attribution is performed by (1) training a supervised model before (2) interrogating the model's behavior (\eg via internals, forward or backward passes, or input perturbations) and inferring class-specific features. Both model-specific (\eg gradients \citep{Simonyan2014-ta}, attention maps \citep{Jetley2018-aw}) and model-agnostic (\eg SHAP \citep{Lundberg2017-bw}, LIME \citep{Ribeiro2016-hq}) methods of today are either computationally prohibitive \citep{Ancona2019-kg,Chen2023-ak} or poor localizers of class-specific features \citep{Arun2020-jp, Zech2018-vb, Jain2019-to, Zhou2021-lb, Bilodeau2022-lm}.

\begin{figure}
    \centering
    \vspace{0pt}
    \includegraphics[width=0.48\textwidth, trim={0 0 0 0},clip]{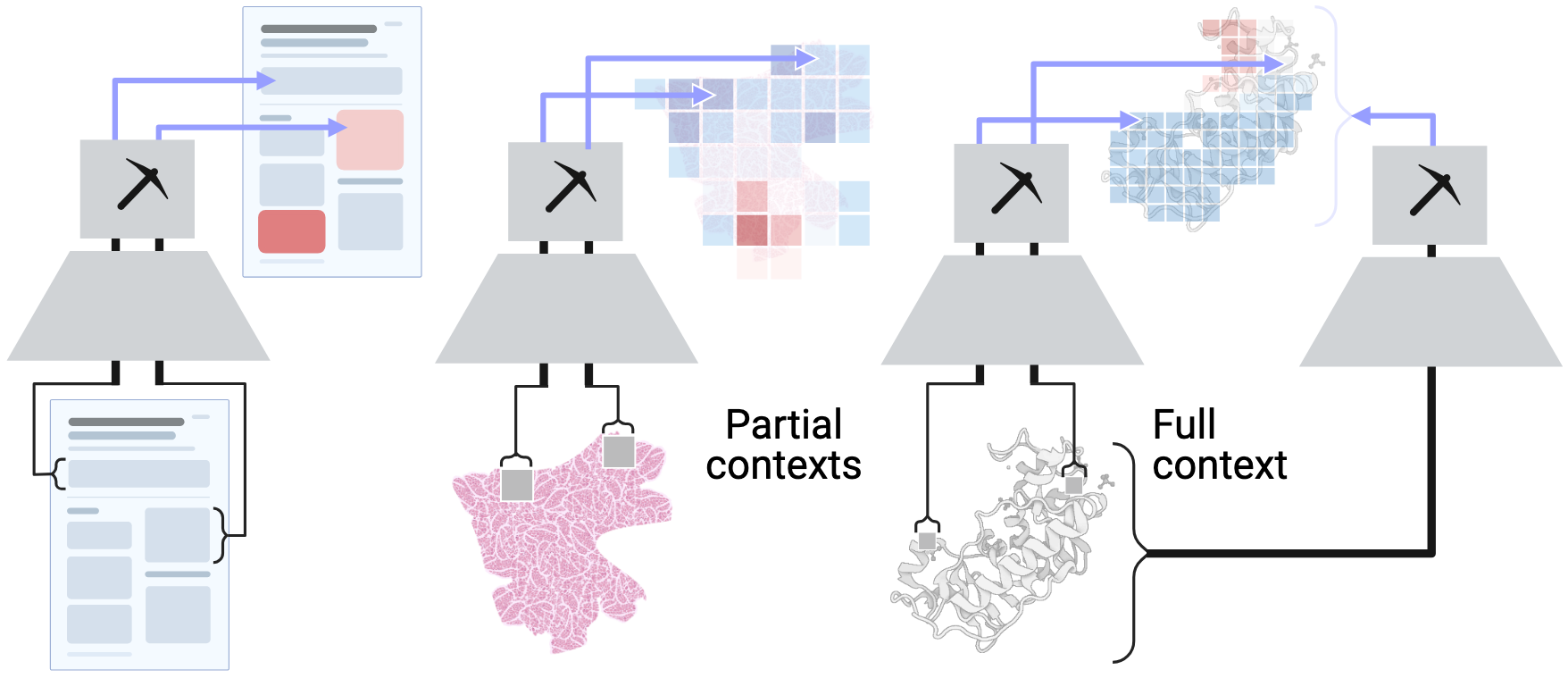}
    \vspace{-5pt}
    \setlength{\belowcaptionskip}{-10pt}
    \caption{Prospectors are modality-generalizable, amenable to sequences (\eg text), images (\eg pathology), and graphs (\eg protein structures). They can also operate on embeddings from either partial- or full-context encoders.}
    \label{fig:modality}
\end{figure}

\textbf{Modern encoders \& context sizes:} 
Most modern encoders for unstructured data operate on \textbf{tokens}, or relatively small pieces of a datum, and their representations. Tokens can be user-prespecified and/or constructed by the encoder itself (potentially with help from a tokenizer), where these encoders are respectively referred to as \textit{partial-context} and \textit{full-context} (\cref{fig:modality}). Due to computational constraints, high-dimensional unstructured data (\eg gigapixel images) often require user-prespecified tokens (\ie patches) and partial-context encoders that embed each token \citep{Lu2023-ar, Huang2023-iv, Klemmer2023-oo, Lanusse2023-xt}. 

Gradient-based saliency and attention maps have been used to explain partial-context classifiers for high-dimensional unstructured data like gigapixel imagery \citep{Campanella2019-sj, Chen2022-rv}. However, studies report low specificity and sensitivity \citep{Machiraju2022-yd} in part because attribution for the entire datum is built by concatenating attributions across prespecified tokens. Partial-context strategies incorrectly assume prespecified tokens are independent and identically distributed (IID).





\textbf{Concept-based modeling:} The use of \textbf{concepts} in ML inherently increases model interpretability by forcing models to reason over unstructured data with respect to said concepts. Concepts themselves can be human-derived, machine-derived \citep{Ghorbani2019-cd, Talukder2024-ou}, or co-derived with humans in the loop \citep{Lam2024-bt}.

Early concept-based methods examine models' use of concepts in prediction \citep{Kim2017-ze}, while recent methods can also attribute concept importance \textit{in situ} \citep{Ghorbani2019-cd, Crabbe2022-lh, Brocki2019-fb, Zhou2018-dz}. Sets of concepts can also form a hidden layer, \ie \say{bottleneck} \citep{Koh2020-bh, Kim2024-wq, Talukder2024-ou}, offering a form of multi-modal grounding when concepts are human-derived. More recently, concepts are being assigned to pre-specified tokens in high-dimensional data, \eg subsequences \citep{Talukder2024-ou} and sentences \citep{Lam2024-bt}. These \say{spatially resolved} concepts have allowed for hierarchical concept formation when paired with LLMs \citep{Lam2024-bt}.

%% file: sections/3-methods.tex
\subsection{Prospection: Attribution sans Explanation}
Prospectors are designed to perform few-shot feature attribution for high-dimensional data while meeting challenges (a-c). Instead of explaining end-to-end classifiers, prospectors interface with encoders by adapting their \textit{token embeddings}. Crucially, prospectors foster a form of inductive reasoning over token embeddings to learn class-specific patterns. The use of tokens as the core unit of analysis depends on the key assumption that the equipped encoders have learned adequate distributional semantics in large-scale pretraining. Prospectors can then learn class-specific patterns in small labeled datasets via a simple two-layer module. In layer (I), prospectors transform token embeddings into spatially resolved concepts learned from the training set, constructing a parsimonious \say{vocabulary} or \say{codebook} that can be user-verified and/or user-defined. Layer (II) then attributes scores to each token using a novel form of graph convolution that operates on concept frequencies and co-occurrences. The following sections describe the inference and fitting procedures of each layer. 
\begin{figure}
    \centering
    \vspace{0pt}
    \includegraphics[width=0.36\textwidth, trim={0 0 0 0},clip]{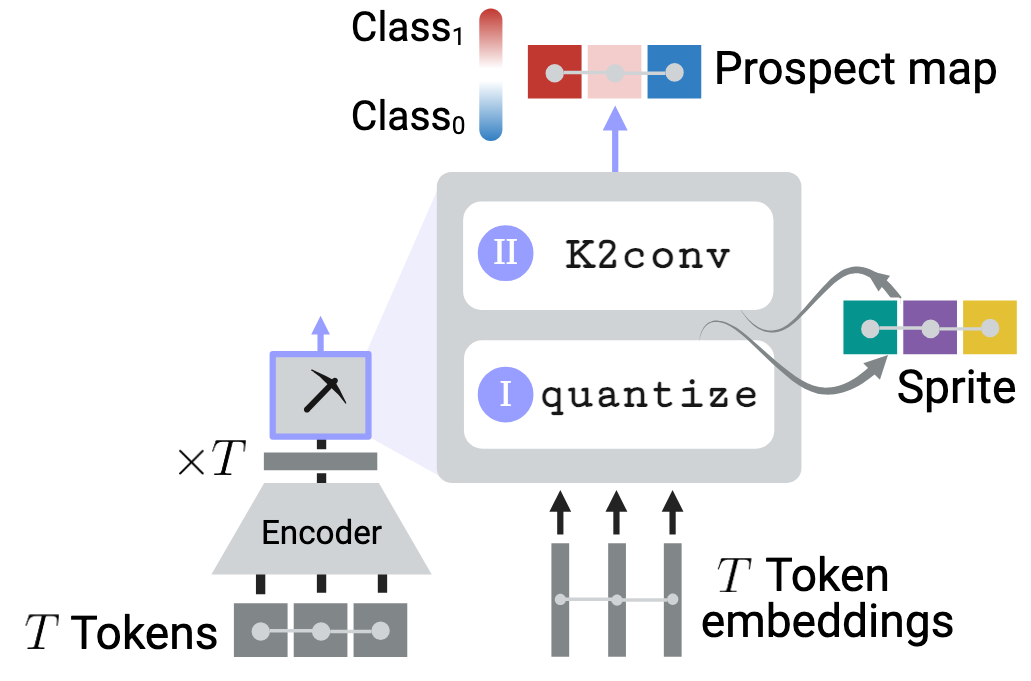}
    \vspace{5pt}
    \setlength{\belowcaptionskip}{-12pt}
    \caption{Prospector-equipped encoders produce attribution maps (called \say{prospect maps}) over two layers. Details for fitting and inference are in Sections \ref{sec:inference}, \ref{sec:fit}, and \ref{supp:internals}.}
    \label{fig:prospect}
\end{figure}

\subsection{Preliminaries}
\label{sec:prelims}
To enable any encoder to perform feature attribution regardless of input modality, we first define a generalized language for unstructured data. Any unstructured datum can be represented by a \textit{map graph} $G(\mathcal{V},\mathcal{E})$ where each vertex $v \in \mathcal{V}$ represents a discrete \textit{token}, or piece of that datum in Euclidean space (\cref{def:map}). $G$ is composed of $T = |\mathcal{V}|$ tokens. For example, in image data, tokens can be defined as pixels or patches. An edge $e_{i \leftrightarrow j} \in \mathcal{E}$ connects vertex $v_i$ to $v_j$. Both $G$'s token resolution and token connectivity are defined based on data modality (\cref{fig:unstruct}). 

\textbf{Problem setup:} Suppose we have a dataset containing map graphs $G$ and binary class labels $y$. We assume that a class$_y$ graph $G(\mathcal{V},\mathcal{E})$ contains a set of class$_y$-specific vertices $\mathcal{V}_y \subseteq \mathcal{V}$, with $|\mathcal{V}_y| \geq 1$ \citep{Zhou2016-wy}. One main goal of feature attribution is to locate $\mathcal{V}_y$ in each datum given a set of $(G,y)$ pairs as a training dataset. This task is inherently \textit{coarsely supervised} \citep{Robinson2020-ef} and is discussed further in \cref{supp:rw}.

\subsection{Prospector Inference}
\label{sec:inference}
\subsubsection{Receiving Token Embeddings}
Prospectors receive token embeddings $\token_1 \ldots \token_T$ from an equipped encoder and update map graph $G$ such that each vertex $v_i \in \mathcal{V}$ is featurized by an embedding $\token_i \in \mathbb{R}^d$. This vertex-specific \say{feature loading} uses the notation: $G[v_i] := \mathbf{x}_i$. Details for partial- and full-context encoders are specified in \cref{supp:definitions}.

\subsubsection{Layer I: Quantizing Embeddings}
\begin{figure}
    \centering
    \vspace{0pt}
    \includegraphics[width=0.4\textwidth, trim={0 0 0 0},clip]{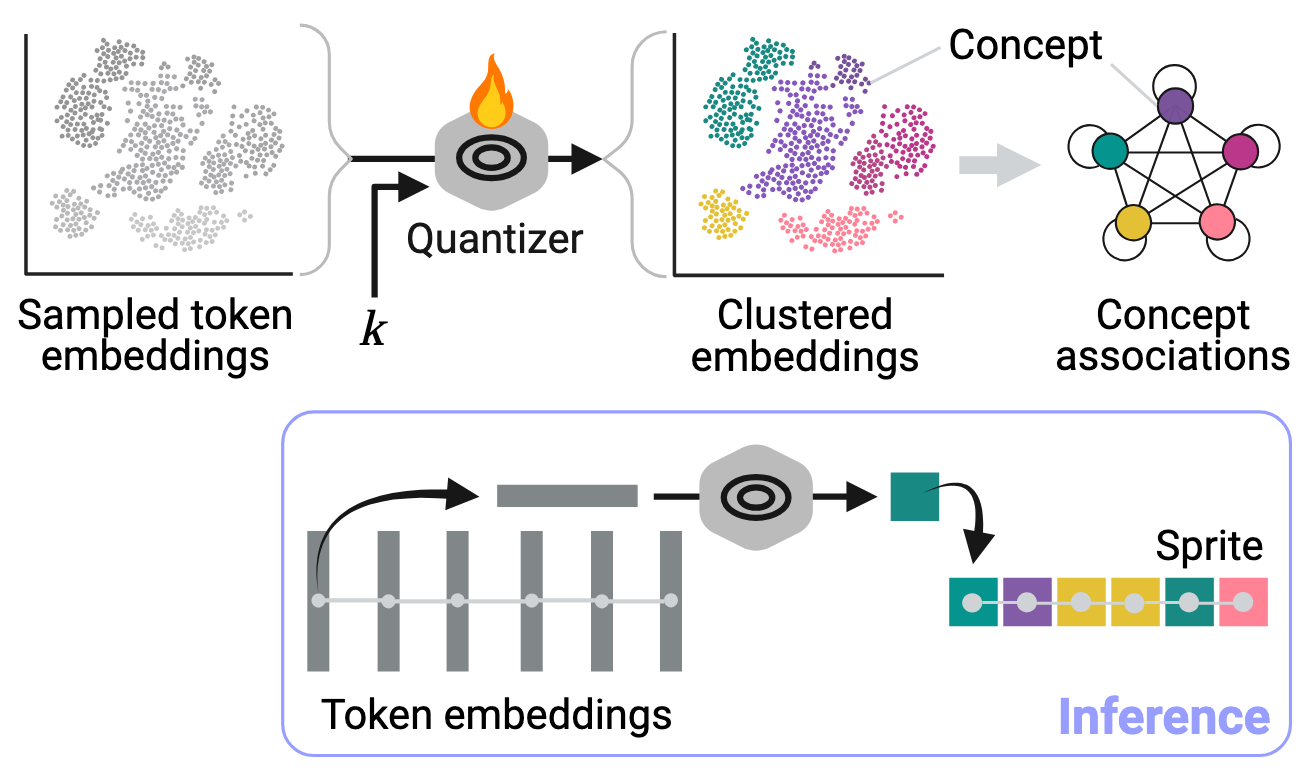}
    \vspace{5pt}
    \setlength{\belowcaptionskip}{-13pt}
    \caption{Layer (I) fitting and inference ($K=5$). Quantized token embeddings define spatially resolved concepts, which together form data sprites.}
    \label{fig:layeri}
\end{figure}

Next, prospectors use an encoder’s learned semantics to define $K$ spatially resolved concepts $\mathcal{C} = \{1,\ldots,K\}$. This is achieved by quantizing each token embedding $\token \in \mathbb{R}^d$ as a scalar concept $c \in \mathcal{C}$: $c_i = \texttt{quantize}(\token_i) \: \forall i=1\ldots T.$ When the \texttt{quantize} layer (\cref{sec:fit_quantize}) is applied over the full graph $G$, it is transformed into graph $S$ with the same topology as $G$, but with categorical vertex features $S[v_i] := c_i \: \forall i$. We refer to $S$ as a \textit{data sprite} due to its low feature dimensionality compared to $G$ (a data compression ratio of $d$). Intuitively, the heterogeneity of $S$ is parameterized by the choice of $K$. This layer is depicted in \cref{fig:layeri}.

\subsubsection{Layer II: Convolution over Concepts}
\label{sec:k2conv}


Prospectors next perform feature attribution using a form of graph convolution over sprite $S$. This convolution requires a global kernel $\omega$ that computes an attribution score $a \in \mathbb{R}$ for each vertex $v$ based on the concepts $c_i$ (\ie monograms) and co-occurrences $c_i,c_j$ (\ie skip-bigrams) present within the graph neighborhood defined by receptive field $r$. The kernel $\omega$ serves as a form of \textit{associative memory} and can be conceptualized as a dictionary, scoring each concept monogram or skip-bigram in the combinatorial space $\mathcal{Z} = \mathcal{C} \cup \{ \mathcal{C} \otimes \mathcal{C}\}$, where $\otimes$ is the Cartesian product. The global kernel is fit over the training set (\cref{sec:fit_kernel}).

\begin{figure}
    \centering
    \vspace{0pt}
    \includegraphics[width=0.48\textwidth, trim={0 0 0 0},clip]{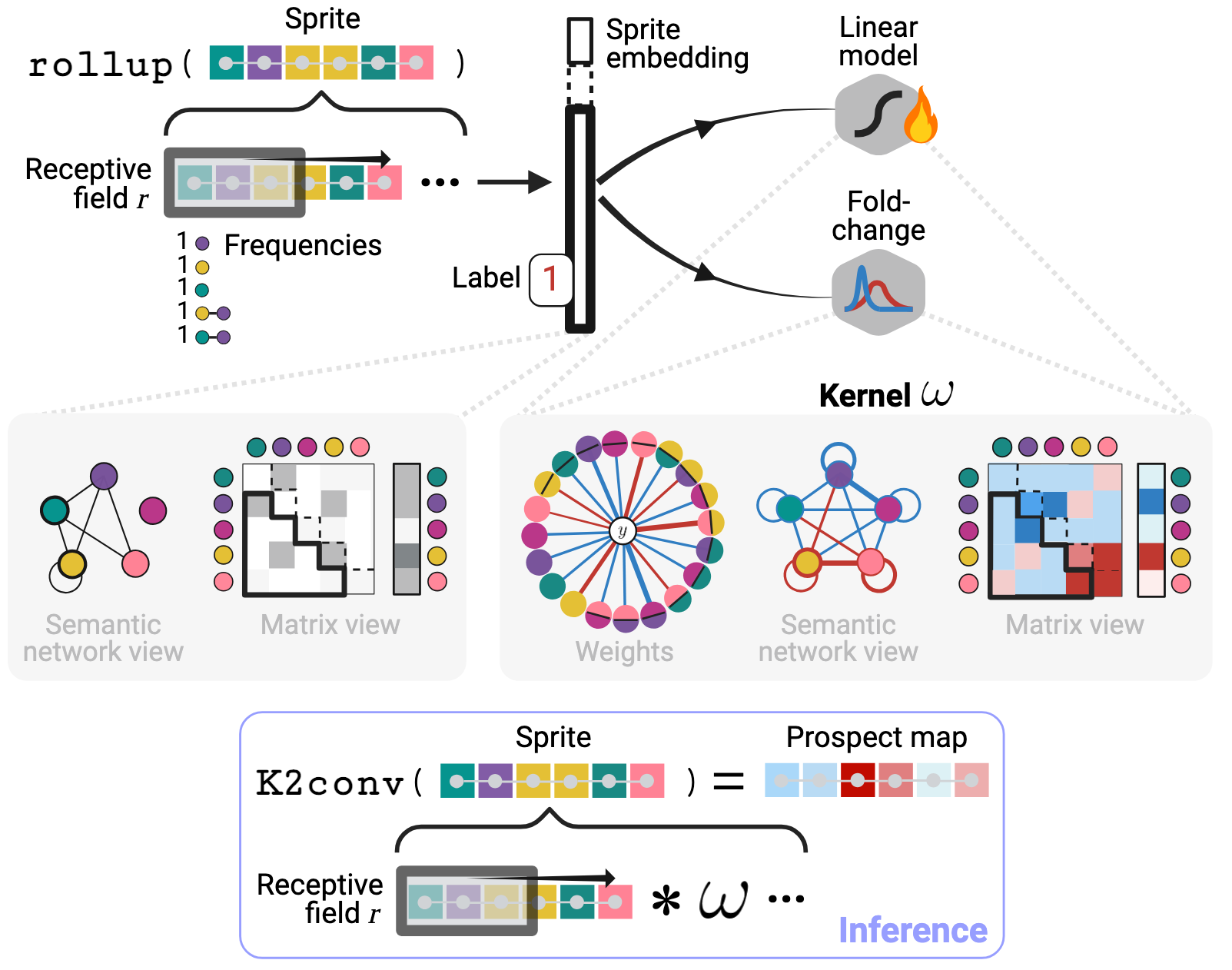}
    \vspace{-5pt}
    \setlength{\belowcaptionskip}{-5pt}
    \caption{Layer (II) fitting and inference ($K=5$). Concept frequencies are used to build sprite embeddings, which are used to fit a \texttt{K2conv} kernel. Flame icon indicates trainable parameters.} 
    \label{fig:layerii}
\end{figure}

To perform feature attribution at inference time, we apply the fitted kernel over each vertex in a datum to produce a \textit{prospect map} $P$. $P$ is a map graph with the same topology as $G$ and $S$ but featurized by scalar continuous attribution scores $P[v] := a$. We call this layer \conv in reference to kernel $\omega$'s implicit structure (\cref{def:self-complete}). An attribution score $a_i$ is computed for each vertex $v_i$ in $S$, where $\mathcal{N}_r$ represents all vertices within the $r$-neighborhood of $v_i$ (including $v_i$ itself): \begin{eqnarray*}
    P[v_i] &:=& a_i = \overbrace{\mathcal{N}_r * \omega}^\text{\conv} \\
    &=& \smashoperator{\sum_{\forall v_i \in \mathcal{N}_r}} \omega \left< \overbrace{\sprite[v_i]}^{c_i} \right> 
    + \: \smashoperator{\sum_{\forall(v_j,v_k) \in \mathcal{N}_r}} \omega \left< (\overbrace{\sprite[v_j]}^{c_j}, \overbrace{\sprite[v_k]}^{c_k}) \right>,
\end{eqnarray*}
where $\omega \langle \cdot \rangle$ denotes dictionary lookup. The above expression resembles the energy function for 2D Markov random fields, but adjusted to allow for longer-range dependencies in the second term via skip-bigrams. The resulting prospect map $P$ targets class-specific region $\mathcal{V}_y$ by assigning high absolute positive or negative values to each token. Intuitively, $r$ parameterizes the level of smoothing over $P$ by modulating the number of neighboring tokens used to compute a token's importance. This layer is depicted in \cref{fig:layerii}.

\subsection{Prospector Fitting}
\label{sec:fit}
In our implementation, layers (I) and (II) are fitted separately and sequentially using the procedures below. Further details for each layer are found in \cref{supp:internals}.

\subsubsection{Quantizer Fitting (Layer I)}
\label{sec:fit_quantize}
Token embeddings sampled from across the training set are partitioned into $K$ subspaces using an unsupervised algorithm (\eg $K$-means clustering). Afterward, each subspace represents a semantic concept $c \in \mathcal{C}$ discovered in the corpus. To reduce computation, clustering can be performed over a representative sample ($>10^3$) of the token embedding space, randomly sampled without replacement. Fitting is depicted in \cref{fig:layeri}.

\subsubsection{Kernel Fitting (Layer II)}
\label{sec:fit_kernel}
Fitting the \conv kernel involves computing the class-attribution weights for each monogram and skip-bigram in $\mathcal{Z}$ across the training set. These weights represent the only learnable parameters of a prospector head. The total number of parameters $|\mathcal{Z}|$ is thus dependent on $K$ and is at maximum (\cref{supp:maxz}): $|\mathcal{Z}| = 2K + {K \choose 2}$. The kernel is fit in two steps, as outlined below. 

\textbf{Step 1: Computing frequencies \& co-occurrences.} For each sprite $S$ in the training set, prospectors first build a datum-level representation in order to learn dataset-wide patterns. This is performed by the \texttt{rollup} operator, which traverses $S$’s vertices, tracks concept monogram and skip-bigrams $z_i \in \mathcal{Z}$, and counts their frequencies over all $r$-neighborhoods. This operation constructs a \textit{sprite embedding} $\feature \in \mathbb{R}^{|\mathcal{Z}|}$, which resemble \say{bag-of-words} vectors with longer-range \say{skip} interactions. Sprite embeddings are rescaled to account for differences in baseline frequencies (\eg using TF-IDF \citep{Sparck_Jones1972-ke}) and thus can be viewed as probabilities: $\mathbb{P}(c_i)$ for monograms and $\mathbb{P}(c_j, c_k)$ for skip-bigrams. The \texttt{rollup} operator and this step as a whole are described in \cref{alg:rollup} and \cref{fig:rollup}.


\textbf{Step 2: Learning kernel weights.} Prospectors next use the datum-level sprite embeddings $\feature$ to learn a vector $\mathbf{w} \in \mathbb{R}^{|\mathcal{Z}|}$ of class-specific weights for each monogram and skip-bigram across the entire training set. After fitting $\mathbf{w}$, we construct $\omega$ as a dictionary mapping each element in $\mathcal{Z}$ to its corresponding weight in $\mathbf{w}$. We implement two approaches to learning weights, which make up the two main prospector variants: a linear classifier $h_\mathbf{w}$ and a parameter-free fold-change computation. These variants are discussed further in \cref{supp:variants} and depicted graphically in \cref{fig:layerii}.

\underline{Linear classifier:} This variant trains a linear classifier $h_\mathbf{w}(\feature) = \mathbf{w}^\intercal \feature$ to learn a mapping from $\feature \mapsto y$ over the training dataset. The learned coefficients $\mathbf{w}$ then represent the class-specific importance of each index in $\feature$. We implement this as a logistic regression with elastic net regularization with the mixing hyperparameter $\lambda$.

\underline{Fold-change computation:} Inspired by bioinformatics \citep{Anders2010-zg}, this variant involves first computing mean sprite embeddings for each class over the training data. For example, for the negative class, $\overline{\feature}_0 = \frac{1}{|\mathcal{D}_0|} \sum_{S^{(i)} \in \mathcal{D}_0}{\feature^{(i)}}$, where $\mathcal{D}_0$ is the subset of the training dataset $(S^{(i)}, y^{(i)})$ for which $y^{(i)} = 0$. This mirrors the \say{baseline vector} commonly used by popular feature attribution methods \citep{Sundararajan2017-ps,Bilodeau2022-lm,Afchar2021-xb}. Then, we compute $\mathbf{w}$ as a fold-changes $\mathbf{w} = \operatorname{log}_2(\overline{\feature}_1) - \operatorname{log}_2(\overline{\feature}_0)$ and select significant weights using a hypothesis test for independent means. The latter step serves as a form of regularization.



\subsection{Meeting Challenges with Intentional Design}
\label{sec:principles}
Prospectors overcome the limitations of current feature attribution methods by observing the following design principles. Firstly, for (a) data efficiency and few-shot capabilities, prospectors are parameter efficient due to the sole use of concept monograms and skip-bigrams to build its kernel — at maximum only requiring $2K + {K\choose 2}$ parameters. Both variants for computing importance weights $\mathbf{w}$ are thus data efficient due to their parsimony. Secondly, prospectors are (b) computationally efficient: by operating as an equippable head, prospectors are \say{plug-in-ready} without encoder retraining \citep{Kim2017-ze} and or backpropagation. The combination of efficient data structures and modeling primitives such as dictionaries and convolutions allow prospectors to efficiently scale feature attribution to high-dimensional data: namely, linear-time with respect to the tunable number of tokens $T$. We outline runtime complexity and speed benchmarking in Sections \ref{supp:runtime} and \ref{supp:speed}. Finally, prospectors achieve (c) improved localization and class-relevance by explicitly training on token embeddings to learn $G_y$ instead of using end-to-end classifiers to identify $G_y$ \textit{post hoc}. We detail other favorable model properties in \cref{supp:props}.

%% file: sections/4-exp.tex


\subsection{Datasets, Encoders, \& Baselines}
\label{sec:datasets}
We evaluate prospectors using three primary tasks, each representing a different data modality (sequences, images, and graphs). Each also poses unique challenges for prospector training and feature attribution: class imbalance (sequences), high input dimensionality with few examples (images), and very coarse supervision (graphs). As is common in scientific and biomedical data, all three datasets are amenable to the \textit{multiple instance assumption} (MIA) — that class$_1$ data largely resemble class$_0$ data with the exception of tokens only found in class$_1$ data \citep{Amores2013-rk, Foulds2010-qi}. Details for each dataset's construction are shared in \cref{supp:data_construct}. For each task, we select representative encoders to which we equip prospector heads and relevant baseline attribution methods. We summarize encoders in \cref{tab:encoders} (and \cref{supp:all_modeling}), baselines in \cref{supp:baseline}, and ruled-out baselines in \cref{supp:oos-baselines}. 
\begin{table}
    \centering
    \vspace{-2pt}
    \scriptsize
    \begin{tabular}{llccc}
        \toprule
        \parbox{0.8cm}{\textbf{Encoder\\Alias}} & \textbf{Architecture} & \parbox{1cm}{\textbf{Learning\\Regime}} & \parbox{0.8cm}{\textbf{Training\\Epochs}} & \parbox{0.8cm}{\textbf{Embed\\Size ($d$)}} \\
        \midrule
        MiniLM & MiniLM-L6-v2 & KD & \xmark & 384 \\
        DeBERTa & DeBERTa-v3-base & SSL & \xmark & — \\
        \midrule
        tile2vec & ResNet-18 & USL & 20 & 128 \\
        ViT & ViT/16 & WSL & 30 & 1024  \\
        CLIP & ViT-B/32 & SSL & \xmark & 512 \\
        PLIP & ViT-B/32  & SSL & \xmark & 512 \\
        \midrule
        COLLAPSE & GVP-GNN & SSL & \xmark & 512 \\
        ESM2 & t33\_650M\_UR50D  & SSL & \xmark & 1028  \\
        AA & — & — & \xmark & 21 \\
        \bottomrule
    \end{tabular}
    \vspace{5pt}
    \caption{Prospector-equipped encoders in descending order by modality: sequences, images, and graphs. Learning regimes are knowledge distillation (KD), unsupervised learning (USL), weakly supervised learning (WSL), and self-supervised learning (SSL). Pre-training denoted by \xmark. Non-applicability denoted by \say{—}. \vspace{-10pt}}
    \label{tab:encoders}
\end{table} 

For both baselines and prospectors, we perform a gridsearch over tunable hyperparameters. Due to the MIA, the best models were selected based on their ability to localize ground truth class$_1$ regions in the training set, since these were not seen by prospectors during training. We use a sequential ranking criteria over four token-level metrics: precision, dice coefficient, Matthews correlation coefficient, and AUPRC. Details of hyperparameter tuning and model selection are found in \cref{supp:hparam} and \ref{supp:model-select}. The results in the remainder of this paper present the localization AUPRC and average precision (AP) over a set of thresholds, for class$_1$ regions in our held-out test data. 
%

\textbf{Sequences (1D): key sentence retrieval in text documents.} Retrieval is an important task in language modeling that provides in-text answers to user queries. For this task, we use the WikiSection \citep{Arnold2019-sz} benchmark dataset created for paragraph-level classification. We repurpose WikiSection to assess the ability to retrieve target sentences specific to a queried class. We specifically use the \say{genetics} section label as a query, and class$_1$ data are defined as documents in the English-language \say{disease} subset that contain this section label. Our goal is to identify sentences that contain genetics-related information given only coarse supervision from document-level labels. After preprocessing the pre-split dataset, our dataset contained 2513 training examples (2177 in class$_0$, 336 in class$_1$) and 712 test examples. The relationship between sentences in each document is represented as a graph with 2-hop connectivity (\cref{fig:unstruct}). 

\begin{figure*}
    \centering
    \vspace{0pt}
    \includegraphics[width=0.98\textwidth, trim={0 14em 0 6em},clip]{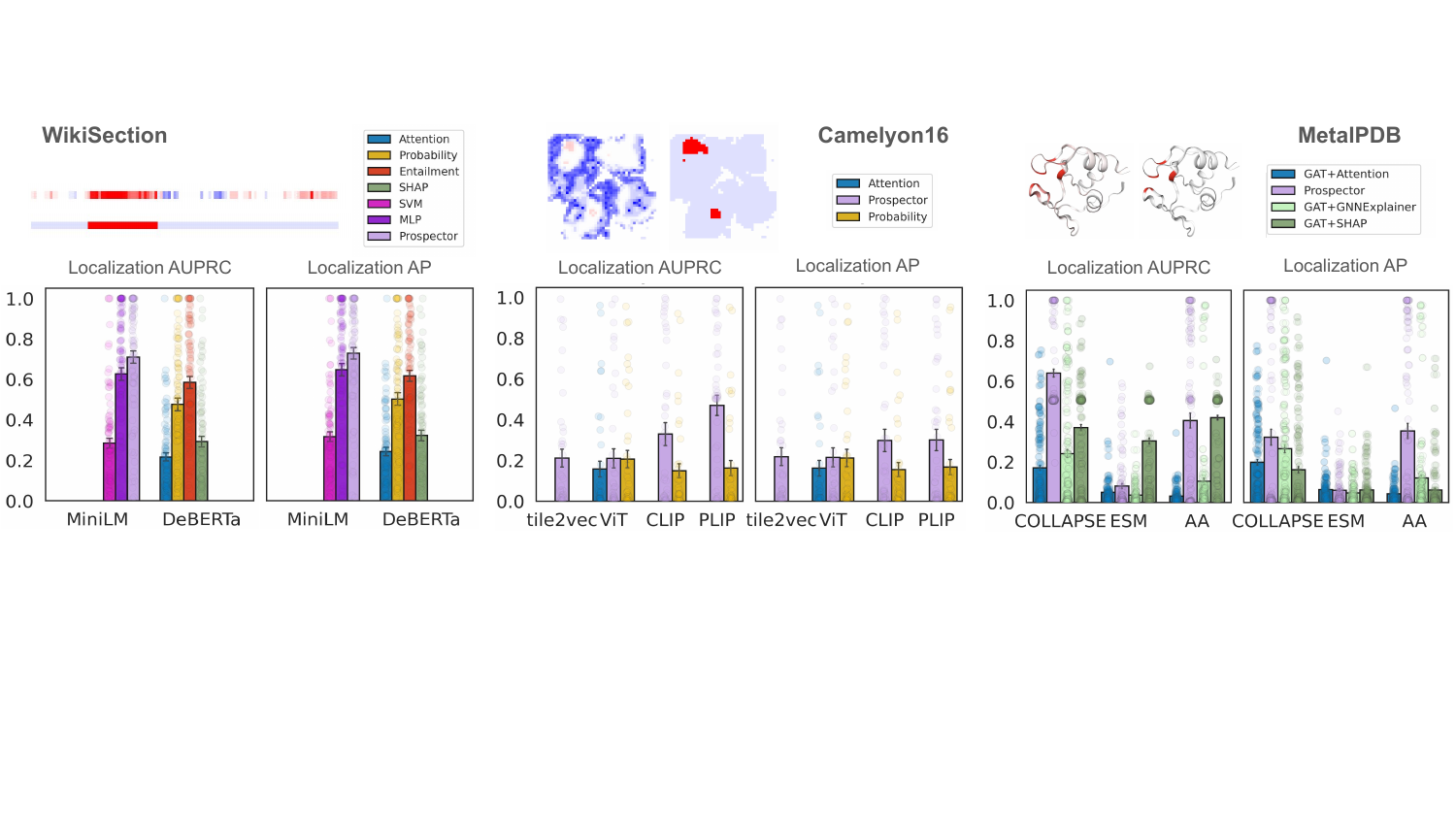}
    \vspace{4pt}
    \setlength{\belowcaptionskip}{-10pt}
    \caption{Prospectors vs. baselines for sequences (left), images (middle), and graphs (right). Dots represent performance on individual test-set examples, while bars represent means with whiskers as standard errors. Numerical results are found in \cref{supp:tab}.}
    \label{fig:perf}
\end{figure*}

\underline{Encoders \& baselines:} We assess two pretrained language models, MiniLM \citep{Wang2020-zz} and DeBERTa \citep{He2020-nn,He2021-wi}, used at partial-context. While DeBERTA is an off-the-shelf LLM for zero-shot classification (ZSC) and natural language inference (NLI), MiniLM is a sentence and paragraph embedding model — thus requiring prospectors to perform feature attribution at the sentence-level.

For baselines attribution methods, we present a mix of (1) supervised heads and (2) off-the-shelf LLM inference. Firstly, supervised heads train on token-level class labels to identify class-specific sentences in testing. Specifically, we train a multi-layer perception (MLP) on labeled token embeddings and a one-class support vector machine (SVM) trained solely on class$_0$ token embeddings. In the latter case, we perform novelty detection to identify class$_1$ tokens. While not traditional explanation methods, the MLP and SVM heads are given a large advantage as semi- and fully supervised baselines (as opposed to prospectors, which are coarsely supervised at the datum-level). For LLM inference, we used DeBERTA to output sentence-level ZSC probabilities (\ie logits), NLI entailment scores, NLI entailment attention, and pooled Shapley values for ZSC. Implementation details are listed in \cref{supp:baseline}.


\textbf{Images (2D): tumor localization in pathology slides.} Identifying tumors is an important task in clinical pathology, where manual annotation is standard practice. We evaluate prospectors on Camelyon16 \citep{Ehteshami_Bejnordi2017-ko}, a benchmark of gigapixel pathology images, each presenting either healthy tissue or cancer metastases. All images are partitioned into prespecified $224 \times 224$ patch tokens and filtered for foreground tissue regions. After pre-processing the pre-split dataset, our dataset contained 218 images for training (111 for class$_0$ and 107 for class$_1$) and 123 images for testing. The relationship between patches in each image is represented as a graph using up to 8-way connectivity (\cref{fig:unstruct}). 

\underline{Encoders \& baselines:} We equip prospectors to four encoders: tile2vec \citep{Jean2019-np}, ViT \citep{Dosovitskiy2020-wn}, CLIP \citep{Radford2021-ln}, and PLIP \citep{Huang2023-iv}. The first two encoders are trained with partial context, where tile2vec is unsupervised while ViT is weakly supervised with image-level label inheritance \citep{Machiraju2022-yd}. Details on encoder training are provided in \cref{supp:all_modeling}. CLIP serves as a general-domain vision-language foundation model (VLM) and PLIP serves as a domain-specific version of CLIP for pathology images. Both VLM encoders are pretrained and used for partial-context inference on prespecified image patches. We choose two popular and computationally feasible explanation-based attribution baselines (\cref{sec:related}): concatenated mean attention \citep{Chen2022-rv} for ViT and concatenated prediction probability \citep{Campanella2019-sj, Machiraju2022-yd, Halicek2019-ru} for ViT, CLIP, and PLIP. 

\textbf{Graphs (3D): binding site identification in protein structures.} Many proteins rely on binding to metal ions in order to perform their biological functions, such as reaction catalysis in enzymes, and identifying the binding-specific amino acids is important for engineering and design applications. We generated a dataset of metal binding sites in enzymes using MetalPDB \citep{Putignano2018-qb}, a curated dataset derived from the Protein Data Bank (PDB) \citep{Berman2002-hh}. Focusing on zinc, the most common metal in the PDB, we generate a gold standard dataset of 610 zinc-binding (class$_1$) enzymes and 653 non-binding (class$_0$) enzymes (see \cref{supp:metalpdb}). Each protein structure is defined using the positions of its atoms in 3D space and subdivided into tokens representing amino acids (a.k.a. “residues”). The relationship between residues is represented as a graph with edges defined by inter-atomic distance (\cref{fig:unstruct}). This task is particularly challenging due to potentially overlapping class-specific features (\ie proteins of both classes are metal-binders), highly heterogeneous background data (proteins in train and test sets adopt a wide variety of structural folds), and relatively small target regions, making this an example of a \say{needle-in-the-haystack} task \citep{Pawlowski2019-by}. 

\underline{Encoders \& baselines:} We apply prospector heads to three encoders: COLLAPSE, an FM which produces embeddings of the local 3D structure surrounding each residue \citep{Derry2023-pg}; ESM2, a protein LLM which produces embeddings for each residue based on 1D sequence \citep{Lin2023-gy}; and a simple amino acid encoder (AA), where each residue is one-hot encoded by amino acid identity. By construction, ESM2 is a full-context encoder while COLLAPSE and AA are partial-context encoders. We present three baselines built on top of a supervised GAT \citep{Velickovic2017-oy} classifier head (trained on protein-level labels) to identify binding residues: Attention, Shapley values (SHAP), and GNNExplainer \citep{Ying2019-cu}. Implementation details are listed in \cref{supp:baseline}.


\begin{figure}
    \centering
    \vspace{0pt}
    \includegraphics[width=0.48\textwidth, trim={16em 19.5em 14.5em 0},clip]{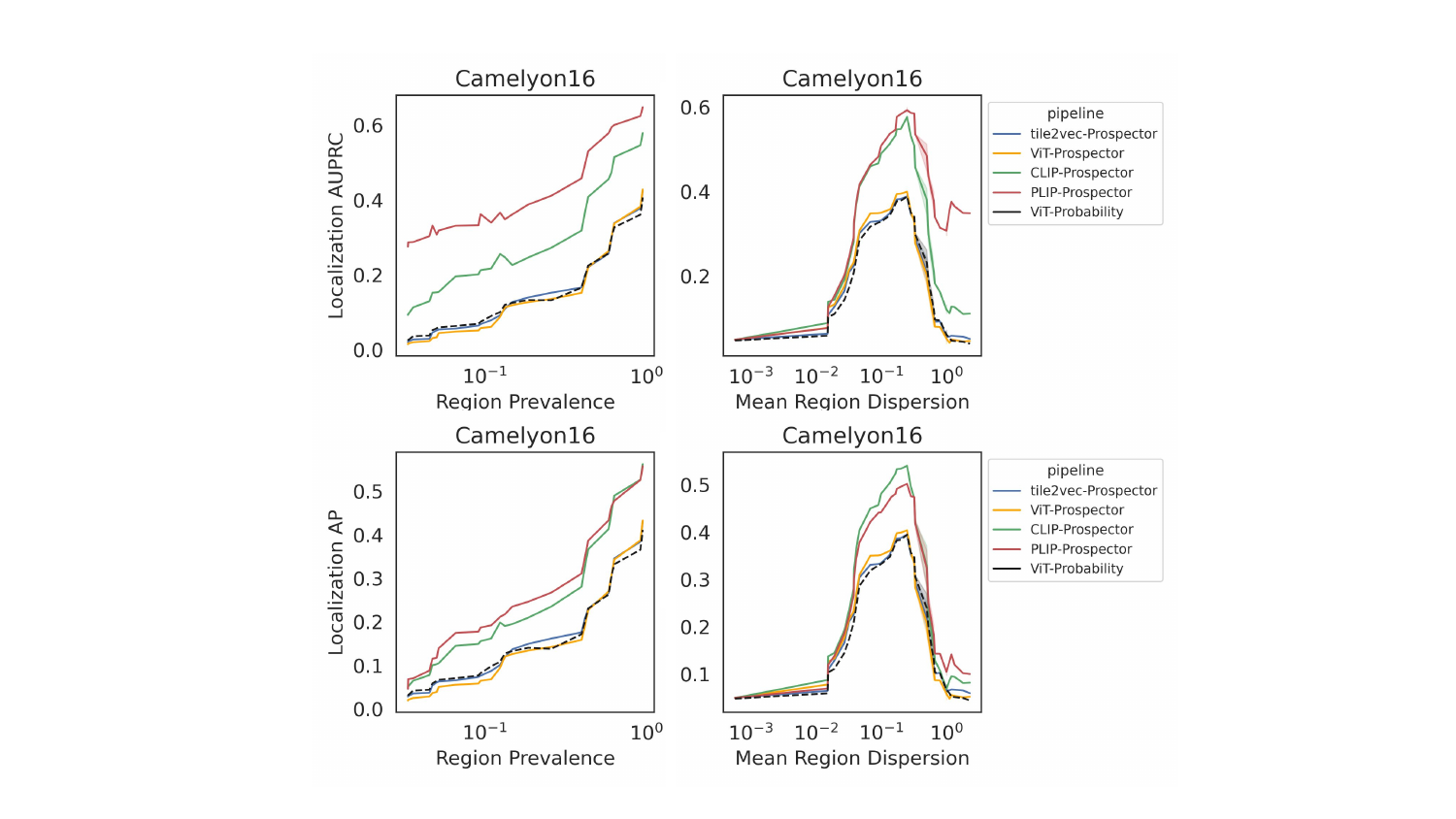}
    \vspace{-9pt}
    \setlength{\belowcaptionskip}{-15pt}
    \caption{Robustness analysis for Camelyon16 data: prospector and top baseline performance with respect to region characteristics.}
    \label{fig:prop-cam-preview}
\end{figure}


%% file: sections/4-results.tex

\subsection{Results}
\begin{figure*}
    \centering
    \vspace{0pt}
    \includegraphics[width=0.82\textwidth, trim={0 17em 0 0},clip]{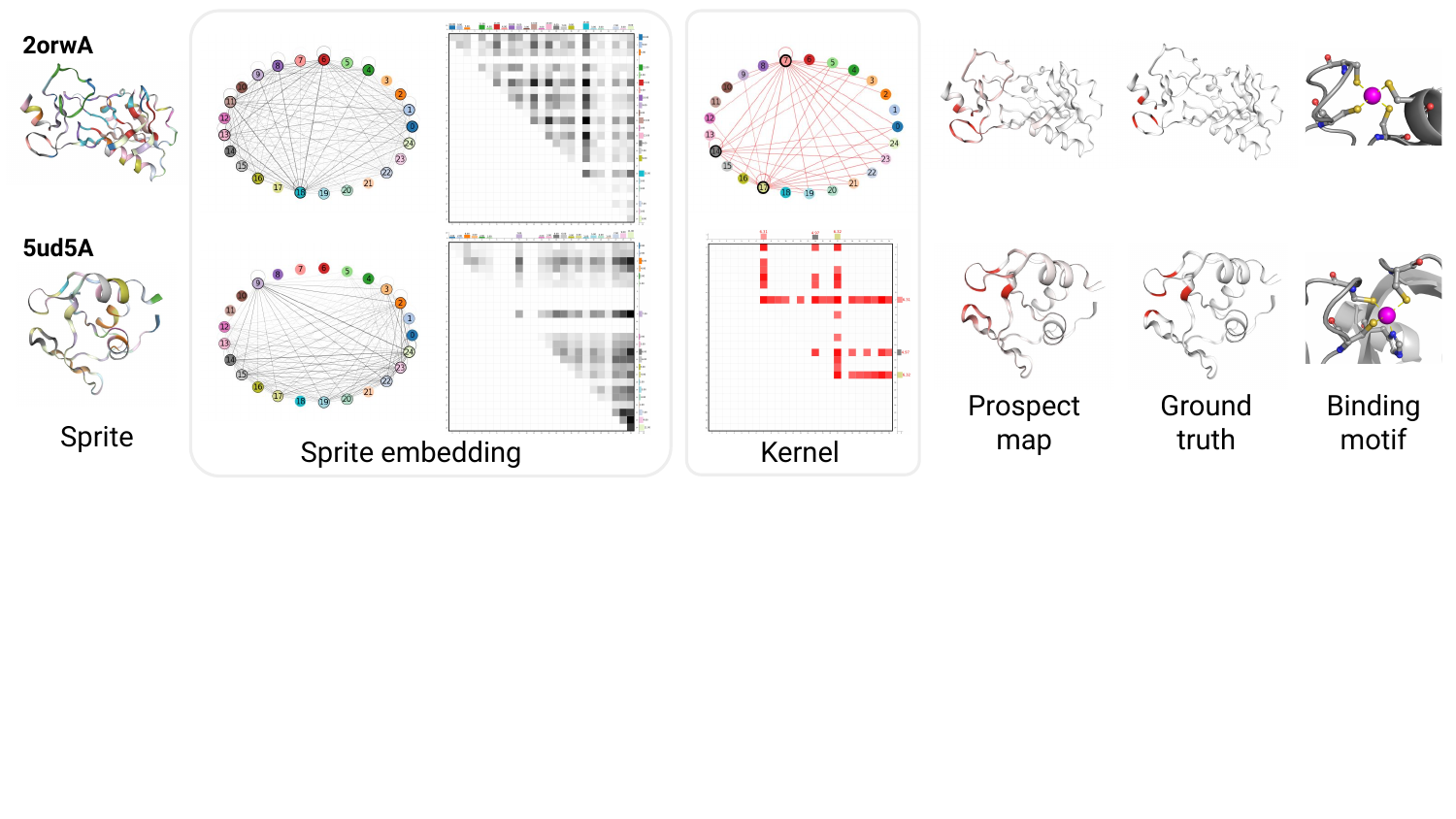}
    \vspace{5pt}
    \setlength{\belowcaptionskip}{-10pt}
    \caption{Prospector visualization for protein data, using the top COLLAPSE configuration (\cref{tab:topmodel}). We show the same five visualizations as before, as well as a visualization of the atomic configuration of the zinc binding site to illustrate the binding motifs discovered by sprite embedding clustering (\cref{fig:supp-clust}). Whole proteins are visualized as cartoons instead of graphs for clarity.}
    \label{fig:prot}
\end{figure*}

\textbf{Prospectors outperform baseline attribution methods in region localization and generalize across data modalities.} In all tasks, prospectors achieve higher AUPRC and AP than baseline methods, often with large improvements (\cref{fig:perf}). For text retrieval, we improve mean test-set AUPRC to 0.711 from 0.626 (\ie 8.5-point gain) with the top supervised baseline (MLP head) and from 0.584 with the top LLM inference baseline (NLI entailment) — in summary, MiniLM with an equipped prospector head is able to outperform DeBERTa's baselines by 12.7 points in AUPRC despite being 5$\times$ smaller in size and with relatively limited pretraining (\cref{tab:results-wiki}). We also observe improved localization over baselines for Camelyon16 (26.3 points in AUPRC and 8.8 points in AP) and MetalPDB (22.0 points in AUPRC and 8.8 points in AP). For the MetalPDB dataset, the optimal methods tend to exhibit bimodal performance, with almost perfect predictions for a subset of the test dataset (particularly cysteine-dependent binding motifs, see \cref{fig:prot}) and poor performance on other subsets, resulting in the clustering of points around 0.5 and 1.0 AUPRC. This behavior suggests that AP more clearly reflects task performance, highlighting the ability of prospectors to identify small conserved binding patterns.





\textbf{The choice of encoder is key to optimal prospector performance.} While prospectors overall improve localization performance over baselines regardless of the chosen encoder, the performance gain is maximized by choosing domain-specialized encoders for each dataset. For Camelyon16 and MetalPDB, the combination of prospectors with FM encoders (CLIP, PLIP, COLLAPSE) showed the strongest localization results, as shown in \cref{fig:perf}. Among FMs, the best-performing encoders are those with the most task-specificity — PLIP has a domain advantage by virtue of being a CLIP-style encoder trained on pathology images, while COLLAPSE accounts for complex 3D atomic geometry rather than simply amino acid sequence (as in ESM2) or one-hot encoding (AA). Interestingly, we note that the AA encoder presents an exception to encoder generalization, supporting that prospectors themselves can identify salient patterns with rudimentary encoder semantics. This is likely due to the fact that many zinc-binding motifs rely on atomic coordination by three to four cysteine residues, which are otherwise rarely found in such arrangements. For tasks which require the detection of less amino acid-dependent structural patterns, we expect the COLLAPSE encoder to result in optimal prospector performance.



\textbf{Prospectors are robust to coarse supervision.} Next, we explore the relationship between the properties of class-specific regions and localization performance. To characterize class-specific regions, we compute two metrics acting as proxies for coarse supervision (\cref{sec:prelims}): \textit{region prevalence} (\# class$_1$ tokens / \# tokens) and \textit{mean region dispersion} (\# connected components / mean component size). For Camelyon16, we plot the relationship between test-set AUPRC and both metrics in \cref{fig:prop-cam-preview}. Full results over all datasets are presented in \cref{supp:robustness}. For each plot, we also display the top baseline method.

Firstly, we observe that most encoders exhibit a positive correlation between region prevalence and localization AUPRC across all modalities. However, some encoders are particularly robust to region prevalance and achieve high AURPC despite low prevalence (MiniLM, PLIP, COLLAPSE), and prospectors are consistently more robust than top baselines over all data modalities. Secondly, mean region dispersion and localization performance (both AUPRC and AP) demonstrate a \textit{parabolic} relationship — indicating that some level of dispersion is needed for detectable regions, while too much dispersion makes the task challenging. These results recapitulate each task's challenges: the pathology task contains a wide range of dispersion values, while the protein task contains the lowest levels of prevalence and highest levels of dispersion (\cref{supp:props}). Despite these task differences, prospector-equipped FMs demonstrate an high levels of robustness to coarse supervision across modalities.


\begin{figure*}
    \centering
    \vspace{0pt}
    \includegraphics[width=0.75\textwidth, trim={0 13em 0 0.5em},clip]{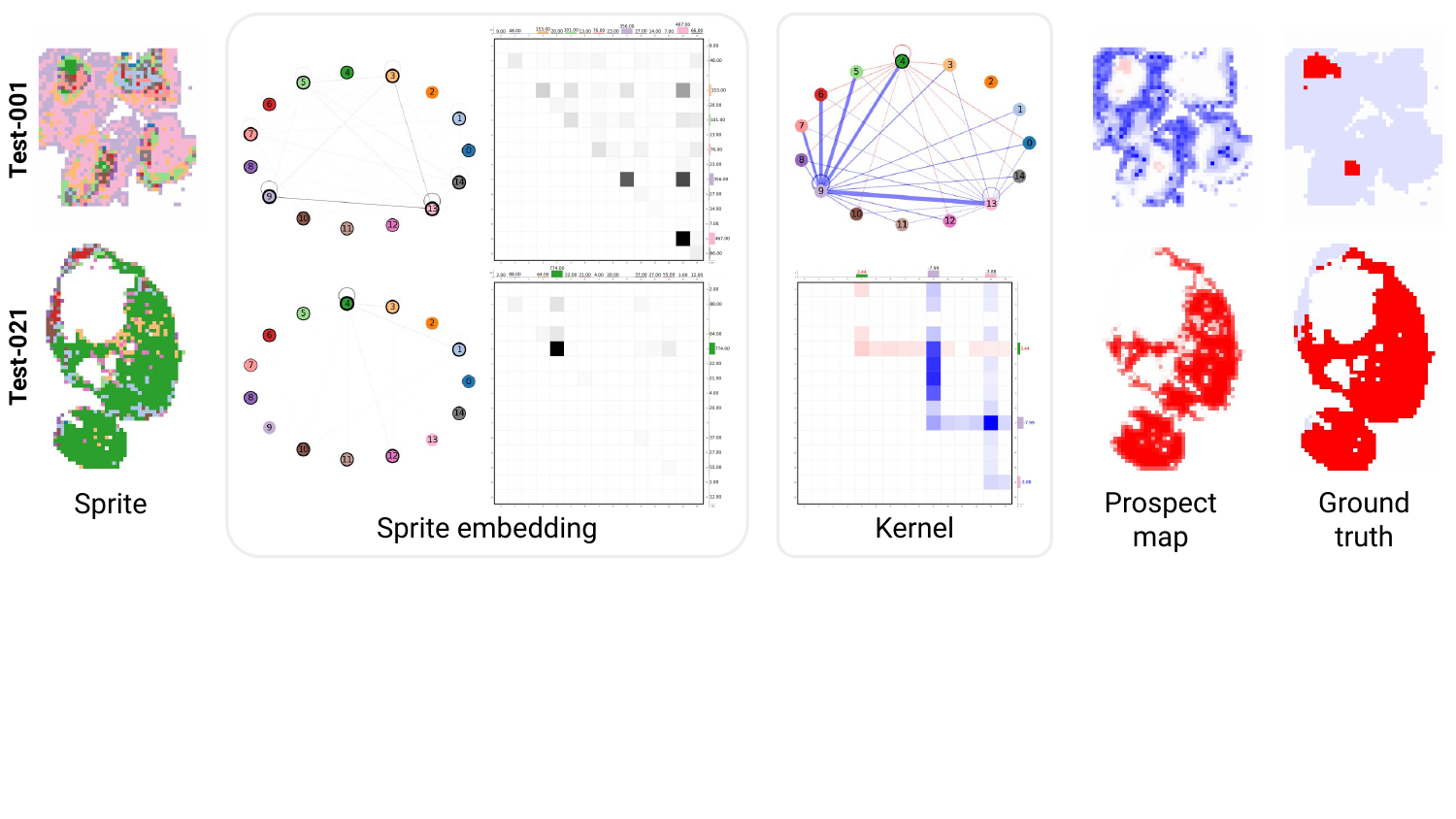}
    \vspace{5pt}
    \setlength{\belowcaptionskip}{-10pt}
    \caption{Prospector visualization for pathology, using the top PLIP configuration (\cref{tab:topmodel}). Visualizations are shown for two test-set examples, from left to right: data sprites; sprite embeddings viewed as semantic networks and heatmaps (\cref{supp:viz}), where line thicknesses or cell shade reflect monogram or skip-bigram count; the kernel viewed as a semantic network and heatmap, where line thickness and cell intensity reflect learned weights; prospect map, with vertex attribution scores mapped back onto tokens in original data; and ground-truth class-specific regions in the image (in red). Sprites and sprite embeddings are colored by the $K=15$ learned concepts. Kernel weights and prospect maps are colored red and blue to reflect class$_1$-specific and class$_0$-specific associations, respectively.}
    \label{fig:path}
\end{figure*}


%



\textbf{Prospectors' sprite embeddings and kernels are interpretable and enable internal visualization.} In addition to improved localization performance, prospectors are inherently \textit{interpretable} because their parameters provide insights into invariant class-specific patterns. Prospect maps visualize the feature attribution outputs in the input token space — but importantly, these maps can be further contextualized by visualizing prospector internals themselves. Due to the use of learned semantic concepts, the global convolutional kernel can be represented as a semantic network or as a heatmap  (\cref{supp:viz}), along with each input example as it passes through layers of the prospector head.

We illustrate this interpretability for pathology images (\cref{fig:path}) and protein structure (\cref{fig:prot}) using two test-set examples. We first visualize data sprites, which reflect the learned concepts mapped onto data inputs (from layer (I)). By analyzing semantic concepts on the data sprite, it is possible to assign domain-specific meaning to each concept. Additionally, by visualizing concept and co-occurrence frequencies in the sprite embedding, we can identify over- or under-represented patterns within each input. By visualizing the global kernel, which captures dataset-wide concept associations and their correlations with class labels, it is possible to cross-reference between the sprite and the class-specific regions of the resulting prospect map. The ability to visualize the internals of a prospector head in terms of concepts facilitates human-in-the-loop model development and the incorporation of domain knowledge, a major advantage relative to \say{black box} models.

\textbf{Prospector kernels allow for parsimony to find \say{hub} concepts.} Our pathology visualization (\cref{fig:path}) demonstrates a kernel with \say{hub,} or densely connected and highly predictive concepts: concept $\#4$ is indicative of class$_1$ while concept $\#9$ is indicative of class$_0$. Such kernels demonstrate how prospectors \textit{do not detract} from the rich semantics offered by FM encoders like PLIP for pathology data.

\textbf{Prospection is robust to concept distributional shifts.} Visualizing kernels for protein structures outlines prevalent class$_1$-specific concepts in training data (\eg concepts $\#7$, $\#17$) that are rare in the test set but nonetheless are critical for classification. Despite their low frequency, top prospectors achieved performant localization for this task. The distributional shift between train and test set is a likely explanation for the bimodal localization performance on this task, and suggests that improvements to kernel design and fitting (\eg feature scaling and choice of $K$) along with constructing optimally representative training datasets (\eg for a more varied class$_0$) would improve prospector performance on more difficult data subsets. 

\textbf{Sprite embeddings also carry class signal.} Further analysis of learned parameters can also help to better understand the nature of discovered patterns. For example, there may be more than one pattern which results in a particular class label, and differentiating examples that exhibit each pattern can uncover mechanistic subgroups of the data. To demonstrate this, we hierarchically cluster the sample-level sprite embeddings in the MetalPDB test set. This identified two major subgroups of zinc binding sites (\cref{fig:supp-clust}) defined by the number of cysteine residues coordinating the bound ion. One subgroup is enriched for proteins which contain four coordinating cysteines, while in the other there are one or more histidine residues involved in the binding interaction. \cref{fig:prot} shows an example from each cluster, including a visualization of the zinc-binding site on the far right. This finding recapitulates known subtypes of zinc binding motifs \citep{Wu2010-gd}, and more broadly demonstrates the potential for prospectors to discover biological mechanisms when applied to less well-studied phenomena.

%% file: sections/5-conc.tex
This work presents prospector heads, encoder-equippable modules for (a) data-efficient, (b) time-efficient, and (c) performant feature attribution. We show that prospectors are both modality-generalizable and encoder-agnostic with particularly dominant performance when equipped to domain-specialized FMs. Finally, we show that prospectors are interpretable through their use of concept-based kernels.

Prospectors' improved localization performance over explanation-based baselines calls into question the underlying assumption of explanations themselves: that end-to-end classifiers implicitly \say{segment} data in the input token space en route to making class predictions, and that these \say{segmentations} can be extracted \textit{post hoc}. Our results suggest that using machine-derived concepts and modeling class-specific associations directly in the input token space helps to avoid such modeling assumptions.

We believe a key driver of prospectors' performance is the combination of token-level representations with the local inductive bias provided by convolution. This combination fosters a form of inductive reasoning through \say{token mixing} and kernel construction. Several other aspects of prospector design draw inspiration from ideas across ML research (\cref{sec:connections}), giving insights into their performance characteristics. Our results suggest that FMs in particular contain strong distributional semantics which yield precise feature attribution even with partial-context encoders and coarse levels of supervision. In other words, FMs (in tandem with quantization) remove the burden of long-context reasoning by reducing input data to mosaics of concepts (\ie sprites). Prospectors can thus functionally operate over long-range dependencies even with a local inductive bias. This claim of capturing short- and long-range dependencies between tokens is backed by prospectors' localization robustness to region prevalence and dispersion. Additionally, because domain-specific FMs do improve performance when they are available (\eg PLIP vs. CLIP), we hypothesize that as FMs continue to improve and be adapted to new applications and data modalities, so will the utility of prospectors across diverse domains.

Prospectors are flexible and modular by design, enabling not only variable encoders but also simple changes in their fitting. Of the two variants we fitted, the non-trainable fold-change variant was superior for almost all evaluated settings (\cref{supp:model-select}). This may be because the variant explicitly learns dataset-wide concept associations and deviations from a class$_0$ \say{baseline vector} \citep{Sundararajan2017-ps,Bilodeau2022-lm,Afchar2021-xb} — which closely reflects the MIA (\cref{sec:datasets}). It is possible that different kernel fitting methods may be better suited to detecting different types of class-specific patterns, but further investigation is needed to explore this question.

One limitation of this work is the lack of sensitivity analysis for all design choices and hyperparameters. For example, due to time and compute constraints, we relied on domain knowledge to select token resolution and connectivity for each task instead of testing their impacts on performance. Furthermore, we did not study the choice of clustering method nor embedding sample size in the quantization step, and we limited our experimentation to open-source encoders only. Future work involves \textit{Pytorch} implementation for GPU acceleration, enabling kernels to learn higher-order $n$-grams, adding new variants for kernel fitting, deployments on varied data modalities, and exploring prospector utility with frontier non-Transformer architectures (\eg state-space models \citep{Gu2021-bq,Poli2023-uw} and their attention hybrids \citep{Poli2023-al}) and API-locked LLMs \citep{Bommasani2023-kb}. 

We anticipate many potential use cases for prospectors, particularly in tandem with vector databases and in other compound AI systems and agentic workflows \citep{Zaharia_undated-wd}. One particular use case is to screen or classify data with FMs equipped with performant classifier heads \citep{Swanson2022-on}, and then swap in prospector heads when feature attribution is required. This process can enable users to investigate multiple class labels (\eg scientific phenomena) without encoder retraining. Another use case is to use prospector-generated attributions to train downstream rationale models \citep{Jain2019-to, Chen2022-qd, Yang2023-oo, Bujel2023-qa}. In general, we believe that prospectors expands the toolkit for improving the transparency and utility of large FMs, high-dimensional data, and large-scale datasets — ultimately inspiring new few-shot inference modes for FMs. For scientific and biomedical applications, including in data-scarce settings, prospectors have the potential to provide mechanistic insights and discover phenomena in complex data \citep{Wang2023-gp}.

%% file: sections/appendix.tex


\section{Related Work (Extended)}
\label{supp:rw}

\subsection{Feature Attribution via Explanation} 

In the current explanation-based paradigm, feature attributions are referred to as \say{explanations} and are performed by (1) training a supervised model before (2) interrogating the model's behavior (\eg via internals, forward or backward passes, or input perturbations) and inferring class-specific features. This framework can be described as \textit{weak} or \textit{coarse supervision} \citep{Robinson2020-ef} due to the sole use of class labels as a supervisory signal in combination with a low signal-to-noise ratio in the datum-label pairs — particularly when the prevalence of class-specific features is low \citep{Pawlowski2019-by}. 

Explanations, and feature attribution more broadly, can be categorized as either \textit{model-specific methods}, which aim to describe how model weights interact with different input features), or \textit{model-agnostic methods}, which aim to describe how each feature contributes to prediction. Explanation-based attribution methods in general are inherently data-inefficient as they require ample labeled training data to train underlying classifiers. It should be noted that few methods can also be applied to all data modalities.

Model-specific methods like gradient-based saliency maps \citep{Simonyan2014-ta}, class-activation maps (CAMs) \citep{Zhou2016-wy, Selvaraju2016-so, Wang2023-gp}, and attention maps \citep{Jetley2018-aw} use a classifier’s internals (\eg weights, layer outputs), forward passes, and/or backpropagated gradients for attribution. Recent work has demonstrated gradients serve as poor localizers \citep{Arun2020-jp, Zech2018-vb} — potentially due to their high sensitivity to inputs \citep{Adebayo2018-ls}, unfaithfulness in reflecting classifiers' reasoning processes \citep{Karimi2022-ek} and propensity to identify spurious correlations despite classifier non-reliance \citep{Adebayo2022-fz}. Furthermore, CAMs pose high computational costs with multiple forward and backward passes \citep{Chen2023-ak}. Finally, attention maps are demonstrably poor localizers \citep{Zhou2021-lb} also perhaps due to their unfaithfulness \citep{Wiegreffe2019-my} and difficulties in assigning class membership to input features \citep{Jain2019-to}. 

On the other hand, model-agnostic methods like SHAP \citep{Lundberg2017-bw} and LIME \citep{Ribeiro2016-hq} perturb input features to determine their differential contribution to classification. Recent work has shown SHAP struggles to localize class-specific regions and is provably no better than random guessing for inferring model behavior or for downstream tasks \citep{Bilodeau2022-lm}. Furthermore, SHAP-style methods can be computationally expensive for a variety of reasons. Some methods face exponential or quadratic time complexities \citep{Ancona2019-kg} with respect to the number of input features (\eg pixels in an image) and are thus infeasible for high-dimensional data, while others require multiple forward and/or backward passes \citep{Chen2022-ej} or require training additional comparably sized deep networks along with the original classifier \citep{Jethani2021-yx}. 




\subsection{Modern Encoders \& Context Sizes}

Most modern encoders for unstructured data operate on \textbf{tokens}, or relatively small pieces of a datum, and their representations. Tokens can be user-prespecified and/or constructed by the encoder itself (potentially with help from a tokenizer) — where these encoders are respectively referred to as \textit{partial-context} and \textit{full-context} (\cref{fig:modality}). Due to computational constraints, high-dimensional unstructured data (\eg gigapixel images) often require user-prespecified tokens (\ie patches) and partial-context encoders that embed each token \citep{Lu2023-ar, Huang2023-iv, Klemmer2023-oo, Lanusse2023-xt}. 

We provide an illustrative example for the image modality. In this setting, determining encoder context is based on practical modeling constraints: computational complexity of an architecture's modeling primitives, input data dimensionality, and hardware. For example, an attention-based Vision Transformer \citep{Dosovitskiy2020-wn} experiences quadratic time complexity \citep{Keles2022-cw} with respect to input dimension. Standard images (\eg $224 \times 224$ pixels) easily fit in modern GPU memory, enabling us to train full-context encoders that construct token embeddings via intermediary layers. However, gigapixel images require user-prespecified tokens (\ie patches) and partial-context encoders.

Full-context encoders now include foundation models (FMs) for a variety of data modalities and domains, including natural imagery \citep{Radford2021-ln}, radiology images \citep{Zhang2023-kp, Zhang2023-ws,Singhal2023-ae, Saab2024-fw}, pathology images \citep{Xu2024-gs}, protein sequences \citep{Rives2021-ck, Lin2023-gy}, and molecular graphs like protein structures \citep{Derry2023-pg}. On the other hand, partial-context encoders train on prespecified tokens like document sentences and image patches. In image domains like histopathology, remote sensing, and cosmology, numerous encoders have been proposed with varying training regimes: unsupervised encoders \citep{Jean2019-np}, weakly supervised classifiers \citep{Chen2022-rv, Campanella2019-sj}, and FMs \citep{Klemmer2023-oo, Jakubik2023-vj, Zhang2023-my, Huang2023-iv, Lu2023-ar, Lanusse2023-xt} all build representations for patches.

Regarding feature attribution for partial context models, gradient-based saliency and attention maps have been used to explain class predictions for high-dimensional unstructured data like gigapixel imagery \citep{Campanella2019-sj, Chen2022-rv}. However, studies report low specificity and sensitivity \citep{Machiraju2022-yd} in part because attribution for the entire datum is built by concatenating attributions across prespecified tokens (\eg image patches). Partial-context strategies incorrectly assume prespecified tokens are independent and identically distributed (IID).

Our work hinges on the assumption that FMs learn particularly rich embeddings and distributional semantics — and thus, sets of concepts — by virtue of their representational power. While feature attribution has not been explored by adapting FM embeddings, this work is inspired by the recent efforts to perform object detection and visual grounding via FM adaptation \citep{Kuo2022-qt,Kalibhat2023-sr}.

\subsection{Broader Connections across ML}
\label{sec:connections}
Prospectors bring together ideas from many classical and modern works in adaptation, interpretability, memory augmentation \citep{Khosla2023-nv}, information retrieval, and language modeling. On the surface, prospectors resemble probing models \citep{Alain2016-xa,Belinkov2021-mo}, but the fact that they learn token associations between \textit{multiple} token embeddings is more akin to constellation models \citep{Weber2000-ah}, self-attention layers \citep{Vaswani2017-ms}, or multiple instance learning approaches \citep{Javed2022-hj}. Layer (I) is inspired by concept bottlenecks (\Cref{sec:related}), but extends the definition of concepts to carry spatial semantics. To learn higher-order associations between concepts, \ie \say{token mixing} and inductive reasoning, layer (II) is inspired by both sliding window attention \citep{Parmar2018-jx, Child2019-jc, Beltagy2020-zj} and the emergent $n$-gram circuits seen in transformer induction heads \citep{Akyurek2024-rq, Olsson2022-fj, Bietti2023-wd}. We foster the pattern-recognition capability via associative memory units \citep{Hopfield1982-tu, Hopfield1984-yp, Kohonen1972-en, Ramsauer2020-dc} built with an encoder's learned representations and graphical models \citep{Liu2018-ad, Graves2013-lo}. The result is that while prospectors are inspired by LLM reasoning, their implementation uses efficient statistical techniques, modeling primitives, and data structures.

\section{Out-of-scope Attribution Methods}
\label{supp:oos-baselines}
For transparency, we also outline our choice to rule out certain baselines for our experiments. A top priority for baseline selection was modality generalizability.

\textbf{LIME:} we rule out LIME \citep{Ribeiro2016-hq} as a baseline for any of our tested data modalities. This is primarily due to the fact that LIME requires ground truth labels to explain each input. Since the inputs to our partial context encoders (and, in turn, LIME) are prespecified tokens (\eg sentences for the WikiSection task), LIME requires token-level labels to explain the importance of sub-tokens (\eg words). This requirement of token-level labels in our setting is fundamentally at odds with prospectors’ goal to predict token labels, \ie learn class-specific tokens \textit{de novo}.



\textbf{FastSHAP:} we do not compare prospectors to modern methods like FastSHAP \citep{Jethani2021-yx}, which requires training additional models. FastSHAP specifically requires training two comparable models to the original encoder (\ie with a classifier head) with respect to parameter count: a \say{surrogate} model that typically mimics the encoder in architecture but trained with a masked-input training regime and an \say{explainer} model that learns to identify class-specific tokens. Such approaches are out of scope for this work, which aims to perform feature attribution with large models like FMs. Training surrogates for FMs is often practically infeasible.

\section{Prospector Heads}
\label{supp:prospect}

\subsection{Core Definitions}
\label{supp:definitions}

\begin{figure}[h]
    \centering
    \vspace{0pt}
    \includegraphics[width=0.45\textwidth, trim={0 0 0 0},clip]{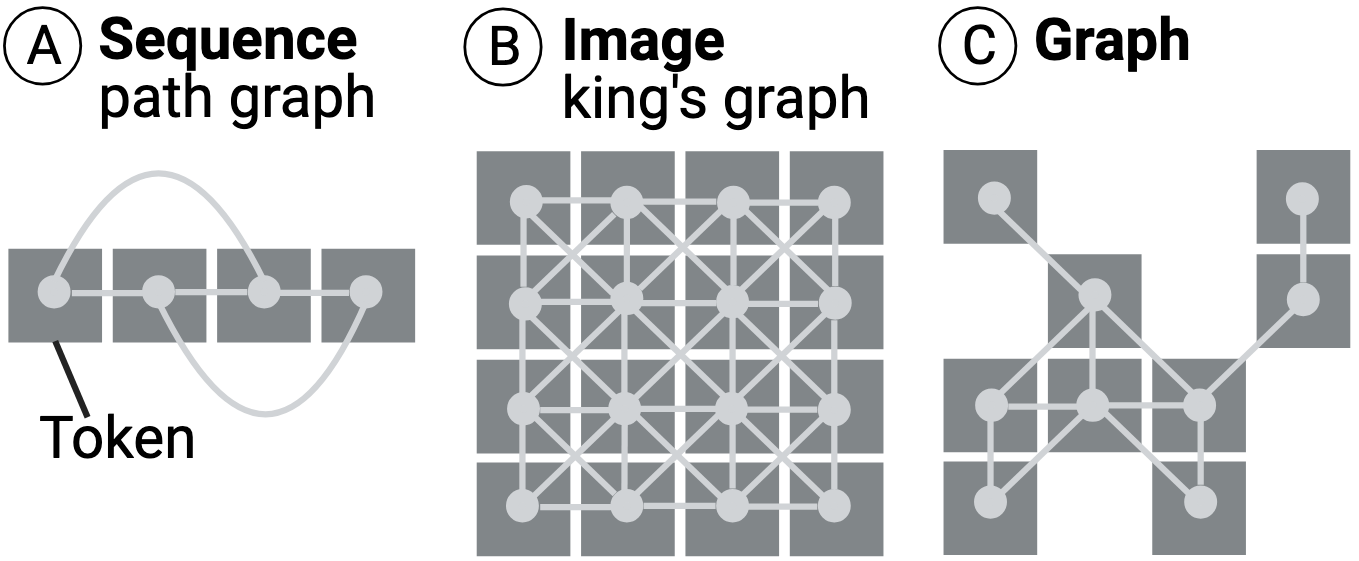}
    \vspace{0pt}
    \setlength{\belowcaptionskip}{0pt}
    \parbox{0.7\textwidth}{ \vspace{5pt} \caption{Unstructured data represented as map graphs. Sequences (A) and images (B) require a specified resolution (\eg words or sentences for text, pixels or patches for images) and connectivity (\eg 2-hop, 8-way) for discretization.} \label{fig:unstruct}}
\end{figure}

All unstructured data can be represented as map graphs of tokens interacting in physical space. We introduce mathematical definitions to describe these representations. Map graphs are also depicted in \cref{fig:unstruct}.

\begin{definition}[Map Graph]
\label{def:map}
\normalfont A map graph $G(\mathcal{V},\mathcal{E})$ is a collection of vertices $\mathcal{V}$ and edges $\mathcal{E}$ connecting neighboring vertices in Euclidean space. Each vertex $v^{(i)} \in \mathcal{V}$ has features $\token^{(i)}$ and each edge $e^{(i \leftrightarrow j)} \in \mathcal{E}$ connects vertices $v^{(i)}$ to $v^{(j)}$.
\end{definition}

\begin{definition}[Connectivity]
\label{def:map}
\normalfont A map graph $G$'s connectivity $\delta$ is its maximal node degree.
\end{definition}

\begin{definition}[Partial-context Encoder]
\label{def:pce}
\normalfont Given a map graph $G$, an encoder $f$ is considered partial-context if it produces an embedding $\token = f(v) \in \mathbb{R}^d$.
\end{definition}

\begin{definition}[Full-context Encoder]
\label{def:fce}
\normalfont Given a map graph $G$, an encoder $f$ is considered full-context if it produces embeddings $[\token_1 \ldots \token_T] = f(G)$, where $\token_i \in \mathbb{R}^d \quad \forall i = 1\ldots T$.
\end{definition}

\subsection{Visualizing Prospectors}
\label{supp:viz}
We choose to visualize any dictionaries created by prospectors (\eg kernel $\omega$ and during \texttt{rollup} \cref{supp:layerii}) in two main styles throughout this work. Firstly, visualization can take the form of (1) \textit{semantic networks}, which easily allow us to visualize either frequencies (in sprites) or importance weights (in kernels) for monogram and skip-bigram associations. These plots are sometimes referred to as \say{chord diagrams} or \say{circos plots}. This data structure is defined mathematically as a self-complete graph:

\begin{definition}[Self-complete graph]
\label{def:self-complete}
\normalfont A self-complete graph $K_K(\mathcal{V},\mathcal{E})$ is a fully connected graph with $K$ vertices, where every pair of distinct vertices $v^{(i)},v^{(j)}$ is connected by a unique edge $e^{(i \leftrightarrow j)}$. It also contains all self-edges that connect any vertex $v^{(i)}$ to itself with edge $e^{(i \leftrightarrow i)}$. Thus, self-complete graphs contain $K$ vertices and $K + {K\choose 2}$ edges. 
\end{definition}

This data structure is referenced in figures \ref{fig:layeri}, \ref{fig:layerii}, \ref{fig:path}, and \ref{fig:prot}. Additionally, we can visualize all associations as (2) \textit{heatmaps}, or unordered symmetric arrays, as seen in figures \ref{fig:layerii}, \ref{fig:path}, and \ref{fig:prot}.


\subsection{Prospector Internals \& Fitting}
\label{supp:internals}

\subsubsection{Layer II}
\label{supp:layerii}
\begin{figure}[h]
        \vspace{0pt}
        \centering
        \includegraphics[width=0.6\textwidth, trim={0 0 0 0},clip]{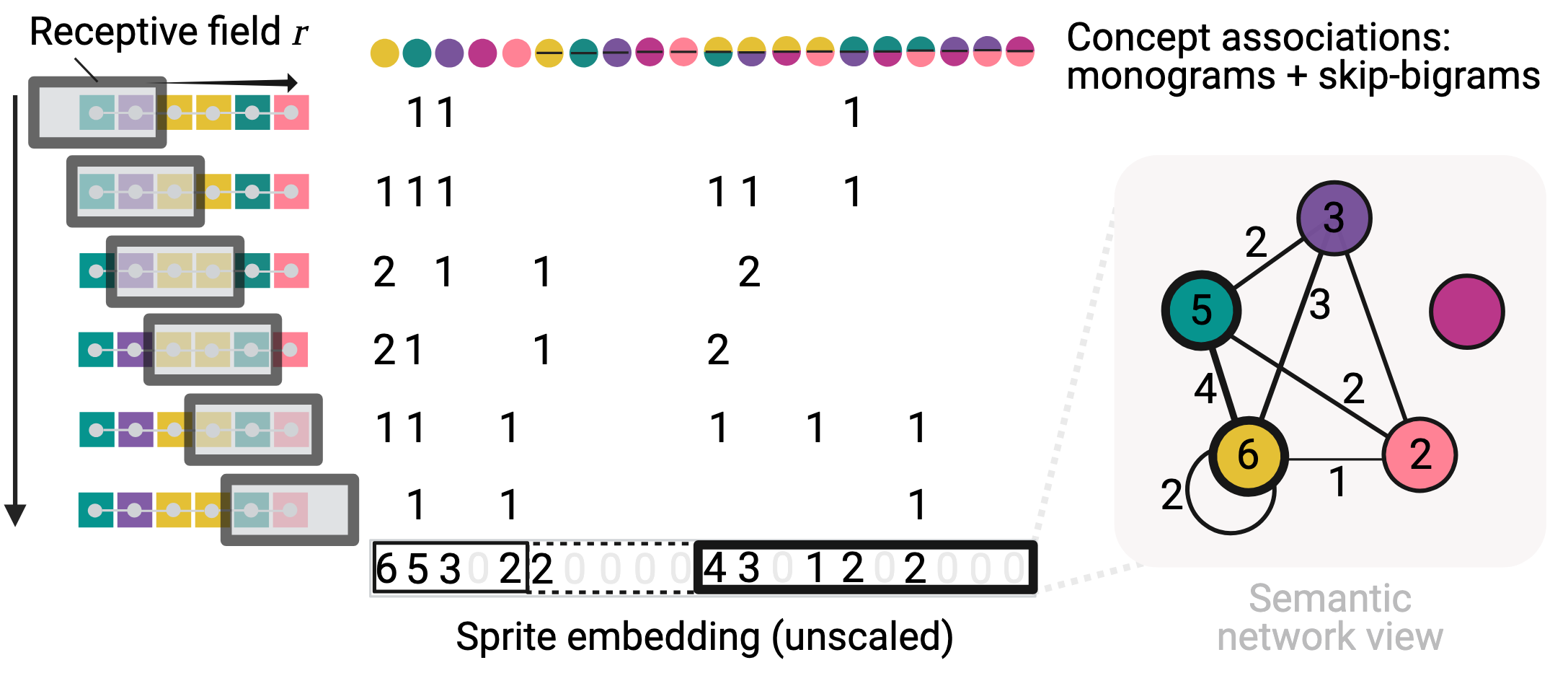}
        \vspace{3pt}
        \setlength{\belowcaptionskip}{-7pt}
        \caption{A depiction of the \texttt{rollup} operation as seen in layer (II).}
        \label{fig:rollup}
\end{figure}

The \texttt{rollup} operator, named after the function of the same name in relational databases, draws similarity to a sliding \textit{bag of words} featurization scheme. Internally, a dictionary $\zeta$ is constructed to capture all monograms and skip-bigrams in each neighborhood of $S$. This operator is described by \cref{alg:rollup} and depicted in \cref{fig:rollup}. For a full view of fitting layer (II), including both steps 1 (\texttt{rollup}) and 2, refer to \cref{fig:layerii}. We note that all sprite embeddings created in \texttt{rollup} were normalized using TF-IDF scaling \citep{Sparck_Jones1972-ke} prior to kernel fitting. 
\begin{algorithm}[h]
    \caption{\texttt{rollup}}
    \label{alg:rollup}
    \footnotesize
    \begin{algorithmic}[1]
        \REQUIRE Sprite $S$, receptive field $r$
        \STATE initialize dictionary $\zeta$
        \FOR{vertex $v_i$ in $S$}
            \FOR{vertex $v_j$ neighborhood $\mathcal{N}_r$}
                \STATE $\zeta \langle S[v_i] \rangle \leftarrow$ occurrences of concept monogram $S[v_i]$ 
                \STATE $\zeta \langle (S[v_i],S[v_j]) \rangle \leftarrow$ occurrences of concept skip-bigram $(S[v_i],S[v_j])$
            \ENDFOR
        \ENDFOR
        \STATE \textbf{return} $\mathbf{z} = \operatorname{linearize}(\zeta)$ \\
        \COMMENT{where $\operatorname{linearize}$ returns values in sorted order}
    \end{algorithmic}
\end{algorithm}

\subsubsection{Parameterization}
\label{supp:maxz}
Prospectors can have the following maximum number of importance weights, depending on $r$ and $K$:
\[ |\mathcal{Z}| = \begin{cases} 
      K & r = 0 \quad \text{(monograms only)} \\
      2K + {K\choose2} & r > 0 \quad \text{(monograms \& bigrams)}
   \end{cases}
\]

\subsubsection{Prospector Variants: Implementation}
\label{supp:variants}
As depicted in \cref{fig:layerii}, and discussed in the main body of this work, we implement two variants of prospector heads: a linear classifier variant and a fold-change variant. We provide additional details here for both variants. Details on hyperparameter selection $\lambda,\tau,\alpha$ are discussed in \cref{supp:hparam}.

\textbf{Linear classifier:} This variant was implemented with the \textit{sklearn} python package. The elastic net classifiers ($\lambda=0.5$) trained for a maximum of 3000 iterations using the saga solver. 

\textbf{Fold-change computation:} In order to supply an alternative to regularization for fold-change variants, we use two-way thresholding as inspired by differential expression analysis \citep{Anders2010-zg}. These thresholds offer a form of \say{masking} importance weights $w_i \in \mathbf{w}$. As described in the main body of this work, the first threshold is $\tau$, or the minimum fold-change required. The other threshold is $\alpha$, which is a threshold used for a statistical hypothesis test, which is tests for independent class means. This test is conducted for each weight entry $w_i$ in $\mathbf{w}$ and significance is assessed via a Mann-Whitney U hypothesis test. Prior to weight masking, given the number of independent tests being conducted, we adjust our chosen significance threshold using the commonly used Bonferroni correction: our original $\alpha$ threshold is divided by the number of entries in $\mathbf{w}$ ($|\mathcal{Z}|$), \ie $\alpha^* := \alpha / |\mathcal{Z}|$. Finally, to perform masking: we use $\pm\tau$ to mask out sufficiently small absolute fold changes (\eg $\pm$1, which indicates a requirement for doubling in $\log_2$-scale), and use $\alpha^*$ to mask out non-significant differences assessed by our hypothesis test.

\subsection{Inferential Time Complexity}
\label{supp:runtime}
Here we conduct a comparative runtime analysis, where we analyze the worst-case time complexity required to explain a single input datum. We focus our analysis on the image modality due to compatibility with many baseline attribution methods. Suppose we have a trained encoder (\eg an end-to-end classifier, unsupervised learner, etc.) and our datum has $T = |\mathcal{V}|$ (tunable) tokens to analyze. Importantly, full-context encoders process all $T$ tokens at once while partial-context encoders process $T$ tokens in sequence. This distinction affects runtime complexity, so we analyze complexity for both partial- and full-context settings. 

\subsubsection{Prospectors}
To analyze prospectors, we consider two main variables in computation: the number of tokens $T$ and the number of operations for a forward pass ($F$) of the underlying encoder. Given these variables, prospectors themselves require only $O(T)$ computations per layer at inference time: $O(T)$ to quantize each token and $O(T)$ to traverse over all tokens during convolution. The latter operation ignores a near-constant term for the worst-case number of interactions, \ie skip-bigrams between central token and tokens in the $r$-neighborhood. The worst-case number of interactions is modality- and user-specific and is dependent on $G$'s topology, $r$, and connectivity $\delta$ (\ie max node degree). Parameters $r$ and $\delta$ are both typically set as small constants, so we can consider them negligible for time complexity. 

Because prospectors are equipped to backbone encoders, inference in totality must account for the encoder’s computational costs as well. Namely, an encoder requires $O(F)$ for inference with full context and $O(TF)$ for partial context (since each prespecified token requires a forward pass). Thus, total computational complexity of an encoder-prospector pipeline operating on a single input datum is $O(TF)$ for partial-context encoders and $O(T+F)$ for full-context encoders.

\subsubsection{Baseline Methods}
To properly characterize baseline methods, we also consider variables for backward pass operations ($B$), sub-tokens ($T^*$), and number of passes ($p$) if applicable. Sub-tokens are any (tunable) constituents within a token (\eg pixels in a patch) and are required by some baselines in partial-context settings. For example, given a text document (datum) and its sentence-level tokens, SHAP \citep{Lundberg2017-bw} may analyze the contribution of word sub-tokens. For comparative summary between prospectors and multiple baselines, please refer to \cref{tab:runtime}. We discuss ruled-out baselines in \cref{supp:oos-baselines}.





\begin{table}
    \centering
    \begin{tabular}{ccc}
    \toprule
    Attribution method & Full-context & Partial-context  \\
    \midrule
    Gradients & $O(F+B)$ & $O(T(F+B))$ \\
    CAM \citep{Zhou2016-wy} & $O(pF)$ & $O(TpF)$ \\
    GradCAM \citep{Selvaraju2016-so} & $O(p F + B)$ & $O(T(p F + B))$ \\
    ScoreCAM \citep{Wang2019-tu} & $O(pF)$ & $O(TpF)$ \\
    Attention & $O(F)$ & $O(TF)$  \\
    SHAP \citep{Lundberg2017-bw} & $O(F 2^{T})$ & $O(T F 2^{T^*})$ \\
    DASP \citep{Ancona2019-kg} & $O(F T^{2})$ & $O(T F {T^*}^{2})$ \\
    G-DeepSHAP \citep{Chen2022-ej} & $O(p(F+B))$ & $O(Tp(F+B))$ \\
    \midrule
    Prospectors (ours) & $O(F+T)$ & $O(TF)$ \\
    \bottomrule
    \end{tabular}
    \parbox{0.7 \textwidth}{ \vspace{5pt}
    \caption{Comparison of inference-time computational complexity between attribution methods. Parameters $T$, $T^*$, $F$, $B$, and $p$ are respectively the number of: tokens, sub-tokens, operations in a forward pass, operations in a backward pass, number of passes. }\label{tab:runtime}}
\end{table}

\subsection{Floating Point Operations (Theoretical)}
\label{supp:flops}
Because prospectors only rely on forward passes from an equipped encoder, our approach to feature attribution is approximately 3-4$\times$ more computationally efficient than gradient-based attribution methods. This analysis is based on empirical results from computing floating point operations (FLOPs) for model inference (\ie passes with frozen weights). Since prospectors’ FLOPs are significantly less than a forward pass at inference time, this efficiency boost is approximated by the forward-backward pass FLOP ratio of 1:2 to 1:3 \citep{Zhou2021-tx, Baydin2017-ie}. This efficiency is especially relevant for multi-billion parameter foundation models that could be used as upstream encoders.

\subsection{Additional Properties}
\label{supp:props}
Prospectors have additional desirable properties:
\begin{itemize}
    \vspace{-4pt}
    \setlength\itemsep{-1pt}
    \item A form of \say{glocal} attribution: prospectors simultaneously build a global, dataset-level kernel of scored concept associations while also building local, datum-specific prospect maps
    \item \textit{Interpretable}: prospector kernels can be inspected and verified by users to interpret class-specific concepts
    \item Arguably \textit{generative} in nature via kernel construction: kernels can be rescaled and interpreted as joint probabilities, \ie $\mathbb{P}(c_i, y)$ for monograms and $\mathbb{P}(c_j, c_k, y)$ for skip-bigrams
    \item Shift- and rotation-equivariant, thus order-free \citep{Bronstein2021-jr}: controlling for any randomness, fitting and inference can start at any origin vertex to yield consistent attributions 
    \item Scale-invariant: input data can be of any number of tokens
    \item Can be \textit{deterministic}: given the above, fold-change variants (which have no trainable parameters) can create deterministic prospect maps if hypothesis testing is forgone
    \item Theoretically can learn \textit{implicit} skip-$n$-grams over a datum, as discussed in \cref{supp:theory}
\end{itemize}



\subsection{Theoretical Insights}
\label{supp:theory}

\subsubsection{Implicit $n$-grams}

Prospector heads are inspired by the induction heads \citep{Olsson2022-fj}, also referred to as $n$-gram heads \citep{Akyurek2024-rq}, found in trained transformers for language modeling — even inspiring our method's name. While induction heads perform a sort of \say{pattern completion} \citep{Olsson2022-fj} using tokens, our approach achieves a form of \say{pattern recognition} and simplifies this computation and parameter space in multiple ways: a quantization of token to a set of $K$ concepts and the explicit learning of monograms and skip-bigrams ($n=1$ and $n=2$). 

We claim that this strategy to learn monograms and skip-bigrams is sufficient for implicitly learning higher order $n$-gram targets. Namely, we argue that skip-bigrams can be “chained” together to form implicit skip-$n$-grams during attribution, \ie during the creation of prospect maps in layer (II). For example, iteration $i$ of convolution may find a skip-bigram of concepts A–B within the $r$-sized receptive field (i.e. A and B may be up to $r$ hops away) and then iteration $i+r$ may find a skip-bigram of concepts B–C. Together, one can argue that both skip-bigrams form an implicit skip-trigram A–B–C. This implicit chaining of skip-$n$-grams can also lead to implicitly capturing longer-range dependencies. In Theorem 1 below, we show that skip-$n$-grams can be implicitly chained up to $(n-1)r$ hops away in a map graph of tokens, $G$. 

\begin{theorem}[Range of implicit $n$-grams]
    \label{thm:implict-n-gram}
    \normalfont Given a map graph $G$ of cardinality $T$, prospectors with receptive field $r$ and an ideal kernel can find all target 1-grams, skip-2-grams, $\dots$, skip-$n$-grams spanning up to $(n-1)r$ node hops.
\end{theorem}

\textbf{\textit{Proof sketch.}} First, we explore the $n:=1$ case (\ie monograms). Here, all target 1-grams are found trivially via kernel look-ups. Next, we take a look at the $n:=2$ case (\ie skip-bigrams). Given the receptive field $r$, skip-bigrams can be found up to $r$ hops away from the central node. Both the $n:=1$ and $n:=2$ cases can be generalized to single \texttt{k2conv} iterations over a large graph $G$ (large $T$) or for small $G$ where $T \leq 2r$ ($G$ fully captured within $r$ hops). Given prospectors natively find monograms and skip-bigrams, multiple convolutional iterations are needed to find $n \geq 3$. We explore these cases next.

For the $n:=3$ case, \ie skip-trigrams, two skip-bigrams must be found in sequence with a shared token between them. We call this process \say{bigram chaining.} Given a skip-bigram can be learned over $r$ hops, prospectors can thus learn a skip-trigram over $2r$ node hops. The desired property trivially generalizes over any choice of $n$ (and $r$) via induction. $\blacksquare$

Through the kernel’s “memorization” of salient monograms and skip-bigram \say{rules,} prospectors offer flexibility without exorbitant parameterization (as with attention) — \ie the kernel does not need to see and learn a particular skip-$n$-gram in training, but at inference-time it can implicitly construct and recognize higher order skip-$n$-grams from its learned bigrams.

\subsubsection{Prospector Failure Modes}
One potential failure mode for prospection is triggered by small receptive fields ($r$), which can prevent prospectors from learning target skip-bigrams or skip-$n$-grams for any $n$. In the previous section, we show how prospectors can potentially \say{chain} skip-bigrams to implicitly learn higher-order skip-$n$-grams (as seen with transformer induction heads). However this expressivity is hinged on a sufficient choice of $r$ — prospectors must ensure the $r$-size field captures the target bigrams at the minimum. We hope to study this potential failure mode with synthetic benchmarks in future works.

\subsubsection{Impossibility Theorems for Feature Attribution}
Finally, another main motivation in prospector design is recent work on impossibility theorems \citep{Bilodeau2022-lm}, showing that (a) complete and (b) linear attribution methods can provably fail to improve on random guessing for inferring model behavior. Our approach sought to develop attribution methods outside of these traditional axioms (a) and (b) \citep{Sundararajan2017-ps}. Prospector heads are \textit{not} complete by nature of not constraining all token attribution scores in a datum to sum to a class prediction. The linear model variant uses its coefficients to attribute tokens, while the fold-change variant does not even output a class prediction.

\section{Experimental details}
\label{supp:all_modeling}

\subsection{Speed benchmarking for Inference}
\label{supp:speed}
We run a speed benchmarking analysis between two main encoder-attribution pipelines: (1) MiniLM with a prospector head and (2) DeBERTa with a zero-shot classification head and PartitionSHAP. Given the Huggingface implementation for DeBERTa's zero-shot classification, PartitionSHAP was automatically selected by the \texttt{shap} Python package (over other SHAP methods like DeepSHAP). We present the speed benchmarking in \cref{tab:speed}.

\begin{table}
\centering
\begin{tabular}{lcccl}
\toprule & \multicolumn{2}{c}{CPU time} & \multicolumn{2}{c}{Wall clock time} \\ 
\cmidrule(lr){2-3} \cmidrule(lr){4-5}
Attribution & token & datum & token  & datum \\
\midrule
SHAP & 25.459 & 1428.347 & 19.222 & 1078.738 \\
Prospectors & & 0.108 & & 0.108 \\
\bottomrule
\end{tabular}
\vspace{0pt}
\parbox{0.7 \textwidth}{ \vspace{5pt}
\caption{Speed benchmarking between PartitionSHAP (denoted as SHAP) and prospectors, as applied to partial-context encoders and text (WikiSection) data. We report mean values, and SHAP's values reflect a random sample of 86\% (621/718) of test-set examples.} \label{tab:speed}}
\end{table}

\subsection{Hyperparameter Tuning via Grid Search}
\label{supp:hparam}

For each task, we conduct a grid-search of hyperparameter configurations to select an optimal prospector model. The prospector kernel has two main hyperparameters—the number of concepts $k$ and the skip-gram neighborhood radius $r$. We also evaluate two prospector variants based on how the kernel is trained: hypothesis testing (with additional hyperparameters for the p-value $\alpha$ and fold change $\tau$ cutoffs) and linear modeling (with elastic net mixing hyperparameter $\lambda)$. We describe all tested hyperparameters in our training grid search in \cref{tab:hparams}.
\begin{table}
    \centering
    \begin{tabular}{lccc}
        \toprule
        Name & WikiSection & Camleyon16 & MetalPDB \\
        \midrule
        Token resolution  & sentence & $224 \times 224$ patch & atom \\
        Token connectivity ($\delta$)  & 2(-hop) & 8(-way) & - \\
        Concept count ($K$) & \{10,15,20,25,30\} & \{10,15,20,25,30\} & \{15,20,25,30\}  \\
        Receptive field ($r$)  & \{0,1,2,4,8\} & \{0,1,2,4,8\} & \{0,1,2,4\}  \\
        Significance threshold ($\alpha$) & \{0.01,0.025,0.05,$\infty$\} & \{0.01,0.025,0.05,$\infty$\} & \{0.001,0.01,0.05,0.5,$\infty$\}  \\
        Fold-change threshold ($\tau$) & \{0,1,2\} & \{0,1,2\} & \{0,1,2,4\}  \\
        Regularization factor ($\lambda$) & 0.5 & 0.5 & \{0.0,0.5,1.0\}  \\
        Edge cutoff ($\epsilon$) & - & - & \{4.0,6.0,8.0\}  \\
        \bottomrule
    \end{tabular}
    \vspace{3pt}
    \parbox{0.85 \textwidth}{ \vspace{5pt}
    \caption{Hyperparameters tuned during training grid search. Note: edge cutoff ($\epsilon$), the distance cutoff to control graph density, only applies to MetalPDB. We use $\infty$ to represent any large number that acts as a non-threshold.} \label{tab:hparams}}
\end{table}

\subsection{Hyperparameter \& Model Selection via Sequential Ranking}
\label{supp:model-select}
To select a top prospector configuration after the training grid-search, we first compute four token-level evaluation metrics for training set localization and apply sequential ranking over those chosen metrics. Applied in order, our chosen metrics were: precision, Matthews correlation coefficient (MCC), Dice coefficient, and AUPRC. These metrics were chosen because they enable segmentation-style evaluation, and we preferentially select on precision because it is especially important for detecting the small-scale class-specific regions in our data. For metrics that require a threshold (precision, MCC, and Dice coefficient), we select models based on the highest value attained over 11 thresholds: $0.0, 0.1, \dots, 1.0$. Top prospectors per encoder, selected from the grid search and sequential ranking, are listed in \cref{tab:topmodel}. 
\begin{table}
    \centering
    \vspace{0pt}
    \begin{tabular}{lcccccc}
        \toprule
        {Encoder Alias} & $K$ & $r$ & $\tau$ & $\alpha$ & $\lambda$ & $\epsilon$ \\
        \midrule
        MiniLM & 25 & 1 & 1 & 0.05 & — & —  \\
        DeBERTa & — & — & — & — & — & —  \\
        \midrule
        tile2vec & 20 & 8 & 0 & $\infty$ & — & — \\
        ViT & 20 & 2 & 0 & 0.05 & — & — \\
        CLIP & 30 & 2 & 2 & $\infty$ & — & — \\
        PLIP & 15 & 1 & 2 & 0.01 & — & — \\
        \midrule
        COLLAPSE & 25 & 4 & 4 & 1.0 & — & 8.0 \\
        ESM2 & 30 & 1 & 0 & 1.0 & — & 8.0\\
        AA & 21$^*$ & 1 & 1 & 0.1 & 1.0 & 6.0 \\
        \bottomrule
    \end{tabular}
    \vspace{3pt}
    \parbox{0.65 \textwidth}{\vspace{5pt}
    \caption{Top prospectors per encoder, after model selection and sequential ranking. All selected prospectors except AA are parameter-free fold-change variants. $^*$The AA encoder does not use clustering for quantization, since amino acids are already a discrete set of 21 tokens (20 standard amino acids + 1 entry for any non-standard amino acid). Symbol \say{—} denotes non-applicable hyperparameter.}     \label{tab:topmodel}}
\end{table}

\subsection{Test Set Evaluation}
\label{supp:testing}
After prospect graphs are created by prospector heads, we map back the values of each token to its original coordinates (referred to in the main body as \say{prospect maps}). Prior to evaluation, we feature scale values to $[0,1]$. Experimenting with other feature scaling schemes, \eg dataset-level scaling based on minimum and maximum values, is left for future work. For reporting results on the held-out test set we focus mainly on AUPRC to provide a threshold-agnostic evaluation of each method. To compute AP, average the precision scores over a set of predetermined thresholds to binarize the prospect maps, as described in the previous section: $0.0, 0.1, \dots, 1.0$.

\subsection{Construction of Task Datasets}
\label{supp:data_construct}

\subsubsection{Sequences (WikiSection)}
\label{supp:wiki}
Wikisection's \say{disease} annotated subset contains $n=3231$ documents total with 2513 training examples and 718 test examples. We preprocess the data into classes by searching each document for the presence of \say{disease.genetics} section labels. If this section label is found, we assign a document-level label of class$_1$ and class$_0$ otherwise. Because our task is at the sentence-level, we then create tokens by breaking sections into sentences by the full-stop delimiter (\say{.}). We then label sentences by their source section labels. Raw-text sentences are then fed into our chose encoder, which handles natural language tokenization.

\subsubsection{Images (Camelyon16)}
This benchmark contains 400 gigapixel whole slide images (270 train, 130 test) of breast cancer metastases in sentinel lymph nodes. All images were partitioned into prespecified patch tokens (size $224 \times 224$) and filtered for foreground tissue regions (as opposed to the glass background of the slide). This process resulted in more than 200K unique patches without augmentation. For ground truth annotations, binary masks were resized with inter-area interpolation and re-binarized (value of 1 is assigned if interpolated value $>0$) to match the dimensionality of data sprites. 

\begin{figure}
    \centering
    \vspace{0em}
    \includegraphics[width=0.65\textwidth, trim={0 0 0 0},clip]{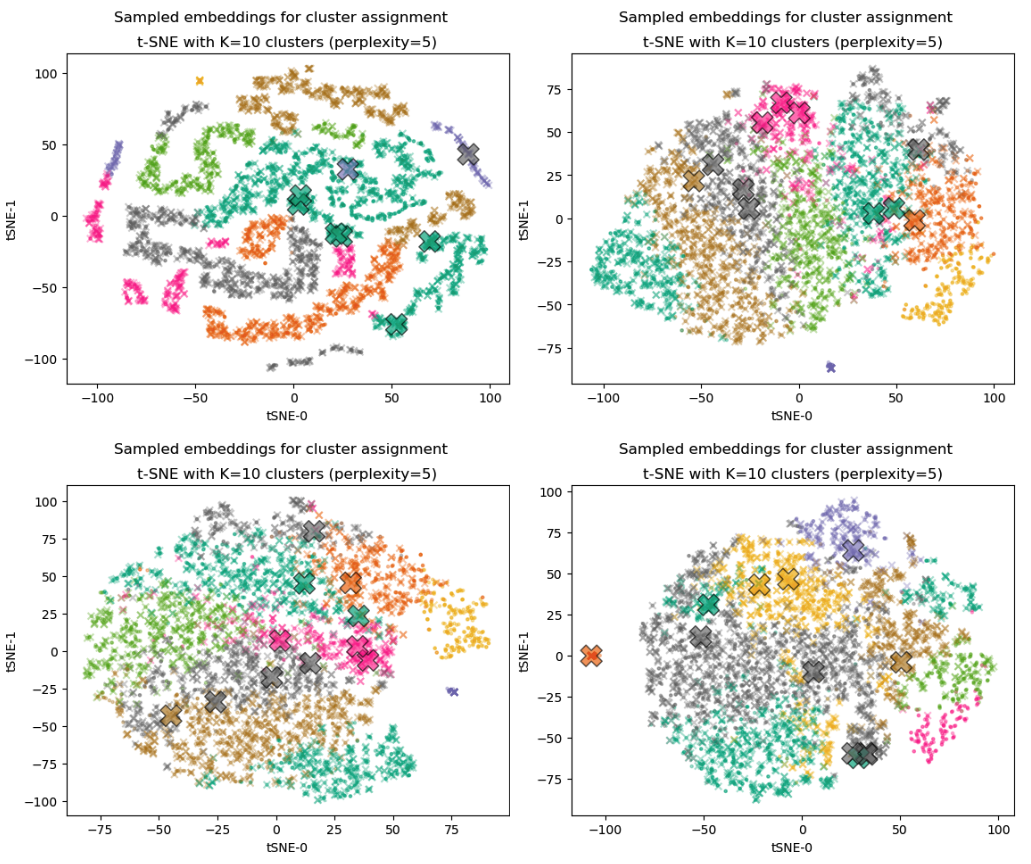}
    \vspace{14pt}
    \setlength{\belowcaptionskip}{-16pt}
    \parbox{0.7 \textwidth}{\vspace{20pt}
    \caption{t-SNE plots, left-to-right then top-to-bottom: tile2vec, ViT, CLIP, PLIP. Marker color denotes concept, marker type denotes ground truth annotation for a patch: $\circ$ for class$_0$, $\times$ for patches from class$_1$ images that do originate from target regions, and the much thicker $\pmb{\times}$ for class$_1$-specific target regions.}  \label{fig:tsne}}
\end{figure}

We also visualize the token embedding spaces of our encoders for the image task in \cref{fig:tsne}. The lack of natural clustering of class$_1$-specific tokens (thick $\pmb{\times}$ markers) from class$_0$ tokens ($\circ$ markers) intuitively depicts the difficulty of our task. In other words, class-specific regions are made up of tokens that are conceptually similar to non-region tokens.

\subsubsection{Graphs (MetalPDB)}
\label{supp:metalpdb}
We constructed a binary classification dataset of zinc-binding and non-binding proteins from the MetalPDB dataset \citep{Putignano2018-qb}. We specifically focus on proteins annotated as enzymes, since metal ions are often critical for enzymatic activity. Such enzymes are known as metallo-enzymes, and our global classification labels reflect whether a metallo-enzyme relies on zinc or a different metal ion. For the positive set, we consider only biologically-relevant zinc ions which occur within a chain (\ie are bound to residues in the main chain of the protein, rather than ligand-binding or crystallization artifacts). We sample only one protein chain from each enzymatic class, as determined by Enzyme Commission numbers \citep{Bairoch2000-un}, selecting the structure with the best crystallographic resolution.  This process resulted in 756 zinc-binding sites from 610 proteins, with 653 corresponding non-zinc-binding proteins sampled from unique enzymatic classes using the same procedure. For each zinc ion in the positive set, we extract all interacting residues annotated in MetalPDB to serve as our ground truth nodes for feature attribution. This dataset was split by enzyme class to ensure that no enzyme exists in both train and test sets, reserving 20\% of chains for held-out evaluation. After removing four structures which produced embedding errors, this produced a training set of 1007 unique protein chains for the train set and 252 for the test set. Each protein is featurized as a graph where each node represents a residue and edges are defined between residues which share any atom within a distance of $\epsilon$ angstroms, where $\epsilon$ varies the density of the graph.

\subsection{Multi-class Settings}
Prospectors can be easily adapted to the multi-class setting by training multiple models for each class of interest. For example, if faced with three classes $a$, $b$, $c$, prospectors could be applied in the following settings (class-1 and class-0, respectively):
\begin{itemize}
    \vspace{-4pt}
    \setlength\itemsep{-2pt}
    \item Prospector trained on $a$ vs. $\{b,c\}$
    \item Prospector trained on $b$ vs. $\{a,c\}$
    \item Prospector trained on $c$ vs. $\{a,b\}$
\end{itemize}
In fact, both our protein (MetalPDB) and text (WikiSection) datasets are adapted from multi-class settings: MetalPDB contains data for many different metals, and WikiSection contains 27 different labels in the English disease document subset. In each case, we selected one class to evaluate for simplicity (zinc-binding proteins and genetics-related text, respectively), but one could easily construct an analogous dataset and train a model for any other class label.

\subsection{Pre-trained Encoders}
\label{supp:pretrained}
We outline specific models and how to access them.

\subsubsection{Text (WikiSection)}
\textbf{MiniLM:} We specifically use the \texttt{all-MiniLM-L6-v2} sentence Transformer model (via the \textit{sentence-transformers} package), which is approximately 22M parameters in size.

\textbf{DeBERTa:} We specifically use the \texttt{DeBERTa-v3-base-mnli-fever-anli} model (via the \textit{transformers} package), which is approximately 98M parameters in size. DeBERTa is able to perform off-the-shelf ZSC and NLI. 

Note: We forgo prospection with DeBERTa since it emits word embeddings without a simple and effective way to construct sentence embeddings. We reiterate that annotations are at the sentence-level (\ie our prespecified tokens).

\subsubsection{Images (Camelyon16)}
\textbf{CLIP:} We specifically use the \texttt{clip-vit-base-patch32} model via the \textit{transformers} package. 

\textbf{PLIP:} We specifically use the \texttt{plip} model via the \textit{transformers} package.

\subsubsection{Graphs (MetalPDB)}
\textbf{COLLAPSE:} We use the implementation and weights available at \texttt{https://github.com/awfderry/COLLAPSE}.

\textbf{ESM:} We use the ESM implementation and weights available at \texttt{https://github.com/facebookresearch/esm}, specifically the 33-layer, 650M parameter ESM2 model (\texttt{esm2\_t33\_650M\_UR50D}).

\subsection{Trained Encoders}
\label{supp:trained}
We train two backbone encoders to equip with prospectors for the image task (Camelyon16):

\textbf{tile2vec:} This encoder uses a ResNet-18 architecture \citep{He2015-hp} trained for 20 epochs on a single NVIDIA T4 GPU. For training, the training set of 200K patches were formed into nearly 100K triplets with a sampling scheme similar to that of \citet{Jean2019-np}. These triplets were then used to train tile2vec with the triplet loss function \citep{Jean2019-np}.

\textbf{ViT:} We trained a custom ViT for trained for 30 epochs on a single NVIDIA T4 GPU. It was trained to perform IID patch predictions under coarse supervision, which involved image-level label inheritance \citep{Machiraju2022-yd} — the process of propagating image-level class labels to all constituent patches.

\subsection{Baseline Feature Attribution Methods}
\label{supp:baseline}
\subsubsection{Sequences (WikiSection)}
\textbf{MiniLM: support vector machine (SVM):} Using our sampled training embeddings from clustering, we train a one-class SVM on class$_0$ token embeddings ($n=4809$) to perform novelty detection on the held-out test set. The SVM was implemented with the \textit{sklearn} package and trained with an RBF kernel and hyperparameter $\gamma=1/d$ (where $d$ is embedding dimension). Training ran until a stopping criterion was satisfied with 1e-3 tolerance. 

\textbf{MiniLM: multi-layer perceptron (MLP):} Using our sampled training embeddings from clustering, we train an MLP on all ($n=5000$) token embeddings (labeled as class$_0$ or class$_1$) to perform fully supervised token classification held-out test set — acting as a stand-in for a segmentation-like baseline. The MLP was implemented with the \textit{sklearn} package and trained with one hidden layer (dimension 100), ReLU activations, adam optimizer, L2-regularization term of 1e-4, initial learning rate 1e-3, and minibatch size of 200. Training ran for a maximum of 1000 iterations, where inputs are shuffled.

\textbf{DeBERTa: zero-shot classification (ZSC):} The DeBERTa model can perform ZSC out of the box, giving us sentence-level ZSC probabilities (\ie logits). We used the ZSC binary labels of [\say{genetics}, \say{other}]. While we considered all possible labels in the WikiSection dataset's disease subset (see below), we ultimately went with binary classification due to higher performance.

Unused multi-class labels: [\say{genetics}, \say{other}, \say{classification}, \say{treatment}, \say{symptom}, \say{screening}, \say{prognosis}, \say{tomography}, \say{mechanism}, \say{pathophysiology}, \say{epidemiology}, \say{geography}, \say{medication}, \say{fauna}, \say{surgery}, \say{prevention}, \say{infection}, \say{culture}, \say{research}, \say{history}, \say{risk}, \say{cause}, \say{complication}, \say{pathology}, \say{management}, \say{diagnosis}, \say{etymology}]

\textbf{DeBERTa: ZSC with PartitionSHAP:} We implemented a Shapley scoring pipeline for DeBERTa since it can perform ZSC end-to-end. The pipeline was implemented via the \textit{transformers} package using the object class \texttt{zeroShotClassificationPipeline}. Shapley computation was performed via the \textit{shap} package. The pipeline defaults to PartitionSHAP — in this setting, PartitionSHAP is applied to the partial-context DeBERTa model and considers sub-tokens (words) over all possible $T$ tokens (sentences), ultimately pooling over sub-tokens to get a token-level score. We run a speed benchmarking analysis for this approach in \cref{supp:speed}.

Note: full-context shapley score computation was also considered. However, due to poor computational scaling (for both DeBERTa and PartitionSHAP), we ruled out this strategy.

\textbf{DeBERTa: NLI entailment:} The DeBERTa model can perform NLI entailment off the shelf, yielding sentence-level NLI entailment scores. We provide the model with an NLI hypothesis (\say{this sentence is about genetics}) and NLI premise (\ie the input sentence). Labels extracted refer to [\say{entailment}, \say{neutral}, \say{contradiction}].

\textbf{DeBERTa: NLI entailment attention:} The DeBERTa model can also output attention scores. Attention scores are computed by max-pooling over the attention weights for the NLI hypothesis (\say{this sentence is about genetics}) given the NLI premise (\ie the input sentence).

\subsubsection{Images (Camelyon16)}
For this task, baselines were chosen due to their popularity and efficiency. 

\textbf{Concatenated mean attention (ViT only):} attention maps are created per input token and their values are averaged. This creates a single attention score per token, after which tokens are concatenated by their spatial coordinates. These values are scaled to values in $[0,1]$.

\textbf{Concatenated prediction probability:} For ViT, each token's prediction probability for class$_1$ is used to score each token, after which tokens are concatenated by their spatial coordinates. For both vision-language models, CLIP and PLIP, we prompt both FMs' text encoders with zero-shot classification labels for class$_0$ and class$_1$, respectively: [\say{normal lymph node}, \say{lymph node metastasis}]. These labels match the benchmark dataset's descriptions of class labels. Similarly to ViT, each token's class$_1$ prediction probability is used to score each token, after which tokens are concatenated by their spatial coordinates

\subsubsection{Graphs (MetalPDB)}

Our baseline for zinc binding residue identification is a graph attention network (GAT) \citep{Velickovic2017-oy} containing two GAT layers, each followed by batch normalization, followed by a global mean pooling and a fully-connected output layer. The input node features for each residue were given by the choice of encoder (COLLAPSE, ESM2, or AA). The GAT model was trained using weak supervision (\ie on graph-level labels $y$) using a binary cross-entropy loss and Adam optimizer with default parameters and weight decay of $1 \times 10^{-4}$. To select the best baseline model, we use a gridsearch over the edge cutoff ($\epsilon$) for the underlying protein graph (6.0 or 8.0 Å), the learning rate ($1 \times 10^{-5}$, $1 \times 10^{-4}$, $5 \times 10^{-4}$, $1 \times 10^{-3}$), and the GAT node feature dimension (100, 200, 500). Feature attribution for all explanation methods was performed using implementations provided by Pytorch Geometric \citep{Fey2019-pa}. The best model was selected using the selection criteria in \cref{supp:hparam}. The final classification models for COLLAPSE, ESM, and AA encoders used edge cutoffs ($\epsilon$) of 8.0 Å, 6.0 Å, and 6.0 Å, learning rates of $5 \times 10^{-4}$, $5 \times 10^{-3}$, and $5 \times 10^{-3}$, and feature dimensions of 100, 500, and 200, respectively.

\textbf{GAT head + GNNExplainer:}  We use GNNExplainer \citep{Ying2019-cu} to produce explanations for nodes (\ie residues) only. We train the GNNExplainer module for 100 epochs with a default learning rate of 0.01.

\textbf{GAT head + Attention:} The attention baseline uses the attention scores of the trained GAT model to produce attribution scores. Attention scores across layers and heads are first max-pooled to produce aggregated attention scores for each edge. Then, we compute the attribution score for each node by averaging over the scores of all edges connected to it.

\textbf{GAT head + SHAP:} We deploy SHAP using a Shapley value sampling approach adapted specifically for graph data and implemented using Captum (\url{https://captum.ai}). SHAP is computationally feasible for this task primarily due to the use of full-context classifier heads to plug into each tested encoder. This allows SHAP to explain individual tokens (amino acids) by aggregating over edge weights using the same procedure as described for our attention baseline.

\subsection{Tabular Results}
\label{supp:tab}
We report quantitative results corresponding to \cref{fig:perf} in tables \ref{tab:results-wiki}, \ref{tab:results-cam}, and \ref{tab:results-metalpdb}. All reported errors reflect the standard error of the mean.
\begin{table}
    \centering
            \begin{tabular}{lcccc}
                \toprule
                Encoder-Attribution & Mean AUPRC & Error AUPRC & Mean AP & Error AP\\
                \midrule
                DeBERTa-ZSC & 0.476 & 0.032 & 0.502 & 0.031 \\
                DeBERTa-SHAP & 0.292 & 0.026 & 0.322 & 0.026  \\
                DeBERTa-NLI & \textbf{0.584} & 0.030 & \textbf{0.617} & 0.023  \\
                DeBERTa-Attention & 0.217 & 0.020 & 0.244 & 0.021  \\
                MiniLM-SVM & 0.284 & 0.024 & 0.317 & 0.024 \\
                MiniLM-MLP & \textit{0.626} & 0.031 & \textit{0.648} & 0.030 \\
                \midrule
                MiniLM-Prospector & \textbf{0.711} & 0.030 & \textbf{0.730} & 0.028 \\
                \bottomrule
            \end{tabular}
            \vspace{6pt}
            \parbox{0.75 \textwidth}{\vspace{3pt} \caption{Tabular results for sequences (WikiSection). Top section contains baseline methods while bottom section contains prospector-equipped encoders. \textbf{Boldface} indicates best-in-encoder results. \textit{Italics} indicates top non-prospector pipeline.} \label{tab:results-wiki}}
\end{table}

\begin{table}
            \centering
            \vspace{0pt}
            \begin{tabular}{lcccc}
                \toprule
                Encoder-Attribution & Mean AUPRC & Error AUPRC & Mean AP & Error AP\\
                \midrule
                ViT-Attention & 0.158 & 0.038 & 0.162 & 0.038 \\
                ViT-Probability & \textit{0.207} & 0.043 & \textit{0.212} & 0.043 \\
                CLIP-Probability & 0.149 & 0.035 & 0.155 & 0.034 \\
                PLIP-Probability & 0.163 & 0.037 & 0.167 & 0.037 \\
                \midrule
                tile2vec-Prospector & \textbf{0.212} & 0.044 & \textbf{0.218} & 0.044 \\
                ViT-Prospector      & \textbf{0.210} & 0.047 & \textbf{0.215} & 0.047 \\
                CLIP-Prospector     & \textbf{0.330} & 0.056 & \textbf{0.298} & 0.055 \\
                PLIP-Prospector     & \textbf{0.470} & 0.050 & \textbf{0.300} & 0.052 \\
                \bottomrule
            \end{tabular}
            \vspace{3pt}
            \parbox{0.75 \textwidth}{\vspace{3pt}\caption{Tabular results for images (Camelyon16). Top section contains baseline methods while bottom section contains prospector-equipped encoders. \textbf{Boldface} indicates best-in-encoder results. \textit{Italics} indicates top non-prospector pipeline.}\label{tab:results-cam}}
\end{table}

\begin{table} 
            \centering
            \begin{tabular}{lcccc}
                \toprule
                Encoder-Attribution & Mean AUPRC & Error AUPRC & Mean AP & Error AP \\
                \midrule
                COLLAPSE-GNNExplainer & 0.242 & 0.020 & \textit{0.266} & 0.020 \\
                COLLAPSE-Attention & 0.171 & 0.013 & 0.199 & 0.013  \\
                COLLAPSE-SHAP & 0.370 & 0.015 & 0.162 & 0.016 \\
                ESM-GNNExplainer & 0.036 &  0.003 & 0.047 & 0.004 \\
                ESM-Attention & 0.050 & 0.005 & \textbf{0.064} & 0.005  \\
                ESM-SHAP & \textbf{0.304} & 0.015  & 0.062 & 0.006 \\
                AA-GNNExplainer & 0.105 &  0.015  & 0.121 & 0.015 \\
                AA-Attention & 0.031 & 0.005 & 0.043 & 0.004 \\
                AA-SHAP & \textit{\textbf{0.420}} & 0.011  & 0.062 & 0.008\\
                \midrule
                COLLAPSE-Prospector & \textbf{0.640} & 0.020 & \textbf{0.323} & 0.039 \\
                ESM-Prospector & 0.082 & 0.013 & 0.060 & 0.007 \\
                AA-Prospector & 0.405 & 0.037 & \textbf{0.354} & 0.037 \\
                \bottomrule
            \end{tabular}
            \vspace{3pt}
            \parbox{0.75 \textwidth}{\vspace{3pt} \caption{Tabular results for protein graphs (MetalPDB). Top section contains baseline methods while bottom section contains prospector-equipped encoders. \textbf{Boldface} indicates best-in-encoder results. \textit{Italics} indicates top non-prospector pipeline.} \label{tab:results-metalpdb}}
\end{table}

\subsection{Robustness to Coarse Supervision}
\label{supp:robustness}
\begin{figure}
    \centering
    \vspace{0pt}
    \includegraphics[width=0.77\textwidth, trim={5em 0 13em 0},clip]{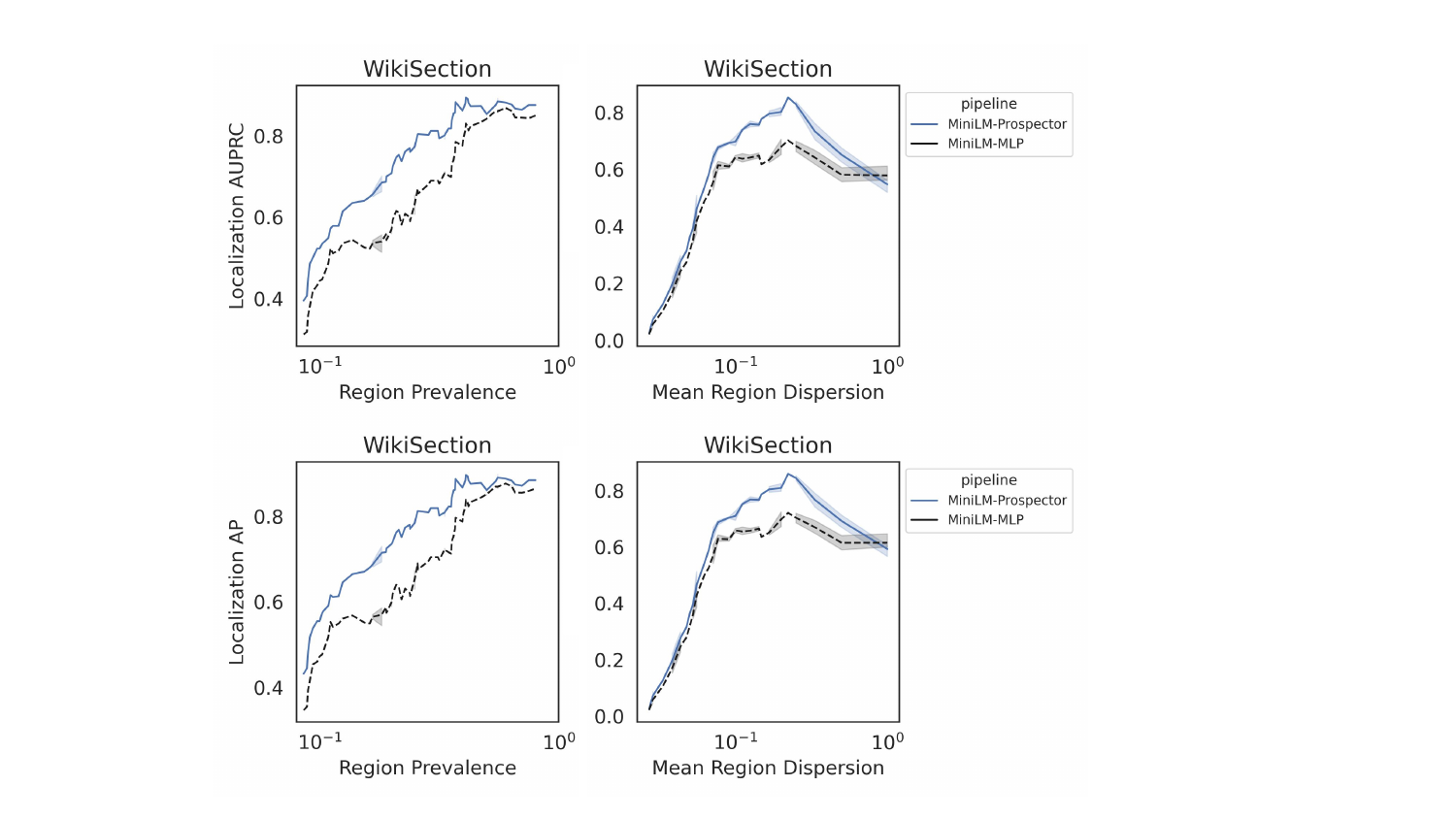}
    \vspace{5pt}
    \setlength{\belowcaptionskip}{-10pt}
    \parbox{0.65 \textwidth}{\vspace{5pt} \caption{Robustness for WikiSection data. Top baseline, MiniLM encoder with MLP head, is denoted by a black dashed line.} \label{fig:prop-wiki}}
\end{figure}

\begin{figure}
    \centering
    \vspace{0pt}
    \includegraphics[width=0.7\textwidth, trim={13em 0.5em 13em 0},clip]{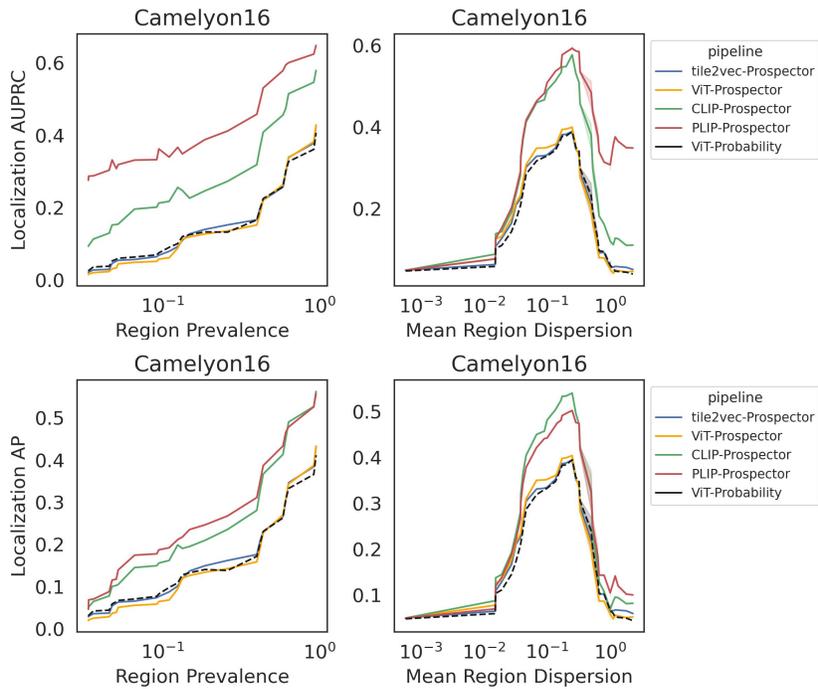}
    \vspace{5pt}
    \setlength{\belowcaptionskip}{-10pt}
    \parbox{0.65 \textwidth}{\vspace{5pt} \caption{Robustness for Camelyon16 data. Top baseline, ViT encoder with prediction probabilities, is denoted by a black dashed line.}\label{fig:prop-cam}}
\end{figure}

We also briefly study the robustness of prospector (and top baseline) test-set performance with respect to salient region characteristics: region prevalence and mean region dispersion. We display these results in figures \ref{fig:prop-wiki}, \ref{fig:prop-cam}, and \ref{fig:prop-metal}. The more that lines gravitate to the top of each plot, the more robust an encoder-attribution pipeline is to target region characteristics. Lines are created by convolving over the test-set examples. 

\subsubsection{Additional analysis for MetalPDB}
\label{supp:prot-props}
\begin{figure}
    \centering
    \vspace{0pt}
    \includegraphics[width=0.8\textwidth, trim={13em 1em 5em 0},clip]{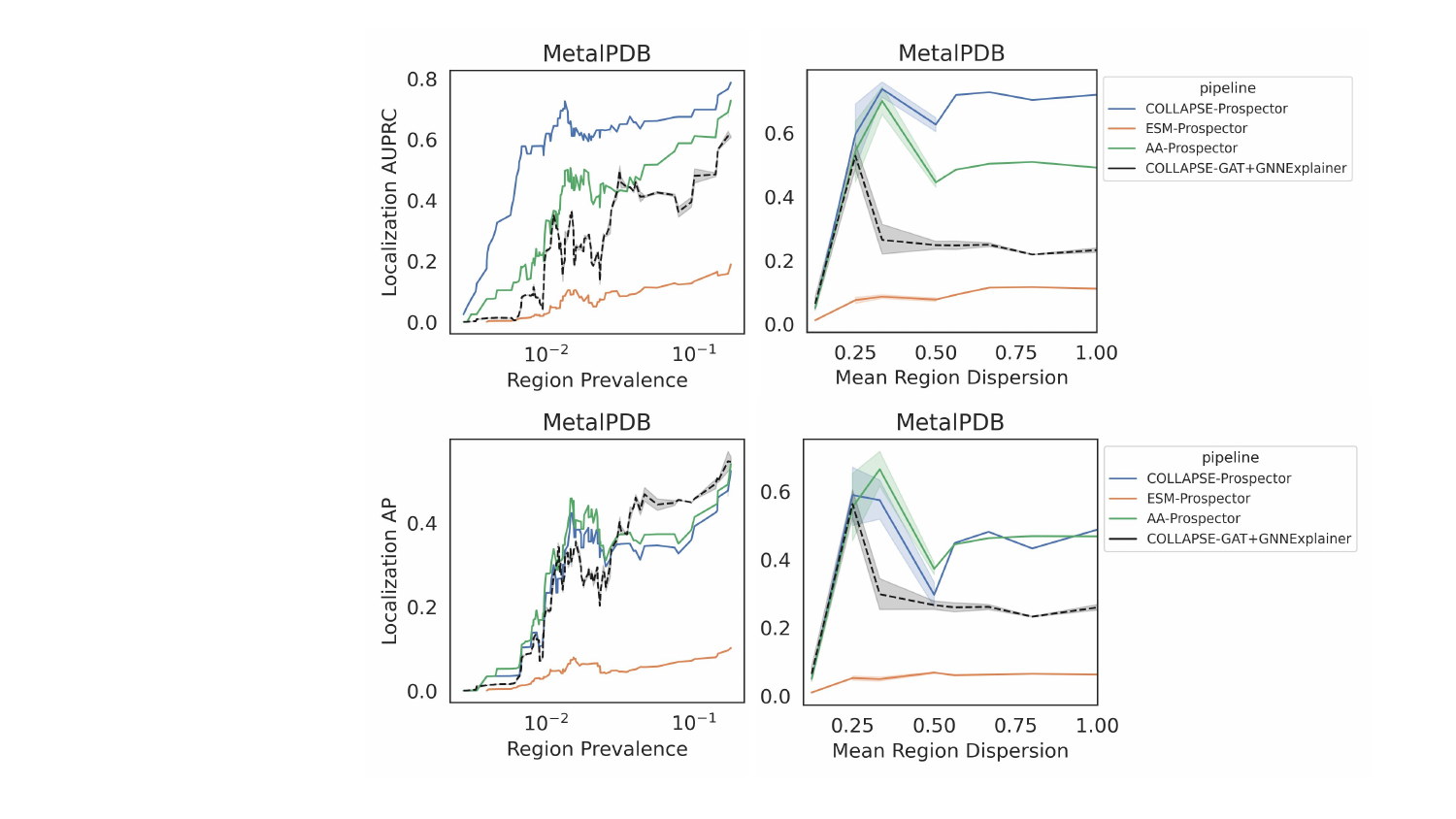}
    \vspace{5pt}
    \setlength{\belowcaptionskip}{-10pt}
    \parbox{0.65 \textwidth}{\vspace{5pt} \caption{Robustness for MetalPDB data. Top baseline, COLLAPSE encoder with GAT head and GNNExplainer, is denoted by a black dashed line.} \label{fig:prop-metal}}
\end{figure}

The metal-binding protein task is particularly challenging as the majority of its class-specific regions are below 0.1 prevalence, but prospectors were nonetheless able to achieve high performance on most test-set examples (\cref{fig:prop-metal}). Interestingly, ESM2 showed bimodal performance, with high AUPRC on one subset and a correlated, low performance on another. This suggests that a subset of data does not contain clear sequence patterns that are correlated with zinc binding, while structure-based encoders can capture local interactions between residues far apart in sequence. In addition to the prevalence of class-specific regions, mean region dispersion provides a view into their spatial organization.

\subsection{Domain-Specific Analysis of Prospector Internals}
\begin{figure}
    \centering
    \vspace{0pt}
    \includegraphics[width=0.9\textwidth, trim={3em 9em 3em 10em},clip]{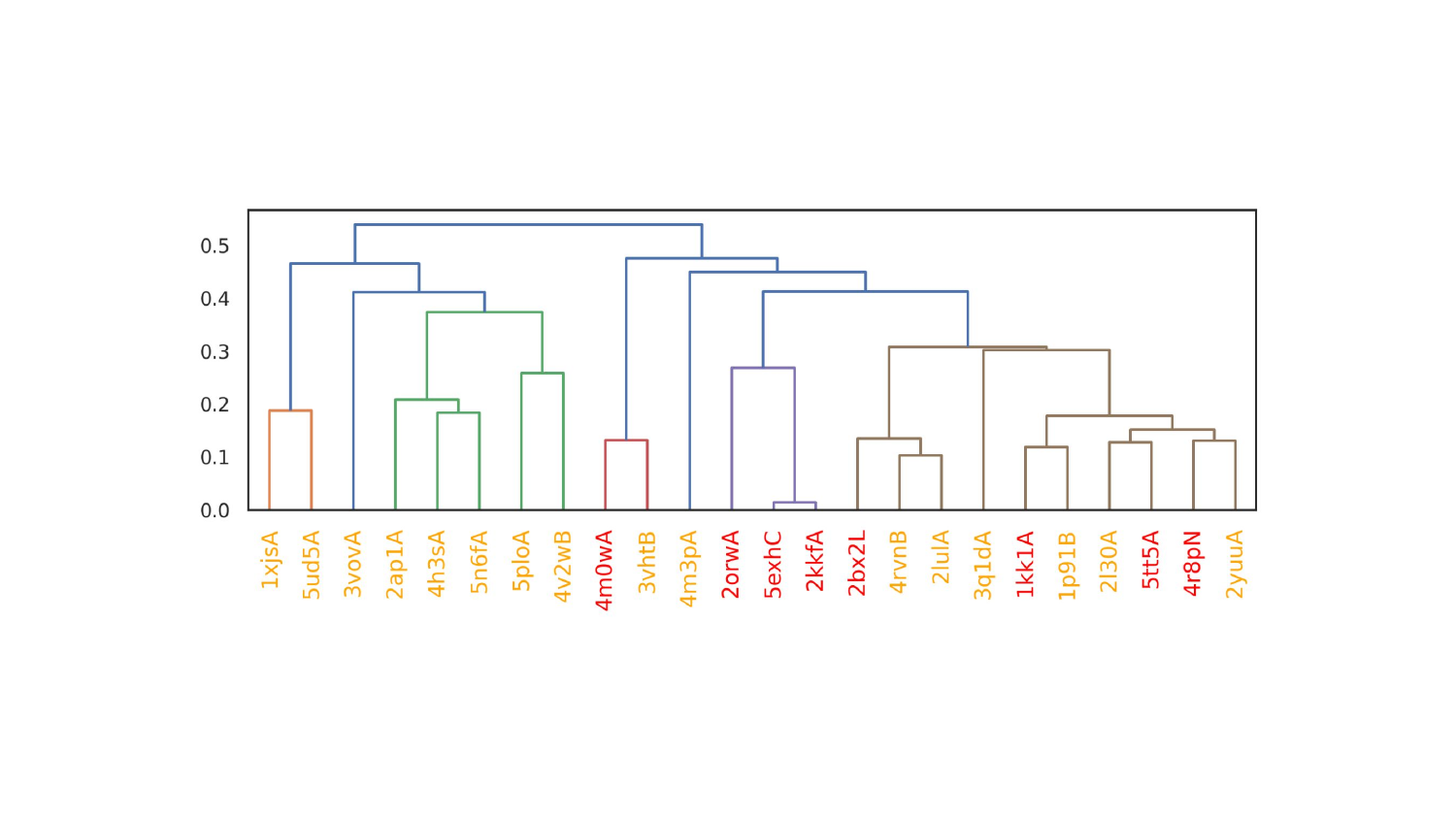}
    \vspace{8pt}
    \parbox{0.85 \textwidth}{\vspace{0pt}  \caption{Hierarchical clustering results for sprite embeddings computed from high-precision test-set examples. X-axis labels are colored by the number of cysteine residues coordinating the central zinc ion, which is a key feature that is correlated with the resulting clusters: orange=3, red=4.}\label{fig:supp-clust}}
\end{figure}

\cref{fig:supp-clust} displays the results of hierarchically clustering sprite embeddings for the zinc binding task. 
\label{subsection:zn-cluster}

%% file: paper.bbl
\begin{thebibliography}{110}
\providecommand{\natexlab}[1]{#1}
\providecommand{\url}[1]{\texttt{#1}}
\expandafter\ifx\csname urlstyle\endcsname\relax
  \providecommand{\doi}[1]{doi: #1}\else
  \providecommand{\doi}{doi: \begingroup \urlstyle{rm}\Url}\fi

\bibitem[Adebayo et~al.(2018)Adebayo, Gilmer, Muelly, Goodfellow, Hardt, and Kim]{Adebayo2018-ls}
Adebayo, J., Gilmer, J., Muelly, M., Goodfellow, I., Hardt, M., and Kim, B.
\newblock Sanity checks for saliency maps.
\newblock \emph{arXiv [cs.CV]}, October 2018.

\bibitem[Adebayo et~al.(2022)Adebayo, Muelly, Abelson, and Kim]{Adebayo2022-fz}
Adebayo, J., Muelly, M., Abelson, H., and Kim, B.
\newblock Post hoc explanations may be ineffective for detecting unknown spurious correlation.
\newblock \emph{arXiv [cs.LG]}, December 2022.

\bibitem[Afchar et~al.(2021)Afchar, Hennequin, and Guigue]{Afchar2021-xb}
Afchar, D., Hennequin, R., and Guigue, V.
\newblock Towards rigorous interpretations: a formalisation of feature attribution.
\newblock \emph{arXiv [cs.LG]}, April 2021.

\bibitem[Aky{\"u}rek et~al.(2024)Aky{\"u}rek, Wang, Kim, and Andreas]{Akyurek2024-rq}
Aky{\"u}rek, E., Wang, B., Kim, Y., and Andreas, J.
\newblock {In-Context} language learning: Architectures and algorithms.
\newblock \emph{arXiv [cs.CL]}, January 2024.

\bibitem[Alain \& Bengio(2016)Alain and Bengio]{Alain2016-xa}
Alain, G. and Bengio, Y.
\newblock Understanding intermediate layers using linear classifier probes.
\newblock \emph{arXiv [stat.ML]}, October 2016.

\bibitem[Amores(2013)]{Amores2013-rk}
Amores, J.
\newblock Multiple instance classification: Review, taxonomy and comparative study.
\newblock \emph{Artif. Intell.}, 201:\penalty0 81--105, August 2013.

\bibitem[Ancona et~al.(2019)Ancona, {\"O}ztireli, and Gross]{Ancona2019-kg}
Ancona, M., {\"O}ztireli, C., and Gross, M.
\newblock Explaining deep neural networks with a polynomial time algorithm for shapley values approximation.
\newblock \emph{arXiv [cs.LG]}, March 2019.

\bibitem[Anders \& Huber(2010)Anders and Huber]{Anders2010-zg}
Anders, S. and Huber, W.
\newblock Differential expression analysis for sequence count data.
\newblock \emph{Genome Biol.}, 11\penalty0 (10):\penalty0 R106, October 2010.

\bibitem[Arnold et~al.(2019)Arnold, Schneider, Cudr{\'e}-Mauroux, Gers, and L{\"o}ser]{Arnold2019-sz}
Arnold, S., Schneider, R., Cudr{\'e}-Mauroux, P., Gers, F.~A., and L{\"o}ser, A.
\newblock {SECTOR}: A neural model for coherent topic segmentation and classification.
\newblock \emph{arXiv [cs.CL]}, February 2019.

\bibitem[Arun et~al.(2020)Arun, Gaw, Singh, Chang, Aggarwal, Chen, Hoebel, Gupta, Patel, Gidwani, Adebayo, Li, and Kalpathy-Cramer]{Arun2020-jp}
Arun, N., Gaw, N., Singh, P., Chang, K., Aggarwal, M., Chen, B., Hoebel, K., Gupta, S., Patel, J., Gidwani, M., Adebayo, J., Li, M.~D., and Kalpathy-Cramer, J.
\newblock Assessing the (un)trustworthiness of saliency maps for localizing abnormalities in medical imaging.
\newblock \emph{bioRxiv}, July 2020.

\bibitem[Bairoch(2000)]{Bairoch2000-un}
Bairoch, A.
\newblock The {ENZYME} database in 2000.
\newblock \emph{Nucleic Acids Res.}, 28\penalty0 (1):\penalty0 304--305, January 2000.

\bibitem[Baydin et~al.(2017)Baydin, Pearlmutter, Radul, and Siskind]{Baydin2017-ie}
Baydin, A.~G., Pearlmutter, B.~A., Radul, A.~A., and Siskind, J.~M.
\newblock Automatic differentiation in machine learning: a survey.
\newblock \emph{J. Mach. Learn. Res.}, 18\penalty0 (1):\penalty0 5595--5637, January 2017.

\bibitem[Belinkov(2021)]{Belinkov2021-mo}
Belinkov, Y.
\newblock Probing classifiers: Promises, shortcomings, and advances.
\newblock \emph{arXiv [cs.CL]}, February 2021.

\bibitem[Beltagy et~al.(2020)Beltagy, Peters, and Cohan]{Beltagy2020-zj}
Beltagy, I., Peters, M.~E., and Cohan, A.
\newblock Longformer: The {Long-Document} transformer.
\newblock \emph{arXiv [cs.CL]}, April 2020.

\bibitem[Berman et~al.(2002)Berman, Battistuz, Bhat, Bluhm, Bourne, Burkhardt, Feng, Gilliland, Iype, Jain, Fagan, Marvin, Padilla, Ravichandran, Schneider, Thanki, Weissig, Westbrook, and Zardecki]{Berman2002-hh}
Berman, H.~M., Battistuz, T., Bhat, T.~N., Bluhm, W.~F., Bourne, P.~E., Burkhardt, K., Feng, Z., Gilliland, G.~L., Iype, L., Jain, S., Fagan, P., Marvin, J., Padilla, D., Ravichandran, V., Schneider, B., Thanki, N., Weissig, H., Westbrook, J.~D., and Zardecki, C.
\newblock The protein data bank.
\newblock \emph{Acta Crystallogr. D Biol. Crystallogr.}, 58\penalty0 (Pt 61):\penalty0 899--907, June 2002.

\bibitem[Bietti et~al.(2023)Bietti, Cabannes, Bouchacourt, Jegou, and Bottou]{Bietti2023-wd}
Bietti, A., Cabannes, V., Bouchacourt, D., Jegou, H., and Bottou, L.
\newblock Birth of a transformer: A memory viewpoint.
\newblock \emph{arXiv [stat.ML]}, June 2023.

\bibitem[Bilodeau et~al.(2022)Bilodeau, Jaques, Koh, and Kim]{Bilodeau2022-lm}
Bilodeau, B., Jaques, N., Koh, P.~W., and Kim, B.
\newblock Impossibility theorems for feature attribution.
\newblock \emph{arXiv [cs.LG]}, December 2022.

\bibitem[Bommasani et~al.(2021)Bommasani, Hudson, Adeli, Altman, Arora, von Arx, Bernstein, Bohg, Bosselut, Brunskill, Brynjolfsson, Buch, Card, Castellon, Chatterji, Chen, Creel, Davis, Demszky, Donahue, Doumbouya, Durmus, Ermon, Etchemendy, Ethayarajh, Fei-Fei, Finn, Gale, Gillespie, Goel, Goodman, Grossman, Guha, Hashimoto, Henderson, Hewitt, Ho, Hong, Hsu, Huang, Icard, Jain, Jurafsky, Kalluri, Karamcheti, Keeling, Khani, Khattab, Kohd, Krass, Krishna, Kuditipudi, Kumar, Ladhak, Lee, Lee, Leskovec, Levent, Li, Li, Ma, Malik, Manning, Mirchandani, Mitchell, Munyikwa, Nair, Narayan, Narayanan, Newman, Nie, Niebles, Nilforoshan, Nyarko, Ogut, Orr, Papadimitriou, Park, Piech, Portelance, Potts, Raghunathan, Reich, Ren, Rong, Roohani, Ruiz, Ryan, R{\'e}, Sadigh, Sagawa, Santhanam, Shih, Srinivasan, Tamkin, Taori, Thomas, Tram{\`e}r, Wang, Wang, Wu, Wu, Wu, Xie, Yasunaga, You, Zaharia, Zhang, Zhang, Zhang, Zhang, Zheng, Zhou, and Liang]{Bommasani2021-ca}
Bommasani, R., Hudson, D.~A., Adeli, E., Altman, R., Arora, S., von Arx, S., Bernstein, M.~S., Bohg, J., Bosselut, A., Brunskill, E., Brynjolfsson, E., Buch, S., Card, D., Castellon, R., Chatterji, N., Chen, A., Creel, K., Davis, J.~Q., Demszky, D., Donahue, C., Doumbouya, M., Durmus, E., Ermon, S., Etchemendy, J., Ethayarajh, K., Fei-Fei, L., Finn, C., Gale, T., Gillespie, L., Goel, K., Goodman, N., Grossman, S., Guha, N., Hashimoto, T., Henderson, P., Hewitt, J., Ho, D.~E., Hong, J., Hsu, K., Huang, J., Icard, T., Jain, S., Jurafsky, D., Kalluri, P., Karamcheti, S., Keeling, G., Khani, F., Khattab, O., Kohd, P.~W., Krass, M., Krishna, R., Kuditipudi, R., Kumar, A., Ladhak, F., Lee, M., Lee, T., Leskovec, J., Levent, I., Li, X.~L., Li, X., Ma, T., Malik, A., Manning, C.~D., Mirchandani, S., Mitchell, E., Munyikwa, Z., Nair, S., Narayan, A., Narayanan, D., Newman, B., Nie, A., Niebles, J.~C., Nilforoshan, H., Nyarko, J., Ogut, G., Orr, L., Papadimitriou, I., Park, J.~S., Piech, C., Portelance, E., Potts, C.,
  Raghunathan, A., Reich, R., Ren, H., Rong, F., Roohani, Y., Ruiz, C., Ryan, J., R{\'e}, C., Sadigh, D., Sagawa, S., Santhanam, K., Shih, A., Srinivasan, K., Tamkin, A., Taori, R., Thomas, A.~W., Tram{\`e}r, F., Wang, R.~E., Wang, W., Wu, B., Wu, J., Wu, Y., Xie, S.~M., Yasunaga, M., You, J., Zaharia, M., Zhang, M., Zhang, T., Zhang, X., Zhang, Y., Zheng, L., Zhou, K., and Liang, P.
\newblock On the opportunities and risks of foundation models.
\newblock \emph{arXiv [cs.LG]}, August 2021.

\bibitem[Bommasani et~al.(2023)Bommasani, Klyman, Longpre, Kapoor, Maslej, Xiong, Zhang, and Liang]{Bommasani2023-kb}
Bommasani, R., Klyman, K., Longpre, S., Kapoor, S., Maslej, N., Xiong, B., Zhang, D., and Liang, P.
\newblock The foundation model transparency index.
\newblock \emph{arXiv [cs.LG]}, October 2023.

\bibitem[Brocki \& Chung(2019)Brocki and Chung]{Brocki2019-fb}
Brocki, L. and Chung, N.~C.
\newblock Concept saliency maps to visualize relevant features in deep generative models.
\newblock In \emph{2019 18th {IEEE} International Conference On Machine Learning And Applications ({ICMLA})}, pp.\  1771--1778. IEEE, December 2019.

\bibitem[Bronstein et~al.(2021)Bronstein, Bruna, Cohen, and Veli{\v c}kovi{\'c}]{Bronstein2021-jr}
Bronstein, M.~M., Bruna, J., Cohen, T., and Veli{\v c}kovi{\'c}, P.
\newblock Geometric deep learning: Grids, groups, graphs, geodesics, and gauges.
\newblock \emph{arXiv [cs.LG]}, April 2021.

\bibitem[Brown et~al.(2020)Brown, Mann, Ryder, Subbiah, Kaplan, Dhariwal, Neelakantan, Shyam, Sastry, Askell, Agarwal, Herbert-Voss, Krueger, Henighan, Child, Ramesh, Ziegler, Wu, Winter, Hesse, Chen, Sigler, Litwin, Gray, Chess, Clark, Berner, McCandlish, Radford, Sutskever, and Amodei]{Brown2020-bh}
Brown, T.~B., Mann, B., Ryder, N., Subbiah, M., Kaplan, J., Dhariwal, P., Neelakantan, A., Shyam, P., Sastry, G., Askell, A., Agarwal, S., Herbert-Voss, A., Krueger, G., Henighan, T., Child, R., Ramesh, A., Ziegler, D.~M., Wu, J., Winter, C., Hesse, C., Chen, M., Sigler, E., Litwin, M., Gray, S., Chess, B., Clark, J., Berner, C., McCandlish, S., Radford, A., Sutskever, I., and Amodei, D.
\newblock Language models are {Few-Shot} learners.
\newblock \emph{arXiv [cs.CL]}, May 2020.

\bibitem[Bujel et~al.(2023)Bujel, Caines, Yannakoudakis, and Rei]{Bujel2023-qa}
Bujel, K., Caines, A., Yannakoudakis, H., and Rei, M.
\newblock Finding the needle in a haystack: Unsupervised rationale extraction from long text classifiers.
\newblock \emph{arXiv [cs.CL]}, March 2023.

\bibitem[Campanella et~al.(2019)Campanella, Hanna, Geneslaw, Miraflor, Silva, Busam, Brogi, Reuter, Klimstra, and Fuchs]{Campanella2019-sj}
Campanella, G., Hanna, M.~G., Geneslaw, L., Miraflor, A., Silva, V. W.~K., Busam, K.~J., Brogi, E., Reuter, V.~E., Klimstra, D.~S., and Fuchs, T.~J.
\newblock Clinical-grade computational pathology using weakly supervised deep learning on whole slide images.
\newblock \emph{Nat. Med.}, 25\penalty0 (8):\penalty0 1301, August 2019.

\bibitem[Chen et~al.(2022{\natexlab{a}})Chen, He, Narasimhan, and Chen]{Chen2022-qd}
Chen, H., He, J., Narasimhan, K., and Chen, D.
\newblock Can rationalization improve robustness?
\newblock \emph{arXiv [cs.CL]}, April 2022{\natexlab{a}}.

\bibitem[Chen et~al.(2022{\natexlab{b}})Chen, Lundberg, and Lee]{Chen2022-ej}
Chen, H., Lundberg, S.~M., and Lee, S.-I.
\newblock Explaining a series of models by propagating shapley values.
\newblock \emph{Nat. Commun.}, 13\penalty0 (1):\penalty0 4512, August 2022{\natexlab{b}}.

\bibitem[Chen et~al.(2023)Chen, Li, Biaz, Bui, and Nguyen]{Chen2023-ak}
Chen, P., Li, Q., Biaz, S., Bui, T., and Nguyen, A.
\newblock {gScoreCAM}: What objects is {CLIP} looking at?
\newblock In \emph{Computer Vision -- {ACCV} 2022}, pp.\  588--604. Springer Nature Switzerland, 2023.

\bibitem[Chen et~al.(2022{\natexlab{c}})Chen, Chen, Li, Chen, Trister, Krishnan, and Mahmood]{Chen2022-rv}
Chen, R.~J., Chen, C., Li, Y., Chen, T.~Y., Trister, A.~D., Krishnan, R.~G., and Mahmood, F.
\newblock Scaling vision transformers to gigapixel images via hierarchical {Self-Supervised} learning.
\newblock \emph{arXiv [cs.CV]}, June 2022{\natexlab{c}}.

\bibitem[Child et~al.(2019)Child, Gray, Radford, and Sutskever]{Child2019-jc}
Child, R., Gray, S., Radford, A., and Sutskever, I.
\newblock Generating long sequences with sparse transformers.
\newblock \emph{arXiv [cs.LG]}, April 2019.

\bibitem[Crabb{\'e} \& van~der Schaar(2022)Crabb{\'e} and van~der Schaar]{Crabbe2022-lh}
Crabb{\'e}, J. and van~der Schaar, M.
\newblock Concept activation regions: A generalized framework for {Concept-Based} explanations.
\newblock \emph{arXiv [cs.LG]}, September 2022.

\bibitem[Derry \& Altman(2023)Derry and Altman]{Derry2023-pg}
Derry, A. and Altman, R.~B.
\newblock {COLLAPSE}: A representation learning framework for identification and characterization of protein structural sites.
\newblock \emph{Protein Sci.}, 32\penalty0 (2):\penalty0 e4541, February 2023.

\bibitem[Dosovitskiy et~al.(2020)Dosovitskiy, Beyer, Kolesnikov, Weissenborn, Zhai, Unterthiner, Dehghani, Minderer, Heigold, Gelly, Uszkoreit, and Houlsby]{Dosovitskiy2020-wn}
Dosovitskiy, A., Beyer, L., Kolesnikov, A., Weissenborn, D., Zhai, X., Unterthiner, T., Dehghani, M., Minderer, M., Heigold, G., Gelly, S., Uszkoreit, J., and Houlsby, N.
\newblock An image is worth 16x16 words: Transformers for image recognition at scale.
\newblock \emph{arXiv [cs.CV]}, October 2020.

\bibitem[Ehteshami~Bejnordi et~al.(2017)Ehteshami~Bejnordi, Veta, Johannes~van Diest, van Ginneken, Karssemeijer, Litjens, van~der Laak, {the CAMELYON16 Consortium}, Hermsen, Manson, Balkenhol, Geessink, Stathonikos, van Dijk, Bult, Beca, Beck, Wang, Khosla, Gargeya, Irshad, Zhong, Dou, Li, Chen, Lin, Heng, Ha{\ss}, Bruni, Wong, Halici, {\"O}ner, Cetin-Atalay, Berseth, Khvatkov, Vylegzhanin, Kraus, Shaban, Rajpoot, Awan, Sirinukunwattana, Qaiser, Tsang, Tellez, Annuscheit, Hufnagl, Valkonen, Kartasalo, Latonen, Ruusuvuori, Liimatainen, Albarqouni, Mungal, George, Demirci, Navab, Watanabe, Seno, Takenaka, Matsuda, Ahmady~Phoulady, Kovalev, Kalinovsky, Liauchuk, Bueno, Fernandez-Carrobles, Serrano, Deniz, Racoceanu, and Ven{\^a}ncio]{Ehteshami_Bejnordi2017-ko}
Ehteshami~Bejnordi, B., Veta, M., Johannes~van Diest, P., van Ginneken, B., Karssemeijer, N., Litjens, G., van~der Laak, J. A. W.~M., {the CAMELYON16 Consortium}, Hermsen, M., Manson, Q.~F., Balkenhol, M., Geessink, O., Stathonikos, N., van Dijk, M.~C., Bult, P., Beca, F., Beck, A.~H., Wang, D., Khosla, A., Gargeya, R., Irshad, H., Zhong, A., Dou, Q., Li, Q., Chen, H., Lin, H.-J., Heng, P.-A., Ha{\ss}, C., Bruni, E., Wong, Q., Halici, U., {\"O}ner, M.~{\"U}., Cetin-Atalay, R., Berseth, M., Khvatkov, V., Vylegzhanin, A., Kraus, O., Shaban, M., Rajpoot, N., Awan, R., Sirinukunwattana, K., Qaiser, T., Tsang, Y.-W., Tellez, D., Annuscheit, J., Hufnagl, P., Valkonen, M., Kartasalo, K., Latonen, L., Ruusuvuori, P., Liimatainen, K., Albarqouni, S., Mungal, B., George, A., Demirci, S., Navab, N., Watanabe, S., Seno, S., Takenaka, Y., Matsuda, H., Ahmady~Phoulady, H., Kovalev, V., Kalinovsky, A., Liauchuk, V., Bueno, G., Fernandez-Carrobles, M.~M., Serrano, I., Deniz, O., Racoceanu, D., and Ven{\^a}ncio, R.
\newblock Diagnostic assessment of deep learning algorithms for detection of lymph node metastases in women with breast cancer.
\newblock \emph{JAMA}, 318\penalty0 (22):\penalty0 2199--2210, December 2017.

\bibitem[Fey \& Lenssen(2019)Fey and Lenssen]{Fey2019-pa}
Fey, M. and Lenssen, J.~E.
\newblock Fast graph representation learning with {PyTorch} geometric.
\newblock \emph{arXiv [cs.LG]}, March 2019.

\bibitem[Foulds \& Frank(2010)Foulds and Frank]{Foulds2010-qi}
Foulds, J. and Frank, E.
\newblock A review of {Multi-Instance} learning assumptions.
\newblock \emph{The Knowledge Engineering Review}, 0\penalty0 (0):\penalty0 1--24, 2010.

\bibitem[Ghorbani et~al.(2019)Ghorbani, Wexler, Zou, and Kim]{Ghorbani2019-cd}
Ghorbani, A., Wexler, J., Zou, J., and Kim, B.
\newblock Towards automatic concept-based explanations.
\newblock \emph{arXiv [stat.ML]}, February 2019.

\bibitem[Gondal et~al.(2023)Gondal, Gast, Ruiz, Droste, Macri, Kumar, and Staudigl]{Gondal2023-bg}
Gondal, M.~W., Gast, J., Ruiz, I.~A., Droste, R., Macri, T., Kumar, S., and Staudigl, L.
\newblock Domain aligned {CLIP} for few-shot classification.
\newblock \emph{arXiv [cs.CV]}, November 2023.

\bibitem[Graves et~al.(2013)Graves, Jaitly, and Mohamed]{Graves2013-lo}
Graves, A., Jaitly, N., and Mohamed, A.-R.
\newblock Hybrid speech recognition with deep bidirectional {LSTM}.
\newblock In \emph{2013 {IEEE} Workshop on Automatic Speech Recognition and Understanding}, pp.\  273--278. IEEE, December 2013.

\bibitem[Gu et~al.(2021)Gu, Goel, and R{\'e}]{Gu2021-bq}
Gu, A., Goel, K., and R{\'e}, C.
\newblock Efficiently modeling long sequences with structured state spaces.
\newblock \emph{arXiv [cs.LG]}, October 2021.

\bibitem[Halicek et~al.(2019)Halicek, Shahedi, Little, Chen, Myers, Sumer, and Fei]{Halicek2019-ru}
Halicek, M., Shahedi, M., Little, J.~V., Chen, A.~Y., Myers, L.~L., Sumer, B.~D., and Fei, B.
\newblock Head and neck cancer detection in digitized {Whole-Slide} histology using convolutional neural networks.
\newblock \emph{Sci. Rep.}, 9\penalty0 (1):\penalty0 14043, October 2019.

\bibitem[He et~al.(2015)He, Zhang, Ren, and Sun]{He2015-hp}
He, K., Zhang, X., Ren, S., and Sun, J.
\newblock Deep residual learning for image recognition.
\newblock \emph{arXiv [cs.CV]}, December 2015.

\bibitem[He et~al.(2020)He, Liu, Gao, and Chen]{He2020-nn}
He, P., Liu, X., Gao, J., and Chen, W.
\newblock {DeBERTa}: Decoding-enhanced {BERT} with disentangled attention.
\newblock \emph{arXiv [cs.CL]}, June 2020.

\bibitem[He et~al.(2021)He, Gao, and Chen]{He2021-wi}
He, P., Gao, J., and Chen, W.
\newblock {DeBERTaV3}: Improving {DeBERTa} using {ELECTRA-Style} {Pre-Training} with {Gradient-Disentangled} embedding sharing.
\newblock \emph{arXiv [cs.CL]}, November 2021.

\bibitem[Hopfield(1982)]{Hopfield1982-tu}
Hopfield, J.~J.
\newblock Neural networks and physical systems with emergent collective computational abilities.
\newblock \emph{Proc. Natl. Acad. Sci. U. S. A.}, 79\penalty0 (8):\penalty0 2554--2558, April 1982.

\bibitem[Hopfield(1984)]{Hopfield1984-yp}
Hopfield, J.~J.
\newblock Neurons with graded response have collective computational properties like those of two-state neurons.
\newblock \emph{Proc. Natl. Acad. Sci. U. S. A.}, 81\penalty0 (10):\penalty0 3088--3092, May 1984.

\bibitem[Huang et~al.(2023)Huang, Bianchi, Yuksekgonul, Montine, and Zou]{Huang2023-iv}
Huang, Z., Bianchi, F., Yuksekgonul, M., Montine, T., and Zou, J.
\newblock Leveraging medical twitter to build a visual--language foundation model for pathology {AI}.
\newblock \emph{bioRxiv}, pp.\  2023.03.29.534834, April 2023.

\bibitem[Jain \& Wallace(2019)Jain and Wallace]{Jain2019-to}
Jain, S. and Wallace, B.~C.
\newblock Attention is not explanation.
\newblock \emph{arXiv [cs.CL]}, February 2019.

\bibitem[Jakubik et~al.(2023)Jakubik, Roy, Phillips, Fraccaro, Godwin, Zadrozny, Szwarcman, Gomes, Nyirjesy, Edwards, Kimura, Simumba, Chu, Karthik~Mukkavilli, Lambhate, Das, Bangalore, Oliveira, Muszynski, Ankur, Ramasubramanian, Gurung, Khallaghi, {Hanxi}, {Li}, Cecil, Ahmadi, Kordi, Alemohammad, Maskey, Ganti, Weldemariam, and Ramachandran]{Jakubik2023-vj}
Jakubik, J., Roy, S., Phillips, C.~E., Fraccaro, P., Godwin, D., Zadrozny, B., Szwarcman, D., Gomes, C., Nyirjesy, G., Edwards, B., Kimura, D., Simumba, N., Chu, L., Karthik~Mukkavilli, S., Lambhate, D., Das, K., Bangalore, R., Oliveira, D., Muszynski, M., Ankur, K., Ramasubramanian, M., Gurung, I., Khallaghi, S., {Hanxi}, {Li}, Cecil, M., Ahmadi, M., Kordi, F., Alemohammad, H., Maskey, M., Ganti, R., Weldemariam, K., and Ramachandran, R.
\newblock Foundation models for generalist geospatial artificial intelligence.
\newblock \emph{arXiv [cs.CV]}, October 2023.

\bibitem[Javed et~al.(2022)Javed, Juyal, Padigela, Taylor-Weiner, Yu, and Prakash]{Javed2022-hj}
Javed, S., Juyal, D., Padigela, H., Taylor-Weiner, A., Yu, L., and Prakash, A.~a.
\newblock Additive {MIL}: Intrinsically interpretable multiple instance learning for pathology.
\newblock \emph{Adv. Neural Inf. Process. Syst.}, 2022.

\bibitem[Jean et~al.(2019)Jean, Wang, Samar, Azzari, Lobell, and Ermon]{Jean2019-np}
Jean, N., Wang, S., Samar, A., Azzari, G., Lobell, D., and Ermon, S.
\newblock {Tile2Vec}: Unsupervised representation learning for spatially distributed data.
\newblock In \emph{Proceedings of the {AAAI} Conference on Artificial Intelligence}, volume~33, pp.\  3967--3974, July 2019.

\bibitem[Jethani et~al.(2021)Jethani, Sudarshan, Covert, Lee, and Ranganath]{Jethani2021-yx}
Jethani, N., Sudarshan, M., Covert, I., Lee, S.-I., and Ranganath, R.
\newblock {FastSHAP}: {Real-Time} shapley value estimation.
\newblock \emph{arXiv [stat.ML]}, July 2021.

\bibitem[Jetley et~al.(2018)Jetley, Lord, Lee, and Torr]{Jetley2018-aw}
Jetley, S., Lord, N.~A., Lee, N., and Torr, P. H.~S.
\newblock Learn to pay attention.
\newblock \emph{arXiv [cs.CV]}, April 2018.

\bibitem[Kalibhat et~al.(2023)Kalibhat, Bhardwaj, Bruss, Firooz, Sanjabi, and Feizi]{Kalibhat2023-sr}
Kalibhat, N., Bhardwaj, S., Bruss, B., Firooz, H., Sanjabi, M., and Feizi, S.
\newblock Identifying interpretable subspaces in image representations.
\newblock \emph{arXiv [cs.CV]}, July 2023.

\bibitem[Karimi et~al.(2022)Karimi, Muandet, Kornblith, Sch{\"o}lkopf, and Kim]{Karimi2022-ek}
Karimi, A.-H., Muandet, K., Kornblith, S., Sch{\"o}lkopf, B., and Kim, B.
\newblock On the relationship between explanation and prediction: A causal view.
\newblock \emph{arXiv [cs.LG]}, December 2022.

\bibitem[Keles et~al.(2022)Keles, Wijewardena, and Hegde]{Keles2022-cw}
Keles, F.~D., Wijewardena, P.~M., and Hegde, C.
\newblock On the computational complexity of {Self-Attention}.
\newblock \emph{arXiv [cs.LG]}, September 2022.

\bibitem[Khosla et~al.(2023)Khosla, Zhu, and He]{Khosla2023-nv}
Khosla, S., Zhu, Z., and He, Y.
\newblock Survey on {Memory-Augmented} neural networks: Cognitive insights to {AI} applications.
\newblock \emph{arXiv [cs.AI]}, December 2023.

\bibitem[Kim et~al.(2017)Kim, Wattenberg, Gilmer, Cai, Wexler, Viegas, and Sayres]{Kim2017-ze}
Kim, B., Wattenberg, M., Gilmer, J., Cai, C., Wexler, J., Viegas, F., and Sayres, R.
\newblock Interpretability beyond feature attribution: Quantitative testing with concept activation vectors ({TCAV}).
\newblock \emph{arXiv [stat.ML]}, November 2017.

\bibitem[Kim et~al.(2024)Kim, Gadgil, DeGrave, Omiye, Cai, Daneshjou, and Lee]{Kim2024-wq}
Kim, C., Gadgil, S.~U., DeGrave, A.~J., Omiye, J.~A., Cai, Z.~R., Daneshjou, R., and Lee, S.-I.
\newblock Transparent medical image {AI} via an image-text foundation model grounded in medical literature.
\newblock \emph{Nat. Med.}, 30\penalty0 (4):\penalty0 1154--1165, April 2024.

\bibitem[Klemmer et~al.(2023)Klemmer, Rolf, Robinson, Mackey, and Ru{\ss}wurm]{Klemmer2023-oo}
Klemmer, K., Rolf, E., Robinson, C., Mackey, L., and Ru{\ss}wurm, M.
\newblock {SatCLIP}: Global, {General-Purpose} location embeddings with satellite imagery.
\newblock \emph{arXiv [cs.CV]}, November 2023.

\bibitem[Koh et~al.(2020)Koh, Nguyen, Tang, Mussmann, Pierson, Kim, and Liang]{Koh2020-bh}
Koh, P.~W., Nguyen, T., Tang, Y.~S., Mussmann, S., Pierson, E., Kim, B., and Liang, P.
\newblock Concept bottleneck models.
\newblock \emph{arXiv [cs.LG]}, July 2020.

\bibitem[Kohonen(1972)]{Kohonen1972-en}
Kohonen, T.
\newblock Correlation matrix memories.
\newblock \emph{IEEE Trans. Comput.}, C-21\penalty0 (4):\penalty0 353--359, April 1972.

\bibitem[Kuo et~al.(2022)Kuo, Cui, Gu, Piergiovanni, and Angelova]{Kuo2022-qt}
Kuo, W., Cui, Y., Gu, X., Piergiovanni, A.~J., and Angelova, A.
\newblock {F-VLM}: {Open-Vocabulary} object detection upon frozen vision and language models.
\newblock \emph{arXiv [cs.CV]}, September 2022.

\bibitem[Lam et~al.(2024)Lam, Teoh, Landay, Heer, and Bernstein]{Lam2024-bt}
Lam, M.~S., Teoh, J., Landay, J., Heer, J., and Bernstein, M.~S.
\newblock Concept induction: Analyzing unstructured text with {High-Level} concepts using {LLooM}.
\newblock \emph{arXiv [cs.HC]}, April 2024.

\bibitem[Lanusse et~al.(2023)Lanusse, Parker, Golkar, Cranmer, Bietti, Eickenberg, Krawezik, McCabe, Ohana, Pettee, Blancard, Tesileanu, Cho, and Ho]{Lanusse2023-xt}
Lanusse, F., Parker, L., Golkar, S., Cranmer, M., Bietti, A., Eickenberg, M., Krawezik, G., McCabe, M., Ohana, R., Pettee, M., Blancard, B. R.-S., Tesileanu, T., Cho, K., and Ho, S.
\newblock {AstroCLIP}: {Cross-Modal} {Pre-Training} for astronomical foundation models.
\newblock \emph{arXiv [astro-ph.IM]}, October 2023.

\bibitem[Lin et~al.(2023)Lin, Akin, Rao, Hie, Zhu, Lu, Smetanin, Verkuil, Kabeli, Shmueli, Dos Santos~Costa, Fazel-Zarandi, Sercu, Candido, and Rives]{Lin2023-gy}
Lin, Z., Akin, H., Rao, R., Hie, B., Zhu, Z., Lu, W., Smetanin, N., Verkuil, R., Kabeli, O., Shmueli, Y., Dos Santos~Costa, A., Fazel-Zarandi, M., Sercu, T., Candido, S., and Rives, A.
\newblock Evolutionary-scale prediction of atomic-level protein structure with a language model.
\newblock \emph{Science}, 379\penalty0 (6637):\penalty0 1123--1130, March 2023.

\bibitem[Liu \& Mukhopadhyay(2018)Liu and Mukhopadhyay]{Liu2018-ad}
Liu, Q. and Mukhopadhyay, S.
\newblock Unsupervised learning using pretrained {CNN} and associative memory bank.
\newblock In \emph{2018 International Joint Conference on Neural Networks ({IJCNN})}, pp.\  01--08. IEEE, July 2018.

\bibitem[Lu et~al.(2023)Lu, Chen, Williamson, Chen, Liang, Ding, Jaume, Odintsov, Zhang, Le, Gerber, Parwani, and Mahmood]{Lu2023-ar}
Lu, M.~Y., Chen, B., Williamson, D. F.~K., Chen, R.~J., Liang, I., Ding, T., Jaume, G., Odintsov, I., Zhang, A., Le, L.~P., Gerber, G., Parwani, A.~V., and Mahmood, F.
\newblock Towards a {Visual-Language} foundation model for computational pathology.
\newblock \emph{arXiv [cs.CV]}, July 2023.

\bibitem[Lundberg \& Lee(2017)Lundberg and Lee]{Lundberg2017-bw}
Lundberg, S. and Lee, S.-I.
\newblock A unified approach to interpreting model predictions.
\newblock \emph{arXiv [cs.AI]}, May 2017.

\bibitem[Machiraju et~al.(2022)Machiraju, Plevritis, and Mallick]{Machiraju2022-yd}
Machiraju, G., Plevritis, S., and Mallick, P.
\newblock A dataset generation framework for evaluating megapixel image classifiers and their explanations.
\newblock In \emph{Computer Vision -- {ECCV} 2022}, pp.\  422--442. Springer Nature Switzerland, 2022.

\bibitem[Niazi et~al.(2019)Niazi, Parwani, and Gurcan]{Niazi2019-ue}
Niazi, M. K.~K., Parwani, A.~V., and Gurcan, M.~N.
\newblock Digital pathology and artificial intelligence.
\newblock \emph{Lancet Oncol.}, 20\penalty0 (5):\penalty0 e253--e261, May 2019.

\bibitem[Olsson et~al.(2022)Olsson, Elhage, Nanda, Joseph, DasSarma, Henighan, Mann, Askell, Bai, Chen, Conerly, Drain, Ganguli, Hatfield-Dodds, Hernandez, Johnston, Jones, Kernion, Lovitt, Ndousse, Amodei, Brown, Clark, Kaplan, McCandlish, and Olah]{Olsson2022-fj}
Olsson, C., Elhage, N., Nanda, N., Joseph, N., DasSarma, N., Henighan, T., Mann, B., Askell, A., Bai, Y., Chen, A., Conerly, T., Drain, D., Ganguli, D., Hatfield-Dodds, Z., Hernandez, D., Johnston, S., Jones, A., Kernion, J., Lovitt, L., Ndousse, K., Amodei, D., Brown, T., Clark, J., Kaplan, J., McCandlish, S., and Olah, C.
\newblock In-context learning and induction heads.
\newblock \emph{arXiv [cs.LG]}, September 2022.

\bibitem[Parmar et~al.(2018)Parmar, Vaswani, Uszkoreit, Kaiser, Shazeer, Ku, and Tran]{Parmar2018-jx}
Parmar, N., Vaswani, A., Uszkoreit, J., Kaiser, {\L}., Shazeer, N., Ku, A., and Tran, D.
\newblock Image transformer.
\newblock \emph{arXiv [cs.CV]}, February 2018.

\bibitem[Pawlowski et~al.(2019)Pawlowski, Bhooshan, Ballas, Ciompi, Glocker, and Drozdzal]{Pawlowski2019-by}
Pawlowski, N., Bhooshan, S., Ballas, N., Ciompi, F., Glocker, B., and Drozdzal, M.
\newblock Needles in haystacks: On classifying tiny objects in large images.
\newblock \emph{arXiv [cs.CV]}, August 2019.

\bibitem[Poli et~al.(2023{\natexlab{a}})Poli, Massaroli, Nguyen, Fu, Dao, Baccus, Bengio, Ermon, and R{\'e}]{Poli2023-uw}
Poli, M., Massaroli, S., Nguyen, E., Fu, D.~Y., Dao, T., Baccus, S., Bengio, Y., Ermon, S., and R{\'e}, C.
\newblock Hyena hierarchy: Towards larger convolutional language models.
\newblock \emph{arXiv [cs.LG]}, February 2023{\natexlab{a}}.

\bibitem[Poli et~al.(2023{\natexlab{b}})Poli, Wang, Massaroli, Quesnelle, Carlow, Nguyen, and Thomas]{Poli2023-al}
Poli, M., Wang, J., Massaroli, S., Quesnelle, J., Carlow, R., Nguyen, E., and Thomas, A.
\newblock {StripedHyena}: {StripedHyena}: Moving beyond transformers with hybrid signal processing models.
\newblock \url{https://github.com/togethercomputer/stripedhyena}, November 2023{\natexlab{b}}.
\newblock Accessed: 2024-6-2.

\bibitem[Putignano et~al.(2018)Putignano, Rosato, Banci, and Andreini]{Putignano2018-qb}
Putignano, V., Rosato, A., Banci, L., and Andreini, C.
\newblock {MetalPDB} in 2018: a database of metal sites in biological macromolecular structures.
\newblock \emph{Nucleic Acids Res.}, 46\penalty0 (D1):\penalty0 D459--D464, January 2018.

\bibitem[Radford et~al.(2021)Radford, Kim, Hallacy, Ramesh, Goh, Agarwal, Sastry, Askell, Mishkin, Clark, Krueger, and Sutskever]{Radford2021-ln}
Radford, A., Kim, J.~W., Hallacy, C., Ramesh, A., Goh, G., Agarwal, S., Sastry, G., Askell, A., Mishkin, P., Clark, J., Krueger, G., and Sutskever, I.
\newblock Learning transferable visual models from natural language supervision.
\newblock In Meila, M. and Zhang, T. (eds.), \emph{Proceedings of the 38th International Conference on Machine Learning}, volume 139 of \emph{Proceedings of Machine Learning Research}, pp.\  8748--8763. PMLR, 2021.

\bibitem[Ramsauer et~al.(2020)Ramsauer, Sch{\"a}fl, Lehner, Seidl, Widrich, Adler, Gruber, Holzleitner, Pavlovi{\'c}, Sandve, Greiff, Kreil, Kopp, Klambauer, Brandstetter, and Hochreiter]{Ramsauer2020-dc}
Ramsauer, H., Sch{\"a}fl, B., Lehner, J., Seidl, P., Widrich, M., Adler, T., Gruber, L., Holzleitner, M., Pavlovi{\'c}, M., Sandve, G.~K., Greiff, V., Kreil, D., Kopp, M., Klambauer, G., Brandstetter, J., and Hochreiter, S.
\newblock Hopfield networks is all you need.
\newblock \emph{arXiv [cs.NE]}, July 2020.

\bibitem[Ribeiro et~al.(2016)Ribeiro, Singh, and Guestrin]{Ribeiro2016-hq}
Ribeiro, M.~T., Singh, S., and Guestrin, C.
\newblock ``why should {I} trust you?'': Explaining the predictions of any classifier.
\newblock \emph{arXiv [cs.LG]}, February 2016.

\bibitem[Rives et~al.(2021)Rives, Meier, Sercu, Goyal, Lin, Liu, Guo, Ott, Zitnick, Ma, and Fergus]{Rives2021-ck}
Rives, A., Meier, J., Sercu, T., Goyal, S., Lin, Z., Liu, J., Guo, D., Ott, M., Zitnick, C.~L., Ma, J., and Fergus, R.
\newblock Biological structure and function emerge from scaling unsupervised learning to 250 million protein sequences.
\newblock \emph{Proc. Natl. Acad. Sci. U. S. A.}, 118\penalty0 (15), April 2021.

\bibitem[Robinson et~al.(2020)Robinson, Jegelka, and Sra]{Robinson2020-ef}
Robinson, J., Jegelka, S., and Sra, S.
\newblock Strength from weakness: Fast learning using weak supervision.
\newblock \emph{arXiv [cs.LG]}, February 2020.

\bibitem[Saab et~al.(2024)Saab, Tu, Weng, Tanno, Stutz, Wulczyn, Zhang, Strother, Park, Vedadi, Chaves, Hu, Schaekermann, Kamath, Cheng, Barrett, Cheung, Mustafa, Palepu, McDuff, Hou, Golany, Liu, Alayrac, Houlsby, Tomasev, Freyberg, Lau, Kemp, Lai, Azizi, Kanada, Man, Kulkarni, Sun, Shakeri, He, Caine, Webson, Latysheva, Johnson, Mansfield, Lu, Rivlin, Anderson, Green, Wong, Krause, Shlens, Dominowska, Ali~Eslami, Cui, Vinyals, Kavukcuoglu, Manyika, Dean, Hassabis, Matias, Webster, Barral, Corrado, Semturs, Sara~Mahdavi, Gottweis, Karthikesalingam, and Natarajan]{Saab2024-fw}
Saab, K., Tu, T., Weng, W.-H., Tanno, R., Stutz, D., Wulczyn, E., Zhang, F., Strother, T., Park, C., Vedadi, E., Chaves, J.~Z., Hu, S.-Y., Schaekermann, M., Kamath, A., Cheng, Y., Barrett, D. G.~T., Cheung, C., Mustafa, B., Palepu, A., McDuff, D., Hou, L., Golany, T., Liu, L., Alayrac, J.-B., Houlsby, N., Tomasev, N., Freyberg, J., Lau, C., Kemp, J., Lai, J., Azizi, S., Kanada, K., Man, S., Kulkarni, K., Sun, R., Shakeri, S., He, L., Caine, B., Webson, A., Latysheva, N., Johnson, M., Mansfield, P., Lu, J., Rivlin, E., Anderson, J., Green, B., Wong, R., Krause, J., Shlens, J., Dominowska, E., Ali~Eslami, S.~M., Cui, C., Vinyals, O., Kavukcuoglu, K., Manyika, J., Dean, J., Hassabis, D., Matias, Y., Webster, D., Barral, J., Corrado, G., Semturs, C., Sara~Mahdavi, S., Gottweis, J., Karthikesalingam, A., and Natarajan, V.
\newblock Capabilities of gemini models in medicine.
\newblock \emph{arXiv [cs.AI]}, April 2024.

\bibitem[Selvaraju et~al.(2016)Selvaraju, Cogswell, Das, Vedantam, Parikh, and Batra]{Selvaraju2016-so}
Selvaraju, R.~R., Cogswell, M., Das, A., Vedantam, R., Parikh, D., and Batra, D.
\newblock {Grad-CAM}: Visual explanations from deep networks via gradient-based localization.
\newblock \emph{arXiv [cs.CV]}, October 2016.

\bibitem[Simonyan \& Zisserman(2014)Simonyan and Zisserman]{Simonyan2014-ta}
Simonyan, K. and Zisserman, A.
\newblock Very deep convolutional networks for {Large-Scale} image recognition.
\newblock \emph{arXiv [cs.CV]}, September 2014.

\bibitem[Singhal et~al.(2023)Singhal, Azizi, Tu, Mahdavi, Wei, Chung, Scales, Tanwani, Cole-Lewis, Pfohl, Payne, Seneviratne, Gamble, Kelly, Babiker, Sch{\"a}rli, Chowdhery, Mansfield, Demner-Fushman, Ag{\"u}era Y~Arcas, Webster, Corrado, Matias, Chou, Gottweis, Tomasev, Liu, Rajkomar, Barral, Semturs, Karthikesalingam, and Natarajan]{Singhal2023-ae}
Singhal, K., Azizi, S., Tu, T., Mahdavi, S.~S., Wei, J., Chung, H.~W., Scales, N., Tanwani, A., Cole-Lewis, H., Pfohl, S., Payne, P., Seneviratne, M., Gamble, P., Kelly, C., Babiker, A., Sch{\"a}rli, N., Chowdhery, A., Mansfield, P., Demner-Fushman, D., Ag{\"u}era Y~Arcas, B., Webster, D., Corrado, G.~S., Matias, Y., Chou, K., Gottweis, J., Tomasev, N., Liu, Y., Rajkomar, A., Barral, J., Semturs, C., Karthikesalingam, A., and Natarajan, V.
\newblock Large language models encode clinical knowledge.
\newblock \emph{Nature}, 620\penalty0 (7972):\penalty0 172--180, August 2023.

\bibitem[Song et~al.(2023)Song, Jaume, Williamson, Lu, Vaidya, Miller, and Mahmood]{Song2023-sq}
Song, A.~H., Jaume, G., Williamson, D. F.~K., Lu, M.~Y., Vaidya, A., Miller, T.~R., and Mahmood, F.
\newblock Artificial intelligence for digital and computational pathology.
\newblock \emph{Nature Reviews Bioengineering}, 1\penalty0 (12):\penalty0 930--949, October 2023.

\bibitem[Sparck~Jones(1972)]{Sparck_Jones1972-ke}
Sparck~Jones, K.
\newblock A statistical interpretation of term specificity and its application in retrieval.
\newblock \emph{Journal of Documentation}, 28\penalty0 (1):\penalty0 11--21, January 1972.

\bibitem[Sundararajan et~al.(2017)Sundararajan, Taly, and Yan]{Sundararajan2017-ps}
Sundararajan, M., Taly, A., and Yan, Q.
\newblock Axiomatic attribution for deep networks.
\newblock \emph{arXiv [cs.LG]}, March 2017.

\bibitem[Swanson et~al.(2022)Swanson, Chang, and Zou]{Swanson2022-on}
Swanson, K., Chang, H., and Zou, J.
\newblock Predicting immune escape with pretrained protein language model embeddings.
\newblock In Knowles, D.~A., Mostafavi, S., and Lee, S.-I. (eds.), \emph{Proceedings of the 17th Machine Learning in Computational Biology meeting}, volume 200 of \emph{Proceedings of Machine Learning Research}, pp.\  110--130. PMLR, 2022.

\bibitem[Talukder et~al.(2024)Talukder, Yue, and Gkioxari]{Talukder2024-ou}
Talukder, S., Yue, Y., and Gkioxari, G.
\newblock {TOTEM}: {TOkenized} time series {EMbeddings} for general time series analysis.
\newblock \emph{arXiv [cs.LG]}, February 2024.

\bibitem[Vaswani et~al.(2017)Vaswani, Shazeer, Parmar, Uszkoreit, Jones, Gomez, Kaiser, and Polosukhin]{Vaswani2017-ms}
Vaswani, A., Shazeer, N., Parmar, N., Uszkoreit, J., Jones, L., Gomez, A.~N., Kaiser, L., and Polosukhin, I.
\newblock Attention is all you need.
\newblock \emph{arXiv [cs.CL]}, June 2017.

\bibitem[Veli{\v c}kovi{\'c} et~al.(2017)Veli{\v c}kovi{\'c}, Cucurull, Casanova, Romero, Li{\`o}, and Bengio]{Velickovic2017-oy}
Veli{\v c}kovi{\'c}, P., Cucurull, G., Casanova, A., Romero, A., Li{\`o}, P., and Bengio, Y.
\newblock Graph attention networks.
\newblock \emph{arXiv [stat.ML]}, October 2017.

\bibitem[Wang et~al.(2019)Wang, Wang, Du, Yang, Zhang, Ding, Mardziel, and Hu]{Wang2019-tu}
Wang, H., Wang, Z., Du, M., Yang, F., Zhang, Z., Ding, S., Mardziel, P., and Hu, X.
\newblock {Score-CAM}: {Score-Weighted} visual explanations for convolutional neural networks.
\newblock \emph{arXiv [cs.CV]}, October 2019.

\bibitem[Wang et~al.(2023)Wang, Fu, Du, Gao, Huang, Liu, Chandak, Liu, Van~Katwyk, Deac, Anandkumar, Bergen, Gomes, Ho, Kohli, Lasenby, Leskovec, Liu, Manrai, Marks, Ramsundar, Song, Sun, Tang, Veli{\v c}kovi{\'c}, Welling, Zhang, Coley, Bengio, and Zitnik]{Wang2023-gp}
Wang, H., Fu, T., Du, Y., Gao, W., Huang, K., Liu, Z., Chandak, P., Liu, S., Van~Katwyk, P., Deac, A., Anandkumar, A., Bergen, K., Gomes, C.~P., Ho, S., Kohli, P., Lasenby, J., Leskovec, J., Liu, T.-Y., Manrai, A., Marks, D., Ramsundar, B., Song, L., Sun, J., Tang, J., Veli{\v c}kovi{\'c}, P., Welling, M., Zhang, L., Coley, C.~W., Bengio, Y., and Zitnik, M.
\newblock Scientific discovery in the age of artificial intelligence.
\newblock \emph{Nature}, 620\penalty0 (7972):\penalty0 47--60, August 2023.

\bibitem[Wang et~al.(2020)Wang, Wei, Dong, Bao, Yang, and Zhou]{Wang2020-zz}
Wang, W., Wei, F., Dong, L., Bao, H., Yang, N., and Zhou, M.
\newblock {MiniLM}: Deep {Self-Attention} distillation for {Task-Agnostic} compression of {Pre-Trained} transformers.
\newblock \emph{arXiv [cs.CL]}, February 2020.

\bibitem[Weber et~al.(2000)Weber, Welling, and Perona]{Weber2000-ah}
Weber, M., Welling, M., and Perona, P.
\newblock Unsupervised learning of models for recognition.
\newblock In \emph{Computer Vision - {ECCV} 2000}, Lecture notes in computer science, pp.\  18--32. Springer Berlin Heidelberg, Berlin, Heidelberg, 2000.

\bibitem[Wiegreffe \& Pinter(2019)Wiegreffe and Pinter]{Wiegreffe2019-my}
Wiegreffe, S. and Pinter, Y.
\newblock Attention is not not explanation.
\newblock \emph{arXiv [cs.CL]}, August 2019.

\bibitem[Wu et~al.(2010)Wu, Liu, and Altman]{Wu2010-gd}
Wu, S., Liu, T., and Altman, R.~B.
\newblock Identification of recurring protein structure microenvironments and discovery of novel functional sites around {CYS} residues.
\newblock \emph{BMC Struct. Biol.}, 10:\penalty0 4, February 2010.

\bibitem[Xu et~al.(2024)Xu, Usuyama, Bagga, Zhang, Rao, Naumann, Wong, Gero, Gonz{\'a}lez, Gu, Xu, Wei, Wang, Ma, Wei, Yang, Li, Gao, Rosemon, Bower, Lee, Weerasinghe, Wright, Robicsek, Piening, Bifulco, Wang, and Poon]{Xu2024-gs}
Xu, H., Usuyama, N., Bagga, J., Zhang, S., Rao, R., Naumann, T., Wong, C., Gero, Z., Gonz{\'a}lez, J., Gu, Y., Xu, Y., Wei, M., Wang, W., Ma, S., Wei, F., Yang, J., Li, C., Gao, J., Rosemon, J., Bower, T., Lee, S., Weerasinghe, R., Wright, B.~J., Robicsek, A., Piening, B., Bifulco, C., Wang, S., and Poon, H.
\newblock A whole-slide foundation model for digital pathology from real-world data.
\newblock \emph{Nature}, May 2024.

\bibitem[Yang et~al.(2023)Yang, Yin, He, Chang, Ma, and Xiang]{Yang2023-oo}
Yang, C., Yin, F., He, H., Chang, K.-W., Ma, X., and Xiang, B.
\newblock Efficient shapley values estimation by amortization for text classification.
\newblock \emph{arXiv [cs.CL]}, May 2023.

\bibitem[Ying et~al.(2019)Ying, Bourgeois, You, Zitnik, and Leskovec]{Ying2019-cu}
Ying, R., Bourgeois, D., You, J., Zitnik, M., and Leskovec, J.
\newblock {GNNExplainer}: Generating explanations for graph neural networks.
\newblock \emph{Adv. Neural Inf. Process. Syst.}, 32:\penalty0 9240--9251, December 2019.

\bibitem[Zaharia et~al.()Zaharia, Khattab, Chen, Davis, Miller, Potts, Zou, Carbin, Frankle, Rao, and Ghodsi]{Zaharia_undated-wd}
Zaharia, M., Khattab, O., Chen, L., Davis, J.~Q., Miller, H., Potts, C., Zou, J., Carbin, M., Frankle, J., Rao, N., and Ghodsi, A.
\newblock The shift from models to compound {AI} systems.
\newblock \url{https://bair.berkeley.edu/blog/2024/02/18/compound-ai-systems/}.
\newblock Accessed: 2024-5-17.

\bibitem[Zech et~al.(2018)Zech, Badgeley, Liu, Costa, Titano, and Oermann]{Zech2018-vb}
Zech, J.~R., Badgeley, M.~A., Liu, M., Costa, A.~B., Titano, J.~J., and Oermann, E.~K.
\newblock Confounding variables can degrade generalization performance of radiological deep learning models.
\newblock \emph{arXiv [cs.CV]}, July 2018.

\bibitem[Zhang et~al.(2023{\natexlab{a}})Zhang, Yu, Adhikarla, Zhou, Yan, Liu, Liu, He, Davison, Li, Ren, Fu, Zou, Liu, Huang, Chen, Zhou, Liu, Chen, Chen, Li, Liu, and Sun]{Zhang2023-ws}
Zhang, K., Yu, J., Adhikarla, E., Zhou, R., Yan, Z., Liu, Y., Liu, Z., He, L., Davison, B., Li, X., Ren, H., Fu, S., Zou, J., Liu, W., Huang, J., Chen, C., Zhou, Y., Liu, T., Chen, X., Chen, Y., Li, Q., Liu, H., and Sun, L.
\newblock {BiomedGPT}: A unified and generalist biomedical generative pre-trained transformer for vision, language, and multimodal tasks.
\newblock \emph{arXiv [cs.CL]}, May 2023{\natexlab{a}}.

\bibitem[Zhang et~al.(2023{\natexlab{b}})Zhang, Xu, Usuyama, Bagga, Tinn, Preston, Rao, Wei, Valluri, Wong, Lungren, Naumann, and Poon]{Zhang2023-kp}
Zhang, S., Xu, Y., Usuyama, N., Bagga, J., Tinn, R., Preston, S., Rao, R., Wei, M., Valluri, N., Wong, C., Lungren, M.~P., Naumann, T., and Poon, H.
\newblock {Large-Scale} {Domain-Specific} pretraining for biomedical {Vision-Language} processing.
\newblock \emph{arXiv [cs.CV]}, March 2023{\natexlab{b}}.

\bibitem[Zhang et~al.(2023{\natexlab{c}})Zhang, Zhao, Guo, and Yin]{Zhang2023-my}
Zhang, Z., Zhao, T., Guo, Y., and Yin, J.
\newblock {RS5M} and {GeoRSCLIP}: A large scale {Vision-Language} dataset and a large {Vision-Language} model for remote sensing.
\newblock \emph{arXiv [cs.CV]}, June 2023{\natexlab{c}}.

\bibitem[Zhou et~al.(2016)Zhou, Khosla, Lapedriza, Oliva, and Torralba]{Zhou2016-wy}
Zhou, B., Khosla, A., Lapedriza, A., Oliva, A., and Torralba, A.
\newblock Learning deep features for discriminative localization.
\newblock In \emph{2016 {IEEE} Conference on Computer Vision and Pattern Recognition ({CVPR})}. IEEE, June 2016.

\bibitem[Zhou et~al.(2018)Zhou, Sun, Bau, and Torralba]{Zhou2018-dz}
Zhou, B., Sun, Y., Bau, D., and Torralba, A.
\newblock Interpretable basis decomposition for visual explanation.
\newblock In \emph{Proceedings of the European Conference on Computer Vision ({ECCV})}, pp.\  119--134, 2018.

\bibitem[Zhou et~al.(2021{\natexlab{a}})Zhou, Zhang, Chen, Diao, and Zhang]{Zhou2021-tx}
Zhou, X., Zhang, W., Chen, Z., Diao, S., and Zhang, T.
\newblock Efficient neural network training via forward and backward propagation sparsification.
\newblock \emph{arXiv [cs.LG]}, November 2021{\natexlab{a}}.

\bibitem[Zhou et~al.(2021{\natexlab{b}})Zhou, Booth, Ribeiro, and Shah]{Zhou2021-lb}
Zhou, Y., Booth, S., Ribeiro, M.~T., and Shah, J.
\newblock Do feature attribution methods correctly attribute features?
\newblock \emph{arXiv [cs.LG]}, April 2021{\natexlab{b}}.

\end{thebibliography}
